\documentclass[runningheads]{llncs}

\usepackage{eccv}

\usepackage{eccvabbrv}

\usepackage{graphicx}
\usepackage{booktabs}

\usepackage[accsupp]{axessibility}  

\usepackage[pagebackref,breaklinks,colorlinks,citecolor=eccvblue]{hyperref}
\usepackage{enumitem}
\usepackage{tabularx}
\usepackage{multirow}
\usepackage{makecell}
\usepackage[table]{xcolor}
\usepackage{array}
\usepackage{subcaption}
\usepackage{wrapfig}
\usepackage{tcolorbox}

\usepackage{hyperref}

\usepackage{orcidlink}

\definecolor{MyRowBlueGray}{HTML}{8A9EB1}
\newcolumntype{G}{!{\color{gray!70}\vrule width 0.3pt}}
\newcommand{\corrauth}{\textsuperscript{\textdagger}}

\begin{document}

\title{StAR: Segment Anything Reasoner} 

\titlerunning{StAR: Segment Anything Reasoner}

\author{Seokju Yun\inst{1,2} \and
Dongheon Lee\inst{2} \and
Noori Bae\inst{2} \and
Jaesung Jun\inst{2} \and
Chanseul Cho\inst{2} \and
Youngmin Ro\inst{2}\corrauth}

\authorrunning{S.~Yun et al.}

\institute{
KAIST AI
\and
University of Seoul \\
\url{https://github.com/ysj9909/StAR}
}

\maketitle

\begingroup
\renewcommand{\thefootnote}{}
\footnotetext[0]{\mbox{\textdagger\ Corresponding author}}
\endgroup
\setcounter{footnote}{0}

\begin{abstract}
    As AI systems are being integrated more rapidly into diverse and complex real-world environments, the ability to perform holistic reasoning over an implicit query and an image to localize a target is becoming increasingly important.
    However, recent reasoning segmentation methods fail to sufficiently elicit the visual reasoning capabilities of the base model.
    In this work, we present \textbf{S}egmen\textbf{t} \textbf{A}nything \textbf{R}easoner (\textbf{StAR}), a comprehensive framework that refines the design space from multiple perspectives—including parameter-tuning scheme, reward functions, learning strategies and answer format—and achieves substantial improvements over recent baselines.
    In addition, for the first time, we successfully introduce parallel test-time scaling to the segmentation task, pushing the performance boundary even further.
    To e\underline{X}tend the scope and depth of reasoning covered by existing benchmark, we also construct the ReasonSeg-\underline{X}, which compactly defines reasoning types and includes samples that require deeper reasoning.
    Leveraging this dataset, we train StAR with a rollout-expanded selective-tuning approach to activate the base model’s latent reasoning capabilities, and establish a rigorous benchmark for systematic, fine-grained evaluation of advanced methods.
    With only 5k training samples, StAR achieves significant gains over its base counterparts across extensive benchmarks, demonstrating that our method effectively brings dormant reasoning competence to the surface.
\end{abstract}
\section{Introduction}
\begin{figure}[t]
  \centering
  \includegraphics[height=7cm]{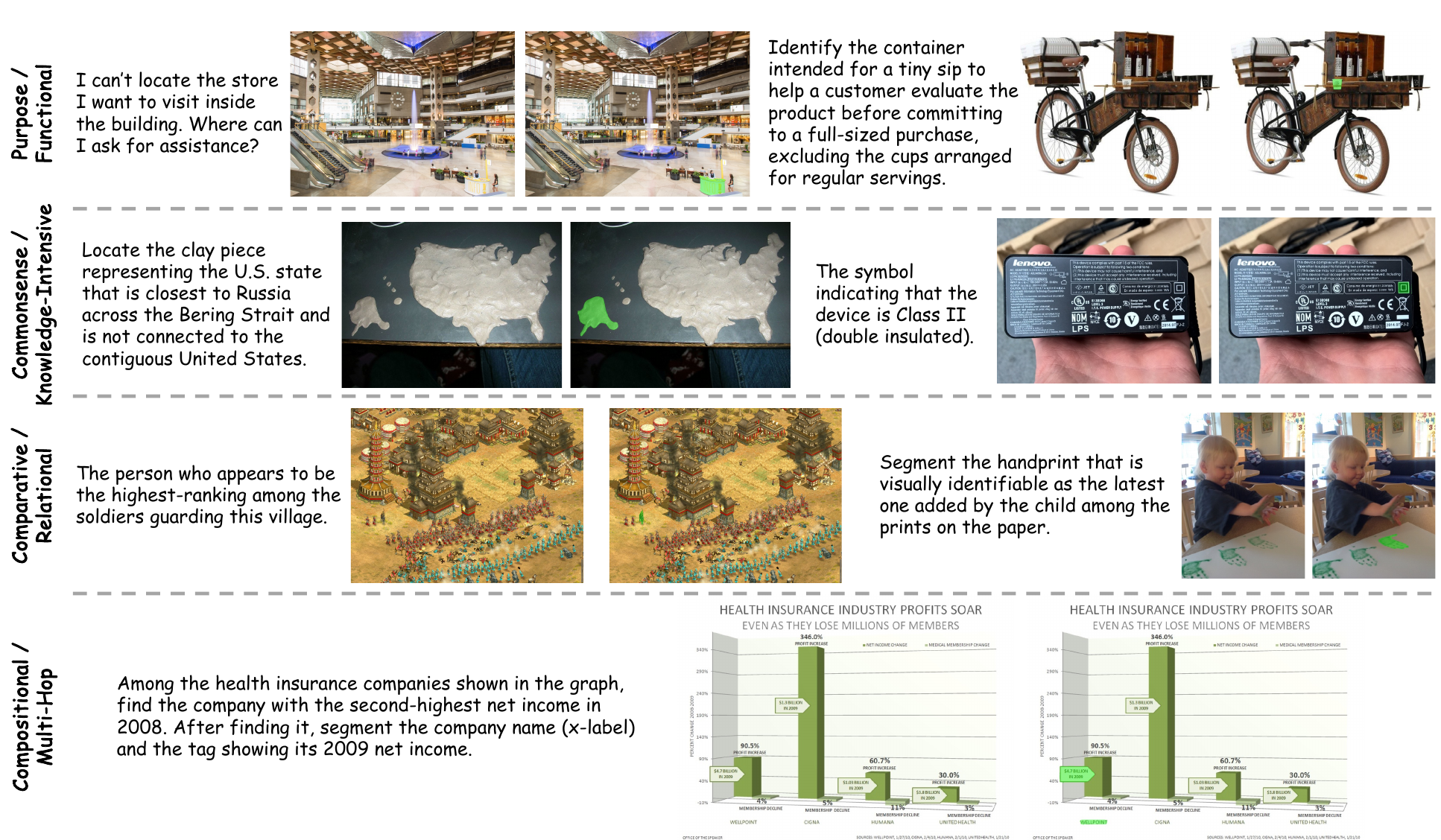}
  \caption{
  We establish a reasoning segmentation benchmark, \emph{ReasonSeg-X}, that demands a broad range of reasoning skills.
  Our dataset addresses goal-oriented reasoning and the ability to flexibly invoke world knowledge, while also covering complex relational reasoning and step-by-step reasoning capabilities.
  Our model, StAR (equipped with Qwen3-VL 32B~\cite{Qwen3-VL}), demonstrates remarkable performance across all these aspects (Our model’s prediction is shown as the green mask in the right image).
  Additional results are provided in the supplementary material.
  }
  \label{fig:teaser}
\end{figure}

Embodied AI agents are increasingly expected to operate beyond bounded lab setups, in diverse daily scenes where user intent is often implicit and incomplete.
In this regime, recognizing object categories is not sufficient; an agent must reason about the user’s goal in context and localize the relevant regions that support safe and effective action.
Reasoning segmentation~\cite{lai2024lisa} formalizes this requirement by demanding adaptive cognitive reasoning that interprets an implicit query in accordance with the context of a given image.
In particular, beyond a superficial interpretation of the scene, the ability to (i) flexibly apply information given in the query or world knowledge, (ii) comprehensively understand the background context and relations among entities, or (iii) reason step-by-step over multiple components has become an important challenge toward general artificial intelligence (see Fig.~\ref{fig:teaser}).

Following the groundwork laid by LISA~\cite{lai2024lisa}, several reasoning segmentation methods~\cite{lai2024lisa, ren2024pixellm, qian2025reasonattend, zhu2025popen, jang2025mmr, visionreasoner, liu2025seg_zero, lu-etal-2025-rsvp, sam3, bao2024cores, huang2025samr1, wang2025segllm, anonymous2026samveteran, lu2025coprs} have demonstrated promising results by integrating the versatile reasoning capabilities of multimodal large language models (MLLMs)~\cite{liu2023llava, liu2024llavanext, bai2025qwen2_5vl, wang2024qwen2vl, Qwen3-VL} with the robust mask generation ability of the SAM series~\cite{ravi2025sam2, kirillov2023sam}.
Furthermore, inspired by the expanded reasoning capacity boundary of leading MLLMs and the remarkable effectiveness of \emph{Reinforcement Learning with Verifiable Rewards} (RLVR)~\cite{guo2025deepseekr1, lambert2024tulu}, recent methods~\cite{visionreasoner, liu2025seg_zero, huang2025samr1} adopt Group Relative Policy Optimization (GRPO)~\cite{shao2024deepseekmath} for MLLMs, effectively enhancing the ability to localize target regions through visual reasoning.
For example, VisionReasoner~\cite{visionreasoner} introduces a decoupled “reasoning module-mask generator” framework and successfully applies GRPO using a compact reward design with a multi-object matching algorithm.
However, despite their empirical success, the bottleneck of current RLVR frameworks remains underexamined.
This raises a fundamental question: \textbf{\emph{Does the current RLVR framework maximally elicit visual reasoning potential that does not readily manifest? If not, which component is responsible for the bottleneck?}}

To systematically answer this question, we investigate all components that constitute the pillars of RLVR and try to identify design decisions that fully leverage the base model’s pre-trained capacity.
We start from VisionReasoner employing Qwen2.5-VL 7B~\cite{bai2025qwen2_5vl} as the MLLM backbone, and gradually “retrofit” the framework to resolve the reasoning bottleneck in each component.

\noindent Specifically, our exploration centers on the following pivotal dimensions:
\begin{itemize}[leftmargin=3mm, itemsep=0.5mm, topsep=0.5mm, partopsep=0mm]
    \item \textbf{Parameter-tuning scheme}: full-parameter tuning requiring careful hyperparameter adjustment $\rightarrow$ \underline{LoRA~\cite{hu2022lora} training} with an appropriate configurations (high-rank and large learning rate) for efficiently enhancing reasoning.
    \item \textbf{Reward design}: MLLM-level $+$ \underline{SAM-level fine-grained reward function} for mask-aware gradients and rollout scalability.
    \item \textbf{Learning strategy}: vanilla GRPO $+$ \underline{rollout-expanded selective-tuning} for propagating more informative training signals through broadened exploration.
    \item \textbf{Answer format}: bbox/point prediction + \underline{label prediction} for promoting reasoning faithfulness and visually grounded reasoning.
\end{itemize}
Through this process, we uncover several key bottlenecks that severely compromise performance.
We then eliminate these bottlenecks in a computationally efficient manner, simply yielding large performance improvements.
As a result, we propose a comprehensive framework, \textbf{S}egmen\textbf{t} \textbf{A}nything \textbf{R}easoner (\textbf{StAR}).

Enabling LLMs to improve their outputs by allocating additional test-time compute~\cite{snell2025ttsoptimally} is a critical step in post-training.
A simple yet effective approach is \emph{majority voting}~\cite{wang2023mv}, which generates $N$ responses in parallel and selects the most frequent one as the final answer.
However, while this strategy is naturally applicable to tasks such as mathematical reasoning—where the equality of two answers can be easily determined—it is not well-suited to pixel-level prediction tasks.
Therefore, instead of a per-pixel voting scheme, we adopt a \emph{mask-level} strategy that selects a representative mask set via IoU-based clustering.
By leveraging the diverse and vast reasoning pattern space of StAR, our test-time scaling (TTS) approach consistently improves performance across extensive benchmarks, enabling StAR to tackle increasingly complex problems through more inference-time compute.
Moreover, our \emph{parallel sampling} approach opens a new research direction that lies on the opposite axis of the recent SAM 3 agentic pipeline~\cite{sam3} centered on \emph{sequential refinement}.

LISA~\cite{lai2024lisa} also introduced ReasonSeg benchmark, where images are annotated with implicit text instructions and target masks.
Unfortunately, ReasonSeg has the following limitations: (i) some samples exhibit low mask quality or contain reasoning errors; (ii) its limited reasoning depth makes it difficult to effectively evaluate methods leveraging frontier MLLMs; and (iii) because the benchmark does not clearly define the reasoning scope and type it covers, it cannot systematically assess recent methods across diverse aspects.
To mitigate these, we first introduce a \underline{R}efined version of ReasonSeg (ReasonSeg-\underline{R}) that addresses issue (i), aiming to prevent currently proposed methods from being misled by incomplete evaluations and thereby support proper research progress (see Fig.~\ref{fig:reasonseg-r_ex}).
In addition, we construct an e\underline{X}tended version of ReasonSeg, termed \textbf{\emph{ReasonSeg-\underline{X}}}, which deepens the reasoning depth and systematically expands the coverage of reasoning, complementing issues (ii)–(iii).
A distinguishing characteristic of the ReasonSeg-X is its balanced coverage of samples across four reasoning types (purpose/functional, commonsense/knowledge-intensive, comparative/relational, and compositional/multi-hop reasoning—see Fig.~\ref{fig:teaser} and Sec.~\ref{sec: reasonsegx}).
We utilize this dataset to support StAR in developing its reasoning segmentation capabilities across various dimensions.

We conduct experiments on several reasoning segmentation benchmarks, including ReasonSeg~\cite{lai2024lisa}, MMR~\cite{jang2025mmr}, MUSE~\cite{ren2024pixellm}, and our proposed ReasonSeg-X/R, and observe that our StAR (i) showcases substantial performance gains over their base counterparts by preserving the base model’s inherent capacity and effectively adapting it to the segmentation task, and (ii) maximizes the efficacy of scaling test-time compute and model size by surfacing “latent, hard-to-elicit” reasoning capabilities through a train-time exploration scaling strategy that increases exposure to diverse reasoning patterns.
\vspace{-1mm}
\section{ReasonSeg-X Benchmark} \label{sec: reasonsegx}

LISA~\cite{lai2024lisa} introduces ReasonSeg, which handles implicit query requiring intricate reasoning, rather than explicit query such as category names.
Building on this, PixelLM~\cite{ren2024pixellm} proposes MUSE dataset to facilitate multi-target reasoning, while MMR~\cite{jang2025mmr} further extends the scope by curating samples that incorporate part-level reasoning.
However, while these studies primarily consider segmentation-level formulation (mask prediction format), we focus on the largely underexplored aspects of \textbf{reasoning depth and reasoning type design}.
Through this, we define the following four compact reasoning taxonomy:
\begin{itemize}[leftmargin=3mm, itemsep=0.5mm, topsep=0.5mm, partopsep=0mm]
    \item \textbf{Purpose/Functional (P/F)}: a reasoning type that requires the ability to locate a target object needed to achieve a specific purpose, or intended for a particular use.
    \item \textbf{Commonsense/Knowledge-Intensive (C/KI)}: a reasoning type that particularly requires commonsense or domain-specific knowledge to localize the target region. Moreover, this covers everyday world knowledge as well as human-like visual priors accumulated through experience.
    \item \textbf{Comparative/Relational (C/R)}: a reasoning type in which the target must be localized through comparisons or relationships between entities in the image. This type includes comparative reasoning over diverse attributes as well as holistic reasoning about inter-entity relations.
    \item \textbf{Compositional/Multi-Hop (C/MH)}: a reasoning type that requires step-by-step composition of multiple evidences, involving challenging cases that integrate complex constraints or apply various reasoning skills sequentially.
\end{itemize}

When designing this reasoning taxonomy, we aim to make each type as mutually exclusive as possible; however, in some cases the boundary inevitably becomes blurred.
For example, “finding an object for a particular use” almost always involves commonsense/knowledge (P/F $\leftrightarrow$ C/KI).
Nevertheless, to foster an agent’s goal-directed reasoning ability, we construct P/F by distinguishing it from C/KI in a strict and careful manner.
Moreover, C/R and C/MH samples exhibit reasoning depth that has been rarely covered in existing datasets, thereby posing new challenges to current methods.
Examples are shown in Fig.~\ref{fig:teaser}.

To generate our dataset, we curate diverse and context-rich images from OpenImages~\cite{OpenImages} and, based on the reasoning type design above, meticulously annotate them with implicit query and high-quality target mask.
For each generated sample, we perform cross-validation among annotators and employ GPT-5~\cite{singh2025openaigpt5} and Gemini 3~\cite{geminiteam2025gemini3} for additional verification to further establish reasoning validity.
The resulting dataset, \textbf{\emph{ReasonSeg-X}}, comprises 1,169 data samples, split into 240/156/773 samples for \texttt{train}/\texttt{val}/\texttt{test}.
Sample counts by reasoning type (in the aforementioned type order) are 53/33/88/66 for \texttt{train}, 37/36/48/35 for \texttt{val}, and 201/181/253/138 for \texttt{test}.

\begin{figure}[t]
  \centering
  \includegraphics[height=3.5cm]{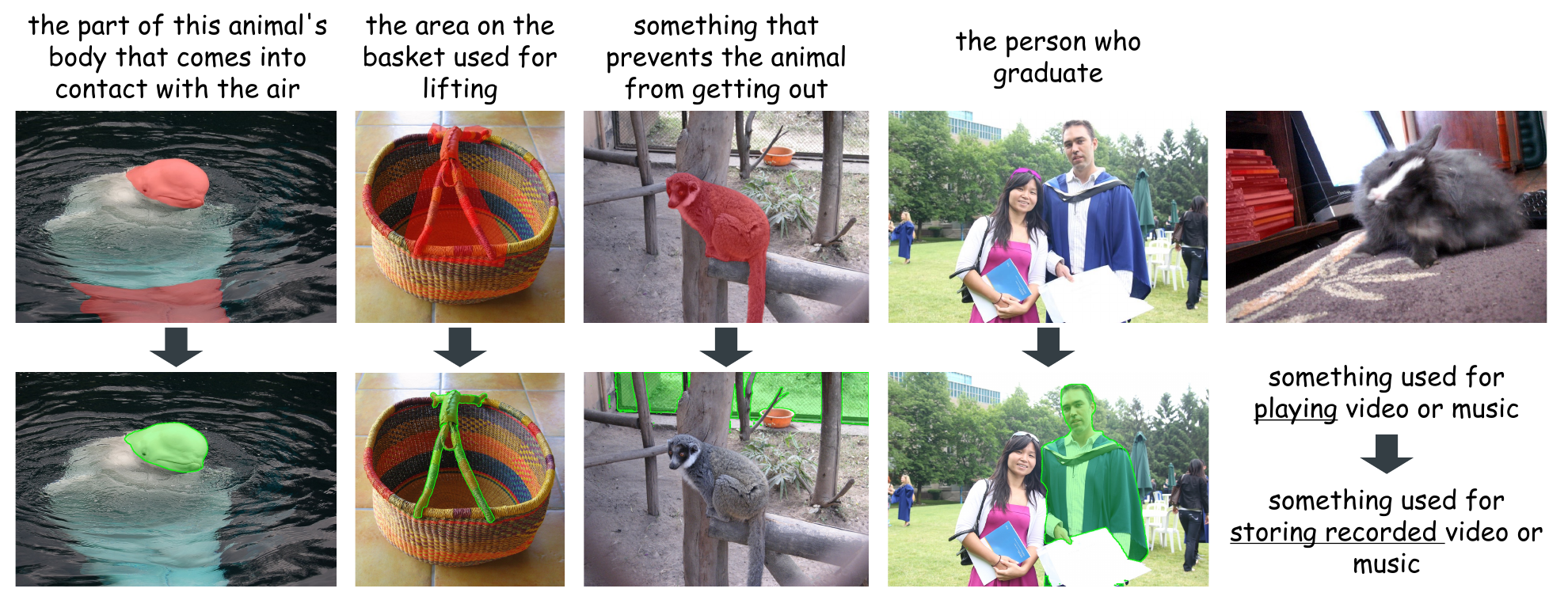}
  \vspace{-1mm}
  \caption{\textbf{ReasonSeg-R examples.}
  We correct masks for cases that are mismatched to the query and also refine mask quality.
  In addition, we modify query expressions that may inadvertently include regions outside the mask (\emph{e.g.}, the video player monitor).
  }
  \label{fig:reasonseg-r_ex}
  \vspace{-2mm}
\end{figure}

\noindent \underline{\emph{ReasonSeg-R}.}
ReasonSeg is the first benchmark specifically designed for reasoning segmentation, providing an important foundation for follow-up studies.
However, our inspection reveals that some data samples suffer from poor mask quality or contain reasoning errors (see Fig.~\ref{fig:reasonseg-r_ex}).
Therefore, to facilitate a more reliable assessment in this research area, we provide a refined version of ReasonSeg, termed \emph{ReasonSeg-R}.
Specifically, we rigorously verify query-reasoning-mask correspondence and further check whether each mask accurately covers the target region along its boundary.
As a result, we refine the masks of 113 samples and the queries of 3 samples, while excluding 13 unsuitable samples.
In addition, since many recent methods do not use the ReasonSeg \texttt{train} set, we merge the \texttt{val} and \texttt{test} sets for a streamlined evaluation process.
Additional details on our presented datasets can be found in the supplementary material.
\section{Methodology}
\vspace{-1mm}
\subsection{Pipeline Overview and Preliminaries}
\begin{figure}[t]
  \centering
  \includegraphics[height=4cm]{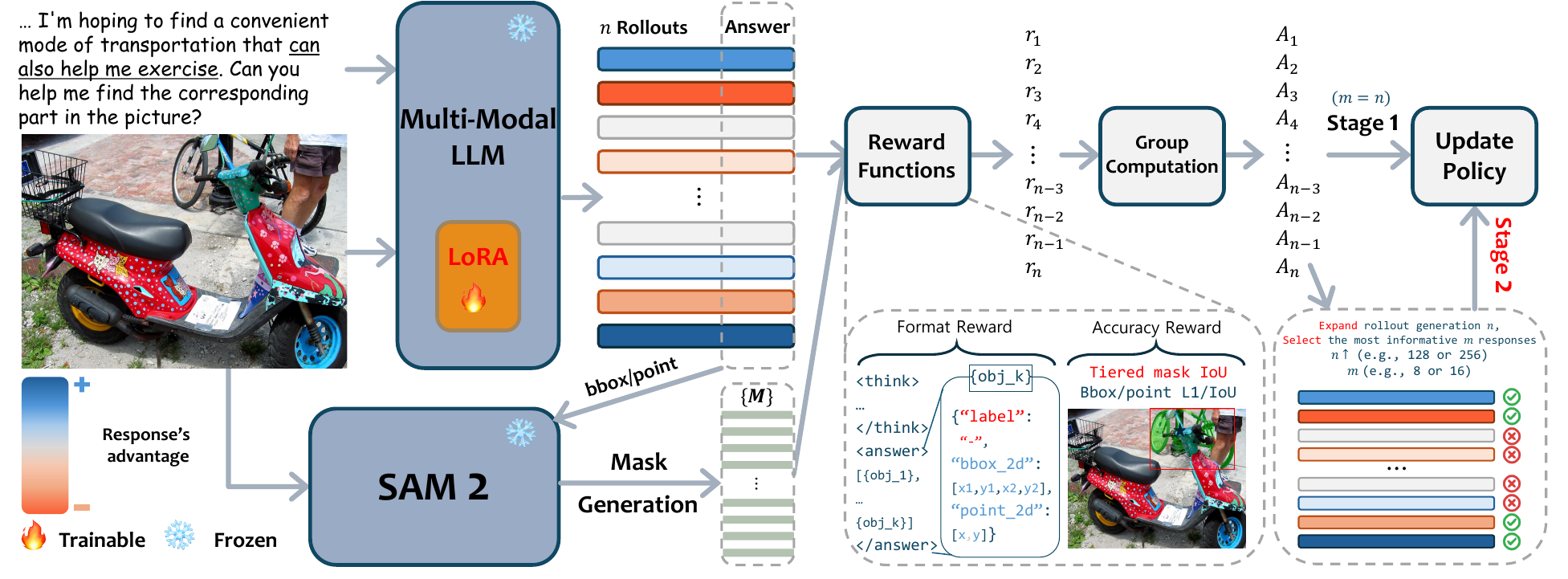}
  \caption{\textbf{Overview of the StAR framework pipeline.}
  We pinpoint and resolve reasoning bottlenecks from all perspectives of the existing reasoning segmentation framework (highlighted in red text).
  See text for details.
  }
  \label{fig:pipeline_overview}
  \vspace{-3mm}
\end{figure}

In this section, we detail the components that constitute the foundation of the StAR framework.
Following prior works~\cite{liu2025seg_zero, visionreasoner}, StAR adopts a reasoning-segmentation decoupled design: an MLLM performs reasoning over the input query-image pair to predict bbox/point coordinates, which are subsequently used as visual prompts for SAM 2~\cite{ravi2025sam2} to extract the final masks.
The pipeline of our framework is illustrated in Fig.~\ref{fig:pipeline_overview}.
Specifically, given a text query $\mathbf{T}$ and the input image $\mathbf{I}$, the reasoning module $\mathcal{F}_{\text{\tiny \emph{MLLM}}}$ generates a chain-of-thought (CoT)~\cite{wei2022cot} reasoning trace and then predicts the bounding boxes $\{\mathbf{B}_i\}_{i=1}^{N_{\text{\tiny \emph{pred}}}}$ and representative points $\{\mathbf{P}_i\}_{i=1}^{N_{\text{\tiny \emph{pred}}}}$ for the target regions associated with the query.
Lastly, these predictions are passed to the segmentation module $\mathcal{F}_{\text{\tiny \emph{SAM}}}$ to produce the final binary masks $\{\mathbf{M}_i\}_{i=1}^{N_{\text{\tiny \emph{pred}}}}$.
This procedure can be formulated as:
\begin{align}
    (\{\mathbf{B}_i, \mathbf{P}_i\})_{i=1}^{N_{\text{\tiny \emph{pred}}}} & = \mathcal{F}_{\text{\tiny \emph{MLLM}}}(\mathbf{I}, \mathbf{T}), \\
    \{\mathbf{M}_i\}_{i=1}^{N_{\text{\tiny \emph{pred}}}} & =\mathcal{F}_{\text{\tiny \emph{SAM}}}(\mathbf{I}, (\{\mathbf{B}_i, \mathbf{P}_i\})_{i=1}^{N_{\text{\tiny \emph{pred}}}}).
\end{align}

With this forward pipeline, we keep SAM 2 frozen and apply RLVR directly on the MLLM to enhance its visual reasoning capability.

\noindent\textbf{Group Relative Policy Optimization (GRPO).}
We adopt GRPO~\cite{shao2024deepseekmath}, a variant of Proximal Policy Optimization~\cite{schulman2017ppo}, as our core RLVR algorithm.
Instead of using a value function, it estimates the advantage using a \emph{normalized reward} within a group of responses to the same prompt.
Specifically, for each question $q$, GRPO first samples a group of $n$ rollouts $\{o_{i}\}^n_{i=1}$ from the old policy $\pi_{\theta_{old}}$; each rollout is then evaluated with reward functions, producing a reward vector $\{r_{i}\}^n_{i=1}$.
Finally, GRPO optimizes the policy model by maximizing the following objective:
\begin{equation}
\resizebox{0.9\linewidth}{!}{$%
\begin{split}
    \mathcal{J}_{GRPO}&(\theta) = \mathbb{E}{[q \sim P(Q), \{o_i\}_{i=1}^n \sim \pi_{\theta_{old}}(O|q)]}  \\
    & \frac{1}{n}\sum_{i=1}^n \left( \min \left( \frac{\pi_\theta(o_i |q)}{\pi_{\theta_{old}}(o_i |q)} A_i, \text{clip} \left( \frac{\pi_\theta(o_i |q)}{\pi_{\theta_{old}}(o_i |q)}, 1 - \epsilon, 1 + \epsilon \right)  A_i \right) - \beta \mathbb{D}_{KL}\left(\pi_{\theta} || \pi_{ref}\right)\right) ,
\end{split}
$}
\label{eq:GRPO-obj}
\end{equation}
where $\epsilon$ and $\beta$ are hyperparameters, and $A_i$ is the advantage, calculated using a group of rewards $\{r_{i}\}^n_{i=1}$ corresponding to the outputs inside each group:

\begin{equation}
A_i = \frac{r_i - \text{mean}(\{r_i\}_{i=1}^n)}{\text{std}(\{r_i\}_{i=1}^n)}. 
\end{equation} 

\noindent\textbf{Training Pipeline.} \label{sec:training_pipeline}
We train our model via \emph{two stages}, employing GRPO throughout both phases.
\underline{Stage 1} is primarily designed to elicit flexible localization capabilities while maximally preserving pre-trained visual perception abilities.
Accordingly, we utilize the training set collected by VisionReasoner~\cite{visionreasoner}, \emph{excluding the reasoning data samples} (LISA$++$~\cite{yang2023lisa++}).
Thus, the Stage-1 training set is sourced from LVIS~\cite{gupta2019lvis}, RefCOCOg~\cite{yu2016refcocog}, and gRefCOCO~\cite{liu2023gref}, and consists of \textbf{5k} samples in total.
\underline{Stage 2} aims to further strengthen the model’s reasoning segmentation capabilities by effectively bringing out the base model’s hidden reasoning potential.
To this end, we continue finetuning the Stage-1 model on our proposed \emph{ReasonSeg-X} \texttt{train} set (240 samples), resulting in \textbf{StAR}, the final model tailored for complex and versatile visual reasoning.

\begin{figure}[t]
  \centering
  \includegraphics[height=6.5cm]{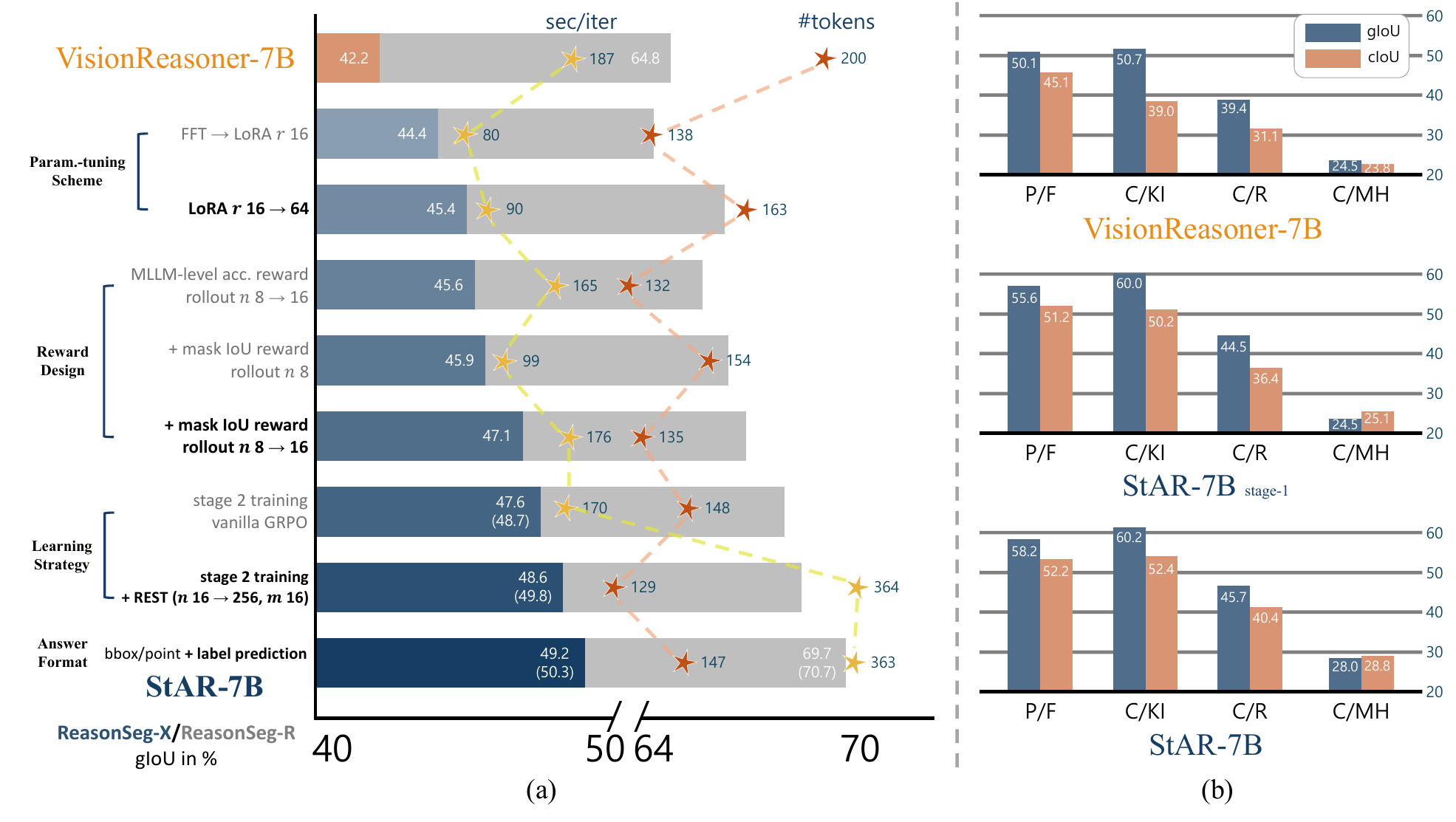}
  \vspace{-2mm}
  \caption{\textbf{(a)} We progressively retrofit all pillars of the standard RLVR framework (VisionReasoner) toward our performant and robust model (StAR), without introducing significant computational overhead.
  The foreground bars show performance on ReasonSeg-X \texttt{test}, while results on ReasonSeg-R are shown with gray bars.
  The performance values in parentheses are obtained with our majority voting strategy (Sec.~\ref{sec: mv}).
  Gray-marked design decisions (y-axis label texts) indicate comparison-only variants that are not adopted.
  \textbf{(b)} The proposed RLVR design space better preserves and exploits the base MLLM’s diverse reasoning capabilities (VisionReasoner $\rightarrow$ StAR stage-1).
  Moreover, Stage 2 training on the ReasonSeg-X \texttt{train} set with rollout-expanded selective-tuning (REST) effectively elicits the MLLM’s hidden reasoning potential, improving performance particularly on complex reasoning (\emph{e.g.}, C/MH).
  The resulting model, StAR, outperforms the baseline by a large margin across all reasoning types.
  }
  \label{fig:roadmap_star}
  \vspace{-2mm}
\end{figure}

\vspace{-1mm}
\subsection{Roadmap to Segment Anything Reasoner (StAR)}
In this section, we present the design evolution that “retrofits” VisionReasoner~\cite{visionreasoner} by alleviating key reasoning bottlenecks.
Our explorations focus on investigating performance-degrading factors along each design axis and mitigating them in a computationally efficient manner.
We start from VisionReasoner-7B and study a series of design decisions summarized as (1) parameter-tuning scheme, (2) reward design, (3) RL strategy, and (4) answer format.
As illustrated in Fig.~\ref{fig:roadmap_star} (a), our design trajectory toward StAR addresses bottlenecks across all axes and yields non-trivial performance gains ($+$7.0/4.9\% gIoU on ReasonSeg-X/R).

\noindent\textbf{Parameter-tuning scheme.}
Unlike supervised fine-tuning scenarios for dense knowledge transfer and for acquiring new reasoning patterns, applying RLVR to the reasoning segmentation (RS) task requires following a principle: “\emph{\textbf{preserve} knowledge and reasoning patterns, and \textbf{learn to leverage} them for segmentation}”.
Yet RLVR-based RS methods~\cite{visionreasoner, huang2025samr1}, as well as most existing open-source reasoning models~\cite{wang2025vlrethinker, muennighoff2025s1tts, yu2025dapo, xu2025llavacot}, typically rely on expensive \emph{full-parameter tuning}.
This approach is notoriously resource-intensive and often requires extra model copies—such as a reference model for KL penalty—greatly increasing memory usage; moreover, adjusting all weights risks substantially damaging the base model’s underlying knowledge.
We therefore investigate \emph{parameter-efficient tuning} to maximally maintain the base model’s inherent capacity while efficiently enhancing reasoning capabilities.

Low-rank adaptation (LoRA)~\cite{hu2022lora} uses a low-rank decomposition to model sparse updates~\cite{liang2025lorasculpt}, thereby largely preserving the base model’s world knowledge and demonstrating robust performance~\cite{yun2025soma, wang2025milora, biderman2024loraforgetless, wang2026tina}.
Building on these observations, we attempt to employ LoRA in the RLVR setting.
However, when using the default hyperparameter configuration, we observe a substantial performance drop compared to the full-parameter tuning baseline.
To better promote task adaptation, we increase the learning rate from $1\times10^{-6}$ to $1\times10^{-5}$ and exclude the KL penalty term.
With this configuration and a LoRA rank of 16, we achieve comparable or even superior performance while reducing training costs by more than 2× compared to VisionReasoner.
We then increase the LoRA rank from 16 to 64 to expand the expressiveness of the update matrices, allowing the model to tackle more complex reasoning~\cite{huang2025hira}.
Overall, our derived tuning scheme achieves surprisingly strong performance (42.2\% $\rightarrow$ 45.4\%/64.8\% $\rightarrow$ 66.1\%) \emph{by following the aforementioned principle.}

\begin{figure}[t]
  \centering
  \includegraphics[height=3.5cm]{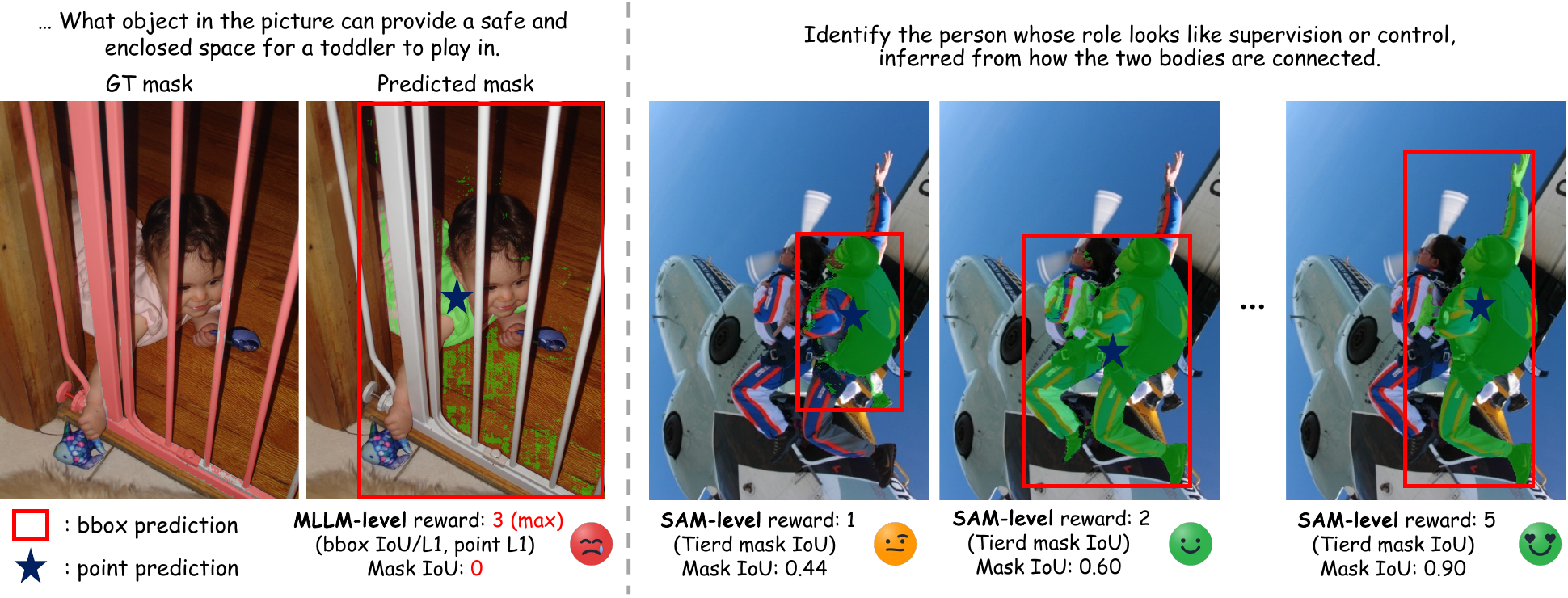}
  \vspace{-2mm}
  \caption{\textbf{Motivation for the mask IoU reward.}
  \emph{Left}: Misalignment between the MLLM-level reward function and the task's final objective potentially induces confusing learning signals.
  \emph{Right}: Distinguishing between diverse mask defects via a fine-grained tiered mask reward promotes stable and stepwise optimization.
  }
  \label{fig:mask_reward}
  \vspace{-3mm}
\end{figure}

\noindent\textbf{Reward design.}
Reward functions are central to RLVR.
However, the accuracy reward functions used in VisionReasoner are not tightly aligned with the task’s ultimate goal—mask generation—increasing the risk of reward hacking (Fig.~\ref{fig:mask_reward}, \emph{left}).
In addition, the prevailing “hard” reward (1 if IoU exceeds a fixed threshold, otherwise 0) neither reflects incremental improvements throughout training nor enables fine-grained evaluation.
To remedy these issues, inspired by~\cite{huang2025samr1}, we incorporate SAM into the RLVR training loop and introduce a piecewise mask-IoU reward to provide mask-aware feedback. 
This design provides graded signals that directly encourage incremental improvements in mask quality (Fig.~\ref{fig:mask_reward}, \emph{right}).

We employ a tiered mask reward function based on IoU to discretize segmentation quality into multiple levels.
The reward is computed as follows (additions relative to SAM-R1~\cite{huang2025samr1} are marked in \textcolor{red}{red}):
\vspace{-1mm}
\begin{equation}
\scalebox{0.6}{$
\text{mask reward} =
\begin{cases}
\textcolor{red}{5}, & \textcolor{red}{\text{IoU} > 0.90},\\
4, & 0.80 < \text{IoU} \le 0.90,\\
3, & 0.70 < \text{IoU} \le 0.80,\\
2, & 0.50 < \text{IoU} \le 0.70,\\
\textcolor{red}{1}, & \textcolor{red}{0.30 < \text{IoU} \le 0.50},\\
0, & \text{otherwise}.
\end{cases}
$}
\label{eq: mask_reward}
\vspace{-2mm}
\end{equation}
Along with this SAM-level reward, we also retain the reward functions from VisionReasoner that evaluate the accuracy of the MLLM’s predictions.
Specifically, we use (i) a bbox IoU reward that assigns 1 when the predicted bbox exceeds 0.5 IoU with the GT bbox, and (ii) a bbox/point L1 reward that assigns 1 when the L1 distance between predicted and GT bbox/point is below 10/30 pixels.
We then solve the multi-object matching problem over $N_{\text{\tiny \emph{pred}}}$ predictions and $N_{\text{\tiny \emph{GT}}}$ instance annotations using a batched Hungarian algorithm~\cite{visionreasoner}.
Based on the resulting assignment, the final accuracy reward is calculated and then divided by $\max\{N_{\text{\tiny \emph{pred}}}, N_{\text{\tiny \emph{GT}}}\}$ to suppress over- and under-segmentation errors.
Additionally, VisionReasoner’s Thinking/Answer/Non-repeat reward functions are applied to ensure a structured and reliable output format.

Maintaining MLLM-level accuracy rewards bolsters coarse localization capabilities, particularly by imposing stronger penalties on reasoning failures.
Moreover, as illustrated in Fig.~\ref{fig:mask_reward} \emph{right}, our fine-grained mask reward delivers a \emph{curriculum-like training signal}: early training emphasizes locating the target via high-quality reasoning, while later stages focus on accurately segmenting target region boundaries.
With this comprehensive reward design, increasing rollout utilization (8 → 16) brings meaningful performance gains—unlike VisionReasoner—suggesting that our rewards discriminate rollouts more finely and offer the
richest feedback.
Notably, by reallocating the resources saved by LoRA training to rollout scaling, we improve performance (45.4\% $\rightarrow$ 47.1\%/66.1\% $\rightarrow$ 66.6\%) while keeping overall training costs modest.

\begin{figure}[t]
  \centering
  \includegraphics[height=2.8cm]{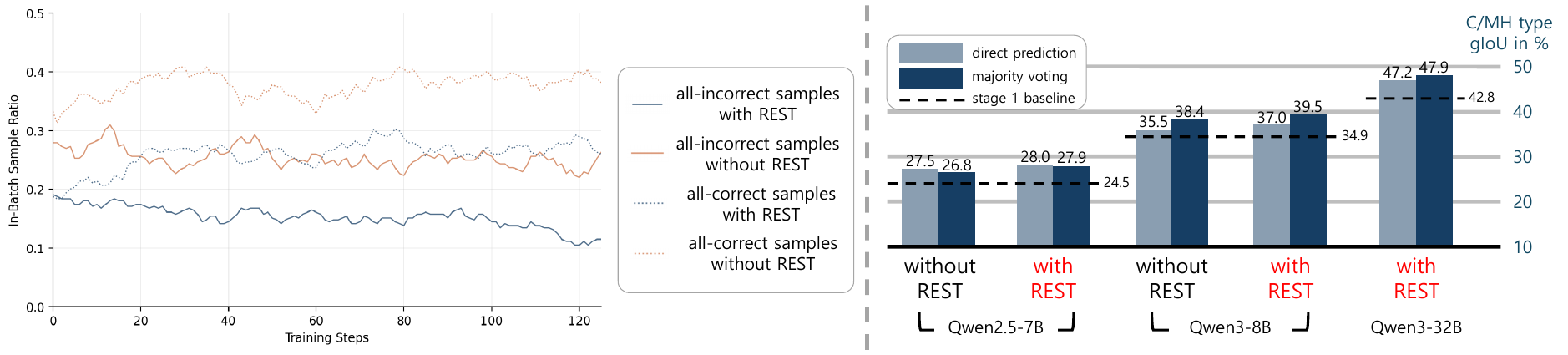}
  \caption{
  \emph{Left}: Rollout-expanded selective-tuning (REST) substantially improves sample utilization.
  In particular, for hard samples, it develops the model’s intricate reasoning by exploring a vast space of reasoning paths.
  “Correct” is defined as IoU > 0.5, and results are reported with the Qwen3-VL 8B model.
  \emph{Right}: REST effectively boosts performance on problems that require multi-step reasoning; moreover, \textbf{its gains synergistically scale with the base model’s reasoning capability boundaries.}
  }
  \label{fig:rest}
  \vspace{-3mm}
\end{figure}

\noindent\textbf{Reinforcement learning (RL) strategy.}
While Stage 1 training primarily leverages referring segmentation datasets and focuses on utilizing the base model’s perceptual capacity for localization, Stage 2 mainly aims to cultivate the ability to solve complex reasoning problems using our proposed dataset.
However, Stage 2 GRPO training often suffers from \emph{advantage vanishing}: for samples that are too easy or too hard, the model generates group responses with consistent rewards, severely reducing learning efficiency.
As depicted in Fig.~\ref{fig:rest} \emph{left}, during vanilla GRPO training, on average, more than half of the batch samples exhibit a near-zero-variance reward distribution.
Moreover, the sparse positive feedback on extremely challenging problems prevents fully drawing out the base model’s hidden reasoning potential.

To mitigate this, we propose Rollout-Expanded Selective-Tuning (REST) as a drop-in modification to GRPO.
REST \textbf{decouples exploration from learning}: we first \emph{expand} the rollout pool from the conventional small-$n$ regime (\emph{e.g.,} 8-16) to a large-$N$ regime (\emph{e.g.,} 128-256) to explore a much wider set of reasoning trajectories.
Instead of updating on all $N$ rollouts, REST then performs \emph{advantage-extremal selection}, updating only on an informative subset of size $m$: the $m/2$ rollouts with the largest advantages and the $m/2$ rollouts with the smallest advantages.
Selecting both tails aligns with the reward variance-maximization strategies~\cite{razin2025what, pods}, yielding stronger contrastive signals for policy improvement.

Furthermore, REST cleverly exploits the compute/memory asymmetry in LLM RL.
\underline{The rollout generation} is highly parallel, requires minimal activation storage, and is typically high-throughput when batched—so generating many rollouts can be amortized efficiently.
By contrast, \underline{policy updates} are memory- and communication-intensive because of optimizer states and cross-device synchronization.
Consequently, exploring a 16× larger reasoning pattern space ($n$ 16 $\rightarrow$ 256) increases the end-to-end training wall-clock time by only about 2× (Fig.~\ref{fig:roadmap_star} (a)).
In other words, REST keeps the expensive part nearly constant while scaling the cheaper, batched inference part.

Importantly, for hard prompts, REST increases the chance of observing rare successful reasoning paths, efficiently eliciting and internalizing “latent, dormant” intricate reasoning skills (Fig.~\ref{fig:rest}).
Leveraging our dataset with REST for Stage-2 training boosts performance on ReasonSeg-X/R by 1.5/1.6\%, respectively.

\noindent\textbf{Answer format.}
Reasoning segmentation is inherently \emph{semantic-first}: the model must interpret the query, perform compositional visual-language inference, and identify target regions consistent with that semantics.
However, the endpoint of the task is \emph{geometric}, which can implicitly bias generation toward a “geometry-only” mode where the model focuses on producing plausible coordinates without committing to a grounded semantic identification.
Meanwhile, recent studies~\cite{xu2025morethinking, tian2026morethought} reveal that as CoT reasoning progresses, visual attention tends to markedly diminish, which in turn exacerbates hallucination.
We therefore introduce Label Prediction (LP) as a simple and lightweight answer-format augmentation.
Rather than predicting only geometry for each target, the model is additionally required to emit a short semantic label that describes the localized entity (\emph{e.g.}, \{“label”: “the owner of the largest dog”, “bbox\_2d”: ..., “point\_2d”: ...\}).
Crucially, while this modification merely adjusts the prompt/response schema, it systematically reshapes generation dynamics by introducing an explicit semantic “naming” step immediately before coordinate prediction.

Interestingly, predicted labels often paraphrase expressions in the query or take the form of answering the query, thereby semantically anchoring the predicted region to the query and ultimately enhancing reasoning faithfulness.
LP also increases StAR-7B’s attention mass over visual tokens during coordinate prediction from 5.3\% to 6.3\%.
This suggests that, in contrast to methods that explicitly strengthen visual dependency during training/inference~\cite{huang2026tokenperception, jung2025visualattention}, a simple form of strategic prompting—requiring label prediction—can efficiently and effortlessly facilitate visually-grounded reasoning.
As a result, this remarkably simple solution improves performance from 48.6/68.2\% to 49.2/69.7\%.
\emph{This design trajectory brings us to our final model, \textbf{StAR}.}

\vspace{-2mm}
\subsection{Majority Voting for Segmentation} \label{sec: mv}
Majority voting~\cite{wang2023mv} has proven effective for tasks with a discrete answer space, where “voting” is naturally well-defined.
However, majority voting remains largely underexplored in segmentation tasks because segmentation produces pixel-level, spatially structured mask outputs for which defining agreement and aggregating multiple predictions are inherently nontrivial.
\emph{Here, we introduce a mask-level clustering strategy that adapts majority voting to pixel-level, multi-object prediction.}
Given a pool of mask candidates collected from all parallelly generated responses, we group all candidates into clusters using a greedy procedure: for each candidate, we compute its IoU with the representative mask of existing clusters; if the IoU exceeds a certain threshold ($=0.85$), we assign it to that cluster, otherwise we create a new cluster.
We then decide the number of target instances $\hat{K}$ by the mode of the per-response predicted target counts, and if \emph{no-target} prediction is in the majority, we also output “no target object”.
Subsequently, we rank clusters by vote count and select the top $\min(\hat{K},\#\text{clusters})$ clusters.
For each selected cluster, we choose the single most confident instance mask, leveraging SAM’s mask quality score~\cite{ravi2025sam2} for robust aggregation.
Finally, we compose the overall prediction by taking the union of the selected instance masks.
\textbf{This strategy is generally applicable to any MLLM-based segmentation framework and brings substantial performance gains when paired with our StAR's vast reasoning capacity boundary (see Tab.~\ref{tab:reasonseg_results}).}

\emph{Further details and a discussion of the methodology can be found in the supplementary material.}

\begin{table*}[t]
\tiny
\centering
\setlength{\tabcolsep}{2pt}
\caption{\textbf{Performance results on the ReasonSeg (RS) series.}
“MV” denotes our majority voting strategy adapted for segmentation.
We compare our StAR against state-of-the-art baselines: LISA~\cite{lai2024lisa}, SegLLM~\cite{wang2025segllm}, READ~\cite{qian2025reasonattend}, CoReS~\cite{bao2024cores}, RSVP~\cite{lu-etal-2025-rsvp}, CoPRS~\cite{lu2025coprs}, SAM-R1~\cite{huang2025samr1}, DPAD~\cite{DPAD}, SAM-Veteran~\cite{anonymous2026samveteran}, SegCompass~\cite{lu2026segcompass},  DR$^2$Seg~\cite{he2026drseg}, VisionReasoner~\cite{visionreasoner}, Seg-ReSearch~\cite{liang2026segresearch}, SAM 3 Agent~\cite{sam3}, and RSAgent~\cite{he2026rsagent}.
}
\vspace{-2mm}
\begin{tabularx}{0.99\linewidth}{
  l
  |c
  |*{2}{>{\centering\arraybackslash}X}
  |*{2}{>{\centering\arraybackslash}X}
  |*{2}{>{\centering\arraybackslash}X}
  |*{2}{>{\centering\arraybackslash}X} 
  G*{2}{>{\centering\arraybackslash}X} 
  G*{2}{>{\centering\arraybackslash}X} 
  G*{2}{>{\centering\arraybackslash}X} 
  G*{2}{>{\centering\arraybackslash}X} 
}
\toprule
\multirow{3}{*}{\textbf{Method}}
& \multirow{3}{*}{\makecell{\textbf{\scalebox{0.85}{Base model}}\\\textbf{\scalebox{0.85}{(Size)}}}}
& \multicolumn{2}{c|}{\textbf{RS}}
& \multicolumn{2}{c|}{\textbf{RS-R}}
& \multicolumn{2}{c|}{\textbf{RS-X}}
& \multicolumn{10}{c}{\textbf{RS-X test}} \\
\cmidrule(lr){3-4}\cmidrule(lr){5-6}\cmidrule(lr){7-8}\cmidrule(lr){9-18}
&& \multicolumn{2}{c|}{test}
& \multicolumn{2}{c|}{--}
& \multicolumn{2}{c|}{val}
& \multicolumn{2}{c|}{overall}
& \multicolumn{2}{c|}{P/F}
& \multicolumn{2}{c|}{C/KI}
& \multicolumn{2}{c|}{C/R}
& \multicolumn{2}{c}{C/MH} \\
\cmidrule(lr){3-4}\cmidrule(lr){5-6}\cmidrule(lr){7-8}
\cmidrule(lr){9-10}\cmidrule(lr){11-12}\cmidrule(lr){13-14}\cmidrule(lr){15-16}\cmidrule(lr){17-18}
&
& \scalebox{0.8}{gIoU} & \scalebox{0.8}{cIoU}
& \scalebox{0.8}{gIoU} & \scalebox{0.8}{cIoU}
& \scalebox{0.8}{gIoU} & \scalebox{0.8}{cIoU}
& \scalebox{0.8}{gIoU} & \scalebox{0.8}{cIoU}
& \scalebox{0.8}{gIoU} & \scalebox{0.8}{cIoU}
& \scalebox{0.8}{gIoU} & \scalebox{0.8}{cIoU}
& \scalebox{0.8}{gIoU} & \scalebox{0.8}{cIoU}
& \scalebox{0.8}{gIoU} & \scalebox{0.8}{cIoU} \\

\midrule
\rowcolor{gray!10}\multicolumn{18}{l}{\textit{\textcolor{gray}{Supervised/Reinforcement Finetuning Methods}}} \\

\scalebox{0.75}{LISA (ft)} & \scalebox{0.65}{Llama2 13B}
& \scalebox{0.9}{51.5} & \scalebox{0.9}{51.3}  
& \scalebox{0.9}{52.5} & \scalebox{0.9}{53.3}  
& \scalebox{0.9}{23.8} & \scalebox{0.9}{24.3}  
& \scalebox{0.9}{25.1} & \scalebox{0.9}{26.0}  
& \scalebox{0.9}{27.9} & \scalebox{0.9}{30.6}  
& \scalebox{0.9}{28.2} & \scalebox{0.9}{29.2}  
& \scalebox{0.9}{26.8} & \scalebox{0.9}{25.8}  
& \scalebox{0.9}{13.8} & \scalebox{0.9}{16.8}  
\\

\scalebox{0.75}{SegLLM} & \scalebox{0.65}{LLaVA1.5 7B}
& \scalebox{0.9}{52.4} & \scalebox{0.9}{48.4}  
& \scalebox{0.9}{--} & \scalebox{0.9}{--}  
& \scalebox{0.9}{--} & \scalebox{0.9}{--}  
& \scalebox{0.9}{--} & \scalebox{0.9}{--}  
& \scalebox{0.9}{--} & \scalebox{0.9}{--}  
& \scalebox{0.9}{--} & \scalebox{0.9}{--}  
& \scalebox{0.9}{--} & \scalebox{0.9}{--}  
& \scalebox{0.9}{--} & \scalebox{0.9}{--}  
\\

\scalebox{0.75}{READ (ft)} & \scalebox{0.65}{LLaVA1.5 13B}
& \scalebox{0.9}{62.2} & \scalebox{0.9}{62.8}  
& \scalebox{0.9}{--} & \scalebox{0.9}{--}  
& \scalebox{0.9}{--} & \scalebox{0.9}{--}  
& \scalebox{0.9}{--} & \scalebox{0.9}{--}  
& \scalebox{0.9}{--} & \scalebox{0.9}{--}  
& \scalebox{0.9}{--} & \scalebox{0.9}{--}  
& \scalebox{0.9}{--} & \scalebox{0.9}{--}  
& \scalebox{0.9}{--} & \scalebox{0.9}{--}  
\\

\scalebox{0.75}{CoReS (ft)} & \scalebox{0.65}{LLaVA1.5 13B}
& \scalebox{0.9}{65.5} & \scalebox{0.9}{--}  
& \scalebox{0.9}{--} & \scalebox{0.9}{--}  
& \scalebox{0.9}{--} & \scalebox{0.9}{--}  
& \scalebox{0.9}{--} & \scalebox{0.9}{--}  
& \scalebox{0.9}{--} & \scalebox{0.9}{--}  
& \scalebox{0.9}{--} & \scalebox{0.9}{--}  
& \scalebox{0.9}{--} & \scalebox{0.9}{--}  
& \scalebox{0.9}{--} & \scalebox{0.9}{--}  
\\

\scalebox{0.75}{RSVP} & \scalebox{0.65}{ GPT-4o}
& \scalebox{0.9}{60.3} & \scalebox{0.9}{60.0}  
& \scalebox{0.9}{--} & \scalebox{0.9}{--}  
& \scalebox{0.9}{--} & \scalebox{0.9}{--}  
& \scalebox{0.9}{--} & \scalebox{0.9}{--}  
& \scalebox{0.9}{--} & \scalebox{0.9}{--}  
& \scalebox{0.9}{--} & \scalebox{0.9}{--}  
& \scalebox{0.9}{--} & \scalebox{0.9}{--}  
& \scalebox{0.9}{--} & \scalebox{0.9}{--}  
\\

\scalebox{0.75}{CoPRS} & \scalebox{0.65}{Qwen2.5-VL 7B}
& \scalebox{0.9}{59.8} & \scalebox{0.9}{55.1}  
& \scalebox{0.9}{--} & \scalebox{0.9}{--}  
& \scalebox{0.9}{--} & \scalebox{0.9}{--}  
& \scalebox{0.9}{--} & \scalebox{0.9}{--}  
& \scalebox{0.9}{--} & \scalebox{0.9}{--}  
& \scalebox{0.9}{--} & \scalebox{0.9}{--}  
& \scalebox{0.9}{--} & \scalebox{0.9}{--}  
& \scalebox{0.9}{--} & \scalebox{0.9}{--}  
\\

\scalebox{0.75}{SAM-R1} & \scalebox{0.65}{Qwen2.5-VL 7B}
& \scalebox{0.9}{60.2} & \scalebox{0.9}{54.3}  
& \scalebox{0.9}{--} & \scalebox{0.9}{--}  
& \scalebox{0.9}{--} & \scalebox{0.9}{--}  
& \scalebox{0.9}{--} & \scalebox{0.9}{--}  
& \scalebox{0.9}{--} & \scalebox{0.9}{--}  
& \scalebox{0.9}{--} & \scalebox{0.9}{--}  
& \scalebox{0.9}{--} & \scalebox{0.9}{--}  
& \scalebox{0.9}{--} & \scalebox{0.9}{--}  
\\

\scalebox{0.75}{DPAD} & \scalebox{0.65}{Qwen2.5-VL 7B}
& \scalebox{0.9}{60.8} & \scalebox{0.9}{57.5}  
& \scalebox{0.9}{--} & \scalebox{0.9}{--}  
& \scalebox{0.9}{--} & \scalebox{0.9}{--}  
& \scalebox{0.9}{--} & \scalebox{0.9}{--}  
& \scalebox{0.9}{--} & \scalebox{0.9}{--}  
& \scalebox{0.9}{--} & \scalebox{0.9}{--}  
& \scalebox{0.9}{--} & \scalebox{0.9}{--}  
& \scalebox{0.9}{--} & \scalebox{0.9}{--}  
\\

\scalebox{0.75}{SAM-Veteran} & \scalebox{0.65}{Qwen2.5-VL 7B}
& \scalebox{0.9}{62.6} & \scalebox{0.9}{56.1}  
& \scalebox{0.9}{--} & \scalebox{0.9}{--}  
& \scalebox{0.9}{--} & \scalebox{0.9}{--}  
& \scalebox{0.9}{--} & \scalebox{0.9}{--}  
& \scalebox{0.9}{--} & \scalebox{0.9}{--}  
& \scalebox{0.9}{--} & \scalebox{0.9}{--}  
& \scalebox{0.9}{--} & \scalebox{0.9}{--}  
& \scalebox{0.9}{--} & \scalebox{0.9}{--}  
\\

\scalebox{0.75}{SegCompass} & \scalebox{0.65}{Qwen2.5-VL 7B}
& \scalebox{0.9}{64.0} & \scalebox{0.9}{64.8}  
& \scalebox{0.9}{--} & \scalebox{0.9}{--}  
& \scalebox{0.9}{--} & \scalebox{0.9}{--}  
& \scalebox{0.9}{--} & \scalebox{0.9}{--}  
& \scalebox{0.9}{--} & \scalebox{0.9}{--}  
& \scalebox{0.9}{--} & \scalebox{0.9}{--}  
& \scalebox{0.9}{--} & \scalebox{0.9}{--}  
& \scalebox{0.9}{--} & \scalebox{0.9}{--}  
\\


\scalebox{0.75}{DR$^2$Seg} & \scalebox{0.65}{Qwen2.5-VL 7B}
& \scalebox{0.9}{66.1} & \scalebox{0.9}{63.6}  
& \scalebox{0.9}{67.4} & \scalebox{0.9}{61.4}  
& \scalebox{0.9}{46.8} & \scalebox{0.9}{43.1}  
& \scalebox{0.9}{45.2} & \scalebox{0.9}{36.6}  
& \scalebox{0.9}{52.1} & \scalebox{0.9}{40.9}  
& \scalebox{0.9}{55.9} & \scalebox{0.9}{47.3}  
& \scalebox{0.9}{43.9} & \scalebox{0.9}{34.1}  
& \scalebox{0.9}{23.7} & \scalebox{0.9}{25.7}  
\\

\scalebox{0.75}{VisionReasoner} & \scalebox{0.65}{Qwen2.5-VL 7B}
& \scalebox{0.9}{63.6} & \scalebox{0.9}{55.7}  
& \scalebox{0.9}{64.8} & \scalebox{0.9}{56.8}  
& \scalebox{0.9}{44.1} & \scalebox{0.9}{37.6}  
& \scalebox{0.9}{42.2} & \scalebox{0.9}{33.8}  
& \scalebox{0.9}{50.1} & \scalebox{0.9}{45.1}  
& \scalebox{0.9}{50.7} & \scalebox{0.9}{39.0}  
& \scalebox{0.9}{39.4} & \scalebox{0.9}{31.1}  
& \scalebox{0.9}{24.5} & \scalebox{0.9}{23.8}  
\\

\rowcolor{MyRowBlueGray!30} \scalebox{0.8}{StAR$_{\text{ stage-1}}$} & \scalebox{0.65}{Qwen2.5-VL 7B}
& \scalebox{0.9}{66.7} & \scalebox{0.9}{60.8}  
& \scalebox{0.9}{69.0} & \scalebox{0.9}{65.6}  
& \scalebox{0.9}{48.5} & \scalebox{0.9}{42.7}  
& \scalebox{0.9}{47.4} & \scalebox{0.9}{40.6}  
& \scalebox{0.9}{55.6} & \scalebox{0.9}{51.2}  
& \scalebox{0.9}{60.0} & \scalebox{0.9}{50.2}  
& \scalebox{0.9}{44.5} & \scalebox{0.9}{36.4}  
& \scalebox{0.9}{24.5} & \scalebox{0.9}{25.1}  
\\

\scalebox{0.75}{Seg-ReSearch} & \scalebox{0.65}{Qwen3-VL 8B}
& \scalebox{0.9}{67.4} & \scalebox{0.9}{59.0}  
& \scalebox{0.9}{--} & \scalebox{0.9}{--}  
& \scalebox{0.9}{--} & \scalebox{0.9}{--}  
& \scalebox{0.9}{--} & \scalebox{0.9}{--}  
& \scalebox{0.9}{--} & \scalebox{0.9}{--}  
& \scalebox{0.9}{--} & \scalebox{0.9}{--}  
& \scalebox{0.9}{--} & \scalebox{0.9}{--}  
& \scalebox{0.9}{--} & \scalebox{0.9}{--}  
\\

\midrule

\rowcolor{gray!10}\multicolumn{18}{l}{\textit{\textcolor{gray}{Inference-time Scaling Methods / Multi-round Tool-calling Agents}}} \\

\scalebox{0.75}{SAM 3 Agent} & \scalebox{0.65}{Qwen2.5-VL 7B}
& \scalebox{0.9}{62.6} & \scalebox{0.9}{56.2}  
& \scalebox{0.9}{63.1} & \scalebox{0.9}{58.0}  
& \scalebox{0.9}{34.1} & \scalebox{0.9}{28.0}  
& \scalebox{0.9}{34.4} & \scalebox{0.9}{29.5}  
& \scalebox{0.9}{41.7} & \scalebox{0.9}{37.9}  
& \scalebox{0.9}{37.7} & \scalebox{0.9}{31.8}  
& \scalebox{0.9}{35.3} & \scalebox{0.9}{26.0}  
& \scalebox{0.9}{17.9} & \scalebox{0.9}{21.9}  
\\

\scalebox{0.75}{SAM 3 Agent} & \scalebox{0.65}{Qwen3-VL 8B}
& \scalebox{0.9}{70.2} & \scalebox{0.9}{67.3}  
& \scalebox{0.9}{69.3} & \scalebox{0.9}{64.1}  
& \scalebox{0.9}{45.9} & \scalebox{0.9}{37.7}  
& \scalebox{0.9}{42.3} & \scalebox{0.9}{39.7}  
& \scalebox{0.9}{50.6} & \scalebox{0.9}{43.6}  
& \scalebox{0.9}{48.4} & \scalebox{0.9}{45.1}  
& \scalebox{0.9}{40.5} & \scalebox{0.9}{40.8}  
& \scalebox{0.9}{25.8} & \scalebox{0.9}{26.0}  
\\

\scalebox{0.75}{SAM 3 Agent} & \scalebox{0.65}{Qwen2.5-VL 72B}
& \scalebox{0.9}{71.8} & \scalebox{0.9}{65.2}  
& \scalebox{0.9}{72.4} & \scalebox{0.9}{65.3}  
& \scalebox{0.9}{52.0} & \scalebox{0.9}{43.0}  
& \scalebox{0.9}{49.8} & \scalebox{0.9}{40.3}  
& \scalebox{0.9}{57.8} & \scalebox{0.9}{50.5}  
& \scalebox{0.9}{55.5} & \scalebox{0.9}{45.4}  
& \scalebox{0.9}{49.6} & \scalebox{0.9}{34.0}  
& \scalebox{0.9}{30.8} & \scalebox{0.9}{35.1}  
\\

\scalebox{0.75}{RSAgent} & \scalebox{0.65}{Qwen2.5-VL 7B}
& \scalebox{0.9}{66.5} & \scalebox{0.9}{57.9}  
& \scalebox{0.9}{--} & \scalebox{0.9}{--}  
& \scalebox{0.9}{--} & \scalebox{0.9}{--}  
& \scalebox{0.9}{--} & \scalebox{0.9}{--}  
& \scalebox{0.9}{--} & \scalebox{0.9}{--}  
& \scalebox{0.9}{--} & \scalebox{0.9}{--}  
& \scalebox{0.9}{--} & \scalebox{0.9}{--}  
& \scalebox{0.9}{--} & \scalebox{0.9}{--}  
\\

\midrule

\rowcolor{MyRowBlueGray!50} \scalebox{0.8}{StAR} & \scalebox{0.65}{Qwen2.5-VL 7B}
& \scalebox{0.9}{67.5} & \scalebox{0.9}{61.3}  
& \scalebox{0.9}{69.7} & \scalebox{0.9}{66.2}  
& \scalebox{0.9}{50.5} & \scalebox{0.9}{48.0}  
& \scalebox{0.9}{49.2} & \scalebox{0.9}{43.6}  
& \scalebox{0.9}{58.2} & \scalebox{0.9}{52.2}  
& \scalebox{0.9}{60.2} & \scalebox{0.9}{52.4}  
& \scalebox{0.9}{45.7} & \scalebox{0.9}{40.4}  
& \scalebox{0.9}{28.0} & \scalebox{0.9}{28.8}  
\\

\rowcolor{MyRowBlueGray!70} \scalebox{0.8}{StAR $+$ MV} & \scalebox{0.65}{Qwen2.5-VL 7B}
& \scalebox{0.9}{68.5} & \scalebox{0.9}{65.0}  
& \scalebox{0.9}{70.7} & \scalebox{0.9}{67.2}  
& \scalebox{0.9}{51.3} & \scalebox{0.9}{48.4}  
& \scalebox{0.9}{50.3} & \scalebox{0.9}{44.9}  
& \scalebox{0.9}{60.6} & \scalebox{0.9}{56.9}  
& \scalebox{0.9}{61.8} & \scalebox{0.9}{54.4}  
& \scalebox{0.9}{46.2} & \scalebox{0.9}{40.1}  
& \scalebox{0.9}{27.9} & \scalebox{0.9}{30.5}  
\\

\rowcolor{MyRowBlueGray!50} \scalebox{0.8}{StAR} & \scalebox{0.65}{Qwen3-VL 8B}
& \scalebox{0.9}{70.6} & \scalebox{0.9}{63.8}  
& \scalebox{0.9}{73.8} & \scalebox{0.9}{69.2}  
& \scalebox{0.9}{60.8} & \scalebox{0.9}{57.2}  
& \scalebox{0.9}{57.9} & \scalebox{0.9}{50.1}  
& \scalebox{0.9}{69.1} & \scalebox{0.9}{55.9}  
& \scalebox{0.9}{66.2} & \scalebox{0.9}{60.3}  
& \scalebox{0.9}{54.4} & \scalebox{0.9}{45.0}  
& \scalebox{0.9}{37.0} & \scalebox{0.9}{41.4}  
\\

\rowcolor{MyRowBlueGray!70} \scalebox{0.8}{StAR $+$ MV} & \scalebox{0.65}{Qwen3-VL 8B}
& \scalebox{0.9}{71.8} & \scalebox{0.9}{66.5}  
& \scalebox{0.9}{74.9} & \scalebox{0.9}{69.7}  
& \scalebox{0.9}{62.6} & \scalebox{0.9}{\textbf{60.1}}  
& \scalebox{0.9}{59.6} & \scalebox{0.9}{53.1}  
& \scalebox{0.9}{69.0} & \scalebox{0.9}{58.6}  
& \scalebox{0.9}{68.5} & \scalebox{0.9}{66.4}  
& \scalebox{0.9}{56.7} & \scalebox{0.9}{47.9}  
& \scalebox{0.9}{39.5} & \scalebox{0.9}{41.7}  
\\

\rowcolor{MyRowBlueGray!50} \scalebox{0.8}{StAR} & \scalebox{0.65}{Qwen3-VL 32B}
& \scalebox{0.9}{72.1} & \scalebox{0.9}{67.3}  
& \scalebox{0.9}{73.8} & \scalebox{0.9}{68.0}  
& \scalebox{0.9}{61.5} & \scalebox{0.9}{54.2}  
& \scalebox{0.9}{61.7} & \scalebox{0.9}{59.2}  
& \scalebox{0.9}{71.7} & \scalebox{0.9}{\textbf{71.6}}  
& \scalebox{0.9}{66.7} & \scalebox{0.9}{66.9}  
& \scalebox{0.9}{58.2} & \scalebox{0.9}{50.8}  
& \scalebox{0.9}{47.2} &\scalebox{0.9}{\textbf{51.5}}  
\\

\rowcolor{MyRowBlueGray!70} \scalebox{0.8}{StAR $+$ MV} & \scalebox{0.65}{Qwen3-VL 32B}
&\scalebox{0.9}{\textbf{72.7}} &\scalebox{0.9}{\textbf{68.1}}  
&\scalebox{0.9}{\textbf{75.0}} &\scalebox{0.9}{\textbf{70.6}}  
&\scalebox{0.9}{\textbf{64.7}} &\scalebox{0.9}{57.5}  
&\scalebox{0.9}{\textbf{64.1}} &\scalebox{0.9}{\textbf{60.8}}  
&\scalebox{0.9}{\textbf{74.0}} &\scalebox{0.9}{69.6}  
&\scalebox{0.9}{\textbf{70.2}} &\scalebox{0.9}{\textbf{71.0}}  
&\scalebox{0.9}{\textbf{60.7}} &\scalebox{0.9}{\textbf{54.2}}  
&\scalebox{0.9}{\textbf{47.9}} &\scalebox{0.9}{50.3}  
\\

\bottomrule
\end{tabularx}
\label{tab:reasonseg_results}
\vspace{-4mm}
\end{table*}

\begin{table*}[t]
\tiny
\centering
\setlength{\tabcolsep}{3pt}
\caption{
Comparison on the Multi-target and Multi-granularity Reasoning (MMR) dataset.
* indicates models trained on a training set that includes the MMR dataset.
}
\vspace{-2mm}
\begin{tabularx}{0.9\linewidth}{
  l
  |c
  |*{2}{>{\centering\arraybackslash}X}
  |*{2}{>{\centering\arraybackslash}X}
  G*{2}{>{\centering\arraybackslash}X}
  G*{2}{>{\centering\arraybackslash}X}
}
\toprule
\multirow{3}{*}{\textbf{Method}}
& \multirow{3}{*}{\makecell{\textbf{Base model}\\\textbf{(Size)}}}
& \multicolumn{2}{c|}{\textbf{val}}
& \multicolumn{6}{c}{\textbf{test}} \\
\cmidrule(lr){3-4}\cmidrule(lr){5-10}
&& \multicolumn{2}{c|}{\textbf{Obj \& Part}}
& \multicolumn{2}{c|}{\textbf{Obj \& Part}}
& \multicolumn{2}{c|}{\textbf{Obj}}
& \multicolumn{2}{c}{\textbf{Part}} \\
\cmidrule(lr){3-4}\cmidrule(lr){5-6}\cmidrule(lr){7-8}\cmidrule(lr){9-10}
&&gIoU &cIoU
&gIoU &cIoU
& gIoU &cIoU
&gIoU &cIoU \\
\midrule

LISA~\cite{lai2024lisa} & Llama2 13B
&15.4 &20.0  
&16.1 &19.8  
&26.1 &27.9  
&7.4 &8.4  
\\

LISA$^{*}$~\cite{lai2024lisa} & Llama2 13B
&22.3 &33.4  
&23.0 &29.2  
&40.2 &45.2  
&10.7 &16.4  
\\

M$^2$SA$^{*}$~\cite{jang2025mmr} & LLaVA 7B
&27.8 &48.6  
&30.9 &46.8  
&41.0 &55.6  
&13.5 &27.0  
\\

M$^2$SA$^{*}$~\cite{jang2025mmr} & Llama2 13B
&28.4 &\textbf{49.1}  
&31.6 &\textbf{47.6}  
&42.3 &\textbf{57.6}  
&13.6 &\textbf{27.2}  
\\

VisionReasoner~\cite{visionreasoner} & Qwen2.5-VL 7B
&26.7 &21.4  
&28.4 &21.7  
&37.2 &26.5  
&11.5 &8.7  
\\

\midrule

\rowcolor{MyRowBlueGray!50} StAR & Qwen2.5-VL 7B
&29.8 &26.3  
&32.4 &27.2  
&44.4 &33.3  
&14.3 &11.7  
\\

\rowcolor{MyRowBlueGray!70} StAR $+$ MV & Qwen2.5-VL 7B
&30.6 &27.1  
&33.0 &27.8  
&45.6 &34.7  
&14.9 &12.0  
\\

\rowcolor{MyRowBlueGray!50} StAR & Qwen3-VL 8B
&31.5 &27.0  
&33.9 &26.8  
&45.0 &33.6  
&15.3 &11.8  
\\

\rowcolor{MyRowBlueGray!70} StAR $+$ MV & Qwen3-VL 8B
&\textbf{32.4} &27.6  
&\textbf{34.6} &27.7  
&\textbf{46.2} &34.0  
&\textbf{15.8} &11.9  
\\

\bottomrule
\end{tabularx}
\label{tab:mmr_results}
\vspace{-2mm}
\end{table*}

\section{Experiments}
\vspace{-1mm}
\subsection{Experimental Setup}
\textbf{Datasets.}
We follow VisionReasoner~\cite{visionreasoner} for stage 1 training, using 5k explicit/referring expression segmentation (RES)~\cite{gupta2019lvis, yu2016refcocog, liu2023gref} samples, and use our proposed \emph{ReasonSeg-X} \texttt{train} set (240 samples) for stage 2.
Extending the widely used ReasonSeg~\cite{lai2024lisa}, we introduce \emph{ReasonSeg-R/X} to evaluate model performance from multiple perspectives and across varying difficulty levels.
We further adopt the MMR~\cite{jang2025mmr} dataset to assess multi-target and multi-granularity reasoning segmentation capabilities.
For out-of-domain evaluation, results on the RES datasets~\cite{yu2016refcocog} and MUSE~\cite{ren2024pixellm} are presented in the supplementary material.

\noindent\textbf{Implementation details.}
We use Qwen2.5-VL 7B~\cite{bai2025qwen2_5vl} and Qwen3-VL 8/32B~\cite{Qwen3-VL} as our base models, with SAM 2 Large as the mask generator.
We set the batch size to 16 and the learning rate to $1\times10^{-5}$ ($5\times10^{-6}$ for stage 2).
We use LoRA rank 64 for all base models, and apply REST only during stage 2 finetuning.
By default, the sampling number of reasoning paths in our majority voting is set to 32.
Detailed training settings are provided in the supplementary material.

\vspace{-2mm}
\subsection{Comparison with State-of-the-Arts}
Following previous works~\cite{lai2024lisa, visionreasoner}, we adopt gIoU and cIoU as our evaluation metrics: gIoU is the average of per-sample IoUs, while cIoU is computed as the cumulative intersection over the cumulative union.

\vspace{2mm}
\begin{table}[t]
\centering
\caption{Robotics grounding performance on RefSpatial-Bench~\cite{zhou2026roborefer}.}
\vspace{-2mm}
\label{tab:referspatial-bench}
\small
{
\setlength{\tabcolsep}{4pt}
\renewcommand{\arraystretch}{0.9}
\begin{tabular}{@{}lcc@{}}
\toprule
\multicolumn{1}{@{}c}{\textbf{Success rate (\%)}} &
\multicolumn{2}{c@{}}{\textbf{RefSpatial-Bench}} \\
\cmidrule(l){2-3}
\textbf{Method} & \textbf{Location} & \textbf{Unseen} \\
\midrule
Gemini-2.5-Pro & 46.96 & 27.14 \\
RoboRefer-8B~\cite{zhou2026roborefer} & 52.00 & 37.66 \\
StAR-8B & \textbf{62.00} & \textbf{38.96} \\
\bottomrule
\end{tabular}
}
\vspace{-2mm}
\end{table}

\begin{figure}[t]
  \centering
  \includegraphics[height=7.6cm, width=12.3cm]{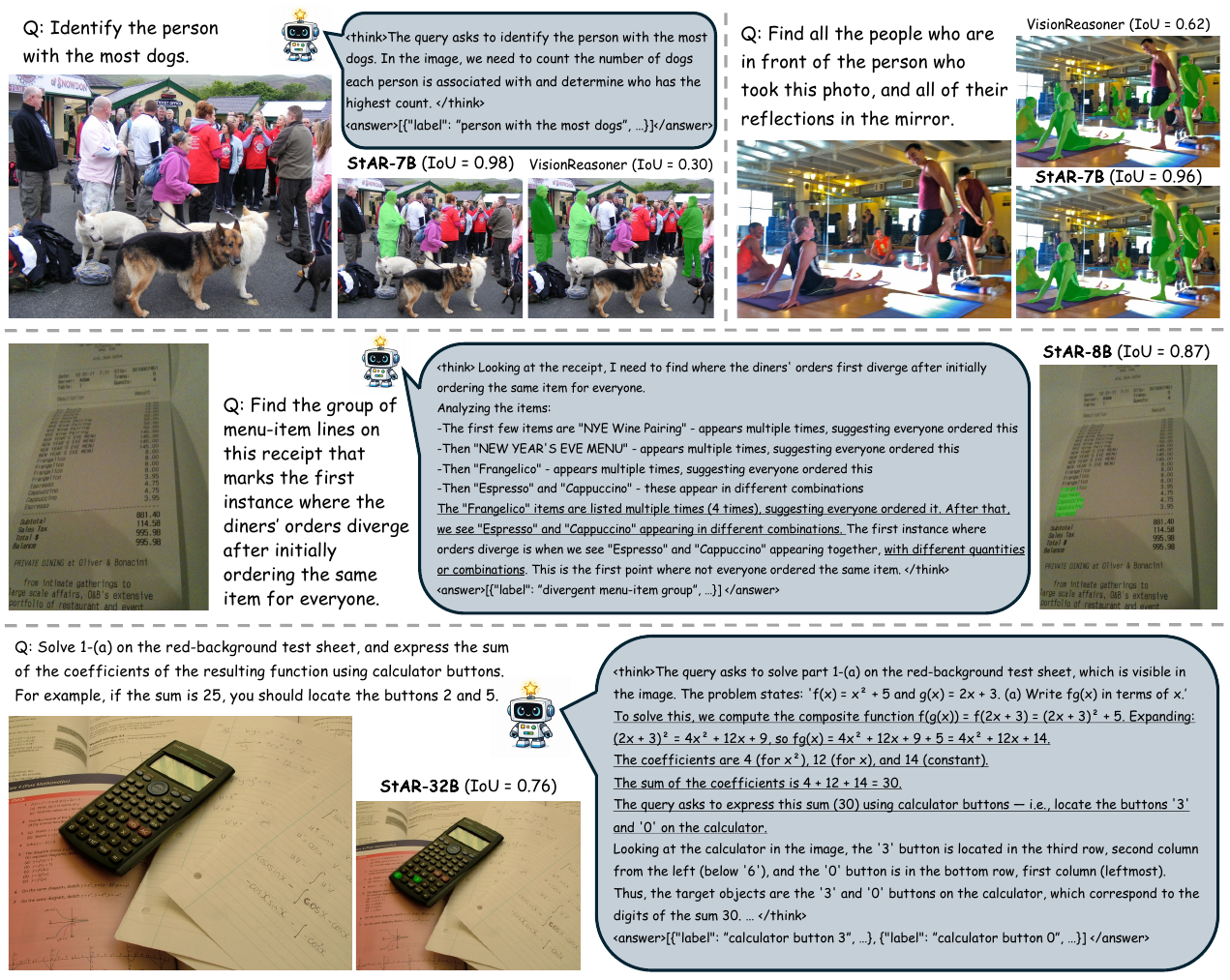}
  \caption{
  Qualitative results on our ReasonSeg-X \texttt{test} set demonstrate that StAR models, via multi-step reasoning, can effectively tackle complex segmentation scenarios necessitating various capabilities, such as comparative/relational reasoning, physics-based 3D awareness, comprehensive pattern analysis, and multi-hop algebraic expansion.
  }
  \label{fig:main_x_qual}
\end{figure}

\noindent\textbf{Results on the ReasonSeg(-R/X) datasets.}
In Tab.~\ref{tab:reasonseg_results}, we compare StAR against previous SOTA baselines.
Notably, our Stage-1 model shows that it can leverage inherent reasoning capabilities \textbf{without using reasoning data}, surpassing methods using the same base model by a large margin.
Finetuning on our dataset with REST further strengthens performance especially on complex reasoning.
Remarkably, StAR-7B with our majority voting attains overall comparable performance on the ReasonSeg-X \texttt{test} set to the SAM 3 Agent (72B), suggesting that \emph{parallel} test-time scaling is also a promising direction.
Moreover, by fully exploiting the base model’s reasoning pattern space, our approach benefits increasingly from larger base model size, and our test-time scaling strategy improves performance in nearly all scenarios.
Fig.~\ref{fig:main_x_qual} illustrates that StAR models are capable of intricate reasoning through step-by-step thinking interleaved with visual inspection.
Meanwhile, it is worth noting that while the performance gap between StAR-8B and StAR-32B is not evident on ReasonSeg-R, it becomes pronounced on ReasonSeg-X, which involves more complex reasoning (\emph{e.g.}, C/R and C/MH).
\emph{This indicates that our benchmark presents a necessary direction for the advancement of future methods.}

\noindent\textbf{Results on the MMR dataset.}
In Tab.~\ref{tab:mmr_results}, we evaluate our models in multi-target and multi-granularity setting~\cite{jang2025mmr}.
StAR models demonstrate strong \textbf{zero-shot} performance, outperforming VisionReasoner as well as models~\cite{jang2025mmr} trained on the MMR dataset (primarily on gIoU).
These results further validate StAR’s flexible segmentation capabilities in addition to its reasoning strengths.

\noindent\textbf{Results on the RefSpatial-Bench dataset.}
In Tab.~\ref{tab:referspatial-bench}, we evaluate StAR’s spatial referring performance, which is crucial for embodied robots to interact with the 3D physical world.
RefSpatial-Bench~\cite{zhou2026roborefer} is a challenging robotics grounding benchmark for evaluating spatial referring with multi-step reasoning.
StAR shows strong \textbf{zero-shot} performance: it is competitive not only on examples requiring target-object localization (location set), but also on more challenging examples involving complex spatial relations, multiple anchors, hierarchical references, or implied placements (unseen set), thereby demonstrating its generalized robotics grounding capability.

Model efficiency comparisons and additional qualitative results can be found in the supplementary material.

\subsection{Additional Analysis}
\textbf{Inference-aware finetuning for majority voting.}
Current finetuning algorithms are agnostic to, and thus decoupled from, the test-time strategy.
However, such a decoupling between training and test-time inference can be suboptimal; when majority voting is employed at test time, training should not optimize only the model’s first attempt, but should also consider multiple attempts under a high-sampling regime.
Accordingly, in Tab.~\ref{tab:main_analysis} \emph{left}, we explore how the number of rollout generations in REST influences performance, with particular emphasis on MV performance.
The results show that REST enables the model to surface high-quality reasoning patterns that express its dormant capability to solve complex reasoning problems, thereby improving performance—especially under majority voting.
The more pronounced improvements observed in MV with increasing rollout generations can be attributed to closer alignment between the sampling distributions used during training and test time.
Moreover, these gains increase synergistically with both the exploration degree and the base model’s reasoning potential (\emph{i.e.}, model size, pre-trained capacity).

\begin{table*}[t]
\centering
\setlength{\tabcolsep}{4pt}
\renewcommand{\arraystretch}{1.12}

\caption{
\emph{Left}: We evaluate performance under varying degrees of REST \textbf{exploration} (the number of rollout generations $n$), focusing on the efficacy of the \emph{inference-time strategy}; the per-step training time for StAR-7B is reported below.
\emph{Right}: We find that reversing the prediction order at test time—after training the model to predict the label before bbox/point coordinates—\textbf{causes a large performance degradation.}
Default settings are marked in \colorbox{MyRowBlueGray!50}{blue}.
}
\label{tab:main_analysis}

\begin{subtable}[t]{0.4\textwidth}
\centering
\tiny
\begin{tabular}{lcccc}
\toprule
\multirow{2}{*}{\makecell{\textbf{RS-X test}\\\textbf{gIoU (\%)}}} & \multicolumn{4}{c}{\textbf{REST} $n$} \\
\cmidrule(lr){2-5}
 & 16 & 64 & 128 &\cellcolor{MyRowBlueGray!50}256 \\
\midrule
StAR-7B     & 48.4 & 49.2 & \textbf{49.4} &49.2 \\
$+$ MV  & 49.3 & 49.5 & 49.8 &\textbf{50.3} \\
StAR-8B     & 57.1 & 56.9 & 57.3 &\textbf{57.9} \\
$+$ MV  & 58.1 & 58.3 & 59.0 &\textbf{59.6} \\
\midrule
Sec/iter & \textbf{170} & 214 & 260 &364 \\
\bottomrule
\end{tabular}
\end{subtable}
\begin{subtable}[t]{0.55\textwidth}
\centering
\tiny
\begin{tabular}{ll|cc|cc}
\toprule
\multirow{2}{*}{\textbf{Model}} &
\multirow{2}{*}{\makecell{\textbf{\scalebox{0.85}{Answer Format}}\\\textbf{\scalebox{0.75}{Prediction Order}}}} &
\multicolumn{2}{c|}{\textbf{RS-R}} &
\multicolumn{2}{c}{\textbf{RS-X test}} \\
\cmidrule(lr){3-4}\cmidrule(lr){5-6}
&& gIoU & cIoU & gIoU & cIoU \\
\midrule
\multirow{2}{*}{StAR-7B} &\cellcolor{MyRowBlueGray!50}label first & \textbf{69.7} & \textbf{66.2} & \textbf{49.2} & \textbf{43.6} \\
& label last  & 68.2 & 63.0 & 46.3 & 40.1 \\
\midrule
\multirow{2}{*}{StAR-8B} &\cellcolor{MyRowBlueGray!50}label first & \textbf{73.8} & 69.2 & \textbf{57.9} & \textbf{50.1} \\
& label last  & 73.1 & \textbf{70.8} & 55.8 & 47.7 \\
\bottomrule
\end{tabular}

\end{subtable}
\vspace{-2mm}
\end{table*}

\noindent \textbf{Answer prediction order.}
As shown in Tab.~\ref{tab:main_analysis} \emph{right}, we note that label prediction per se is not the crucial factor.
Instead, the benefit comes from reformulating localization as a \emph{semantically-conditioned} coordinate prediction step, in which the initially predicted label text serves as an anchor for subsequent grounding.
Conversely, we also observe that for a model trained to predict the label last, prompting label-first prediction at test time substantially improves performance.
\section{Conclusion and Discussion}
In this work, we present StAR, a comprehensive framework that effectively surfaces the base model’s latent reasoning capabilities for segmentation tasks.
We further, for the first time, explore parallel test-time scaling in the reasoning segmentation task and tightly couple training methodologies with inference-time compute strategies.
As a result, each of our design choices provides clear incremental improvements, while their combination yields synergistic benefits, allowing the final model to achieve unprecedented performance across a broad range of reasoning segmentation benchmarks.
Finally, we construct ReasonSeg-X, a highly challenging evaluation benchmark, establishing a clear and systematic target for future works.

\vspace{-2mm}
\subsubsection{Integrating StAR with agentic systems.}
Prior work suggests that for hard questions, appropriately balancing the ratio of sequential to parallel compute is crucial~\cite{snell2025ttsoptimally}.
In this spirit, using StAR as a region-proposal module within a verification-centric sequential pipeline—such as SAM 3 Agent—could be an important next step toward solving harder problems.
Concretely, one can envision an agent that toggles the StAR adapter to propose candidate target regions for a given query, after which the base model verifies the proposed regions.

This integration is intriguing from several perspectives:
(i) It can effectively exploit the verification-generation gap~\cite{setlur2025vggap}, where models are often better at verifying answers than generating them. Verifying StAR-proposed regions using the base model’s generalization capability may particularly improve performance on difficult cases.
(ii) Majority voting can be applied at the region-proposal stage to obtain more diverse proposals, naturally combining parallel scaling (proposal diversity) with sequential scaling (verification).
(iii) Leveraging the StAR adapter for proposals can significantly reduce search cost compared to using SAM 3 alone, improving reliability. Moreover, toggling the StAR adapter on/off across stages enables combining StAR’s strong reasoning capability with the base model’s generalizable verification strength without large memory overhead.

\vspace{-3mm}
\subsubsection{Scope of selective tuning.}
While our current “selective tuning” is applied at the rollout level, it can be extended to improve the learning signal in a more fine-grained and selective manner.
For example, instead of using all tokens in a selected rollout for policy updates, one could prioritize tokens that are more central to reasoning—\emph{e.g.,} those with higher entropy~\cite{wang2025beyond}.
Alternatively, within negative-advantage trajectories, gradients could be masked for partially correct reasoning segments to reduce training noise~\cite{deng2025on}.
In reasoning segmentation, this direction is particularly natural because the task can be decomposed into a reasoning stage and a geometric prediction stage, which induces two fundamental failure modes: reasoning failure and localization failure.
As a result, even when reasoning is successful but coordinate prediction fails, the trajectory may still be penalized in the reasoning portion.
Separating these stages and applying token-level selective tuning—potentially further partitioning the reasoning process into finer sub-parts—could therefore provide a promising avenue to improve the training signal.

\vspace{-3mm}
\subsubsection{Efficiency.}
By scaling only amortizable components, REST can explore a large reasoning pattern space efficiently during training.
However, it can still incur a non-trivial increase in overall training time. An important next step is to devise techniques that anticipate which rollout paths are unlikely to contribute to policy updates, thereby pruning wasted computation and reducing cost.
Moreover, stronger rollout decoding strategies that improve rollout utility in large-sampling regimes could reduce the required rollout budget $n$ while preserving exploration.
We also expect these directions to become even more synergistic when integrated with selective token tuning.

\section{Acknowledgments}
This work was supported by the National Research Foundation of Korea (NRF) grant funded by the Korea government (MSIT) [RS-2025-00562400, RS-2022-NR068754], and also supported by Electronics and Telecommunications Research Institute (ETRI) grant funded by the Korean government [26CS1100, Development of Proprietary Physical AI-based Small-scale Computers and Integrated Soft Suits]

\clearpage


\title{Supplementary Materials for \\
StAR: Segment Anything Reasoner} 

\titlerunning{StAR: Segment Anything Reasoner}

\author{Seokju Yun\inst{1,2} \and
Dongheon Lee\inst{2} \and
Noori Bae\inst{2} \and
Jaesung Jun\inst{2} \and
Chanseul Cho\inst{2} \and
Youngmin Ro\inst{2}\corrauth}

\authorrunning{S.~Yun et al.}

\institute{
KAIST AI
\and
University of Seoul
}

\maketitle

\renewcommand{\thesection}{\Alph{section}}
\renewcommand{\thesubsection}{\thesection.\arabic{subsection}}
\renewcommand{\thesubsubsection}{\thesubsection.\arabic{subsubsection}}

\renewcommand{\thefigure}{S.\arabic{figure}}
\renewcommand{\thetable}{S.\arabic{table}}
\renewcommand{\theequation}{S.\arabic{equation}}


\setcounter{section}{0}
\setcounter{figure}{0}
\setcounter{table}{0}
\setcounter{equation}{0}

\makeatletter
\renewcommand{\theHsection}{supp.\Alph{section}}
\renewcommand{\theHsubsection}{\theHsection.\arabic{subsection}}
\renewcommand{\theHsubsubsection}{\theHsubsection.\arabic{subsubsection}}
\makeatother


\hypersetup{linkbordercolor=black,linkcolor=black}

\par\noindent{\LARGE\bfseries Contents}\par
\vspace{0.6em}

\begingroup
\parindent=0pt
\setlength{\parskip}{0.15em}

\newcommand{\SuppTOCSec}[2]{
  \noindent\hyperref[#1]{\textbf{#2}\dotfill\textbf{\pageref*{#1}}}\par
}

\newcommand{\SuppTOCSub}[2]{
  \noindent\hspace*{1.6em}\hyperref[#1]{#2\dotfill\pageref*{#1}}\par
}

\SuppTOCSec{sec:supp-A}{A.\ Implementation Details}

\SuppTOCSec{sec:supp-B}{B.\ Prompt Design}

\SuppTOCSec{sec:supp-C}{C.\ ReasonSeg-X Annotation Details}

\SuppTOCSec{sec:supp-D}{D.\ Related Work}

\SuppTOCSec{sec:supp-E}{E.\ Model Efficiency Comparison}

\SuppTOCSec{sec:supp-F}{F.\ Additional Experiments}
\SuppTOCSub{sec:supp-F1}{F.1.\ Results on MUSE Dataset}
\SuppTOCSub{sec:supp-F2}{F.2.\ Results on RES Datasets}

\SuppTOCSec{sec:supp-G}{G.\ Additional Methodology Details and Discussion}
\SuppTOCSub{sec:supp-G1}{G.1.\ LoRA Finetuing}
\SuppTOCSub{sec:supp-G2}{G.2.\ Reward Function}
\SuppTOCSub{sec:supp-G3}{G.3.\ Rollout-Expanded Selective-Tuning}
\SuppTOCSub{sec:supp-G4}{G.4.\ Label Prediction}
\SuppTOCSub{sec:supp-G5}{G.5.\ Majority Voting for Segmentation}

\SuppTOCSec{sec:supp-H}{H.\ Additional Qualitative Results}

\SuppTOCSec{sec:supp-I}{I.\ Failure Case Analysis}

\endgroup

\hypersetup{linkcolor=red}

\section{Implementation Details} \label{sec:supp-A}
We train our models using the verl~\cite{sheng2024verl} codebase.
LoRA is applied to all linear layers of the language model, with a default rank of 64.
We disable the KL penalty term and use a weight decay coefficient of 0.001 for all models.
For rollout-expanded selective-tuning (REST), we sample 256 rollouts and perform policy updates on only 16 selected responses; for the StAR-32B model, we reduce the rollout budget to 128 for efficiency.
For all majority voting sampling results, we use temperature 1.0 and top-p 0.9.
We train all models for 1 epoch in Stage 1 and 10 epochs in stage 2.
The end-to-end training of StAR-7B (up to stage 2) requires approximately 33 hours on 8 A6000 GPUs, comparable to VisionReasoner, which requires $\sim$29 hours.

\noindent \underline{\emph{SAM 3 Agent}~\cite{sam3} \emph{evaluation.}}
We first attempt to reproduce the reported results on existing datasets by running the official code.
Given that the original implementation is restricted to sample-level inference, we develop a comprehensive test pipeline incorporating an iterative loop for full dataset evaluation.
During this process, we identify occasional failures in mask production, primarily stemming from the omission of <tool> tags or the exhaustion of the max round.
Therefore, to promote faithful evaluation on our proposed datasets, we allow a single retry for such failure cases.
All other aspects of the evaluation strictly follow the official code.
We also tested Qwen3-VL models, but due to substantially more failures than Qwen2.5-VL and inconsistencies with reported results, we exclude them from performance comparisons.

\begin{table}[t]
\centering
\tcbset{
  colback=gray!5!white,    
  colframe=MyRowBlueGray,           
  width=1.\linewidth,    
  boxrule=1pt,             
  arc=4mm,                 
  left=5pt,                
  right=5pt,               
  top=5pt,                 
  bottom=5pt,              
}
\caption{Prompt template for stage 1 training. ``\textit{\{Question\}}'' is replaced by user query.}
\vspace{-2mm}
\footnotesize 
\begin{tcolorbox}[title=Prompt Template]

"Please find ``\textit{\{Question\}}'' with bbox(es) and point(s)." \\
"Also provide a short label for each object." \\
"\textcolor{blue}{Compare the difference between object(s) and find the most closely matched object(s).}" \\
"Return ALL matching instances; double-check none are missed." \\
"Output the thinking process in <think> </think> and final answer in <answer> </answer> tags." \\
"Output the bbox(es) and point(s) inside the interested object(s), along with a short label, in JSON format." 

\begin{verbatim}
i.e., <think> thinking process (step-by-step reasoning) here </think>
<answer>[{"label": "chair", "bbox_2d": [10,100,200,210],
"point_2d": [30,110]}, {"label": "train track",
"bbox_2d": [225,296,706,786], "point_2d": [302,410]}]</answer>
\end{verbatim}

\normalsize
\label{tab:stage1_template}
\end{tcolorbox}
\vspace{-4mm}
\end{table}

\begin{table}[t]
\centering
\tcbset{
  colback=gray!5!white,    
  colframe=MyRowBlueGray,           
  width=1.\linewidth,    
  boxrule=1pt,             
  arc=4mm,                 
  left=5pt,                
  right=5pt,               
  top=5pt,                 
  bottom=5pt,              
}
\caption{Prompt template for stage 2 training and inference. ``\textit{\{Question\}}'' is replaced by user query.}
\vspace{-2mm}
\footnotesize 
\begin{tcolorbox}[title=Prompt Template]

"Please find ``\textit{\{Question\}}'' with bbox(es) and point(s)." \\
"Also provide a short label for each object." \\
"\textcolor{blue}{First, understand and summarize what the query —``\textit{\{Question\}}''— is likely referring to (which object or concept).}" \\
"\textcolor{blue}{Then apply this to the image and find the matched target object(s).}" \\
"Return ALL matching instances; \textcolor{orange}{if there are no matches, return an empty list (<answer>[]</answer>).} double-check none are missed." \\
"Output the thinking process in <think> </think> and final answer in <answer> </answer> tags." \\
"Output the bbox(es) and point(s) inside the interested object(s), along with a short label, in JSON format." 

\begin{verbatim}
i.e., <think> thinking process (step-by-step reasoning) here </think>
<answer>[{"label": "chair", "bbox_2d": [10,100,200,210],
"point_2d": [30,110]}, {"label": "train track",
"bbox_2d": [225,296,706,786], "point_2d": [302,410]}]</answer>
\end{verbatim}

\normalsize
\label{tab:stage2_test_template}
\end{tcolorbox}
\vspace{-4mm}
\end{table}

\section{Prompt Design} \label{sec:supp-B}
The prompt templates for Stage 1 training and for Stage 2 training/inference are presented in Tab.~\ref{tab:stage1_template} and Tab.~\ref{tab:stage2_test_template}, respectively.
Our prompt design builds on VisionReasoner’s template (including zero-shot CoT prompting), while introducing several targeted refinements.
We first augment the template with instructions for label prediction and with warnings against under-segmentation.
We then introduce two key modifications.
First, inspired by the effectiveness of solving complex problems by breaking them into subproblems and tackling them sequentially~\cite{wang2023plan-and-solve, zhou2023leasttomost}, we explicitly instruct the model to follow a stepwise process—understand the query, apply it to the image, and then locate the target—rather than simply “find the closely matched object”.
This modification can be seen by comparing the \textcolor{blue}{blue}-highlighted texts in Tab.~\ref{tab:stage1_template} and Tab.~\ref{tab:stage2_test_template}.
Second, to address no-target scenarios in test time, we instruct the model to output an empty list when the queried object is absent from the image (marked in \textcolor{orange}{orange} in Tab.~\ref{tab:stage2_test_template}).
In summary, considering task complexity, we utilize the prompt template in Tab.~\ref{tab:stage1_template} for Stage 1 training, and the prompt template in Tab.~\ref{tab:stage2_test_template} for Stage 2 training and inference.

\section{ReasonSeg-X Annotation Details} \label{sec:supp-C}
Since our goal is to deepen the reasoning depth beyond existing datasets, annotation is inherently non-trivial and expected to be time-consuming.
We initially attempted a GPT-assisted semi-automatic pipeline; however, even with meticulously crafted prompts for GPT-5, the generated problems fell short of our requirements in both reasoning-type compliance and reasoning depth criterion.
Consequently, in most cases, we follow a manual annotation protocol: we annotate queries for collected images and generate binary masks using SAM 2 model from our manually specified bbox/point.

Specifically, our annotation procedure is as follows:
\begin{itemize}[leftmargin=3mm, itemsep=0.5mm, topsep=0.5mm, partopsep=0mm]
    \item We closely inspect each selected image, assign a reasoning type, and carefully annotate the query and its corresponding binary mask(s). In this step, we focus on (i) designing queries that require a comprehensive understanding of the scene and (ii) ensuring that the sample satisfies the assigned reasoning type while demanding sufficiently deep reasoning.
    \item To improve the reliability of query–reasoning–mask correspondence, annotators cross-validate samples and filter or revise those that violate criteria such as logical soundness, clarity, and precision.
    \item To further identify human errors, we employ GPT-5 and Gemini 3 to assess reasoning consistency and mask correspondence/quality and incorporate their feedback when finalizing the dataset.
\end{itemize}

\begin{figure}[t]
  \centering
  \includegraphics[width=12.3cm]{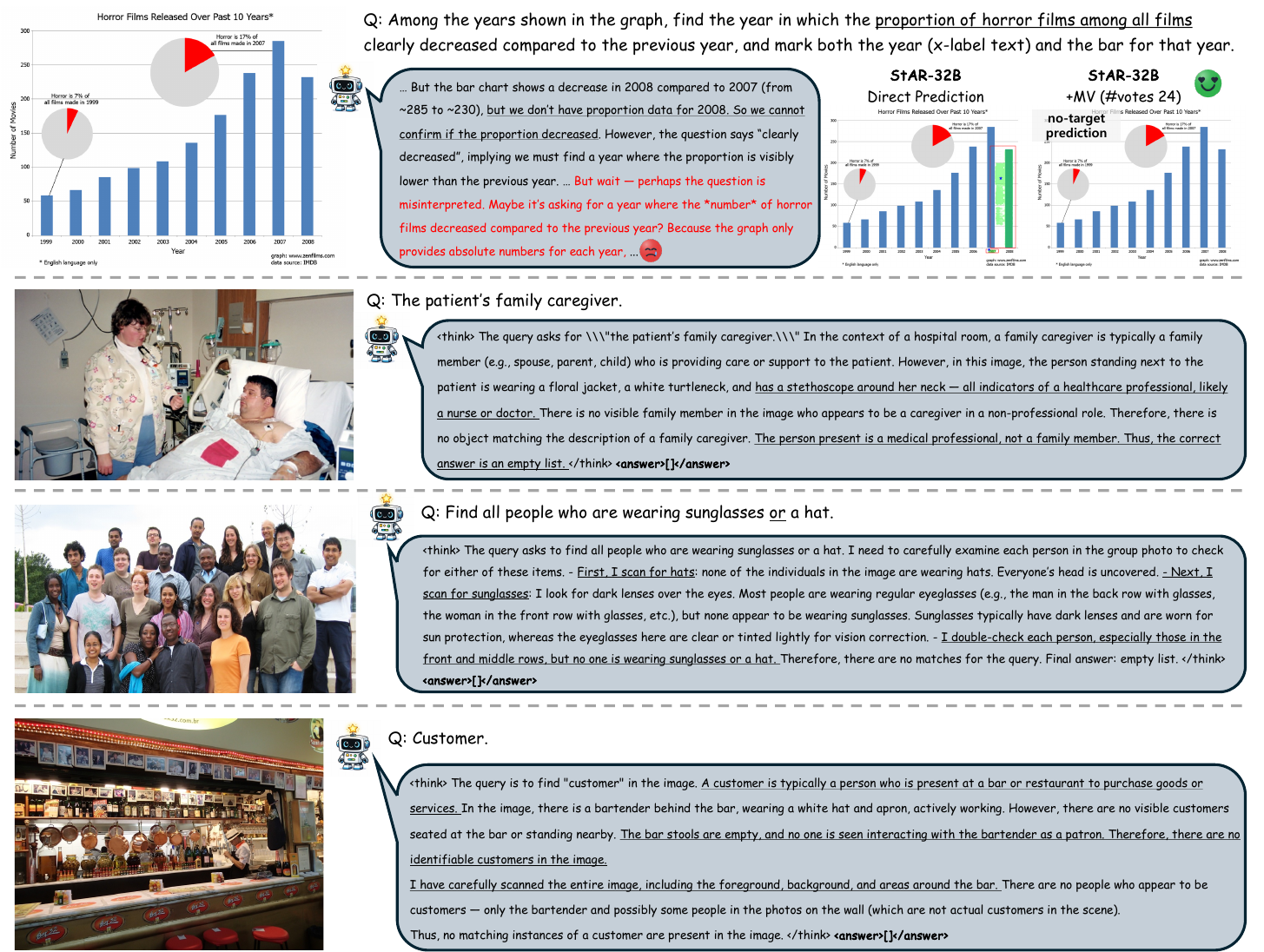}
  \caption{
  Examples of no-target samples in ReasonSeg-X.
  StAR-32B reliably identifies cases where the queried object does not exist in the image.
  The first row further shows that majority voting can mitigate several failure modes of one-shot prediction.
  Notably, the model exhibits diverse reasoning skills, such as fine-grained local inspection, sequential search over multiple constraints, and verification of intermediate conclusions.
  }
  \label{fig:no-target_examples}
\vspace{-4mm}
\end{figure}

\begin{figure}[t]
  \centering
  \includegraphics[height=4cm]{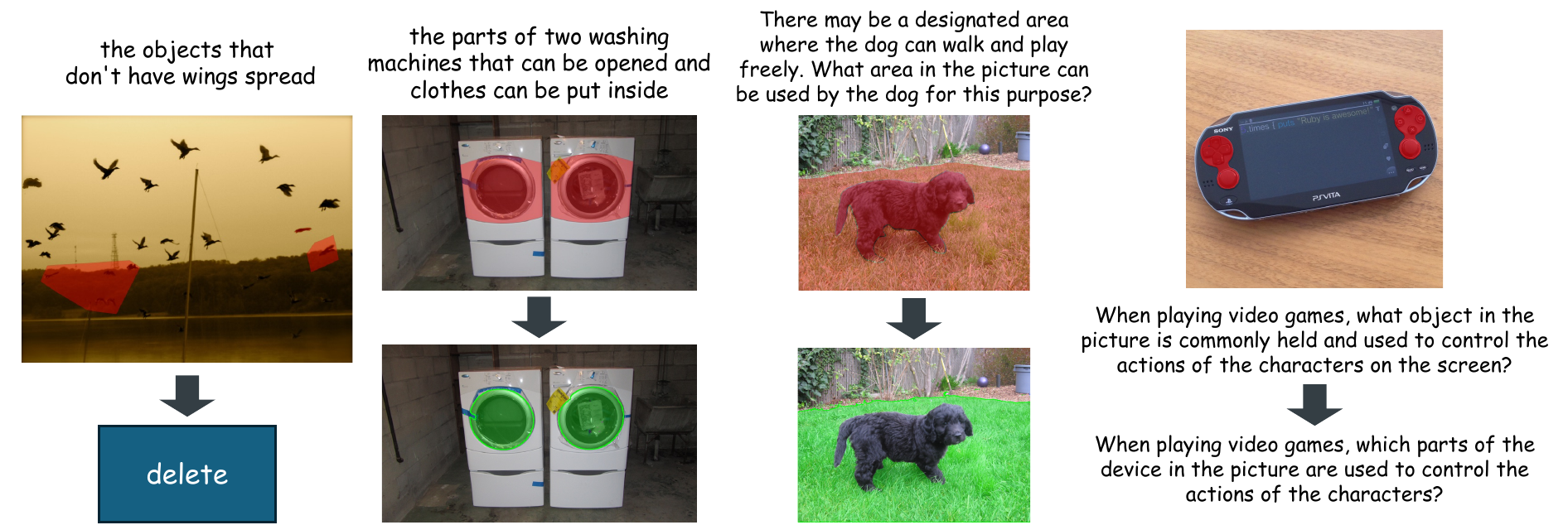}
  \caption{
  We further illustrate typical refinement decisions.
  In the leftmost example,
  we decide to remove the sample because both the mask quality and the query’s required reasoning are visually and semantically ambiguous.
  }
  \label{fig:reasonseg-r_supp_ex}
\vspace{-3mm}
\end{figure}

We strive to balance the reasoning type distribution.
In addition, to support research on realistic reasoning segmentation, we include \emph{no-target} samples where the queried object is absent from the image; examples are shown in Fig.~\ref{fig:no-target_examples}.
Also, additional examples of the ReasonSeg-R refinement process are provided in Fig.~\ref{fig:reasonseg-r_supp_ex}.
The resulting dataset enables more faithful evaluation of a model’s reasoning segmentation capability.

\vspace{-2mm}
\section{Related Work} \label{sec:supp-D}
\vspace{-1mm}
\subsubsection{Reasoning Segmentation.}
LISA~\cite{lai2024lisa} is the first work to introduce the reasoning segmentation task—requiring binary mask generation via complex reasoning—together with the ReasonSeg benchmark dataset.
It further proposes a new <seg> token and an embedding-as-mask paradigm, where the token embedding is provided to SAM~\cite{kirillov2023sam} to produce the final segmentation mask.
Subsequent studies~\cite{qian2025reasonattend, bao2024cores, jang2025mmr, zhu2025popen, wang2025segllm, qian2026anchorseg} further develop this paradigm and demonstrate improved results.
Meanwhile, inspired by the effectiveness of Reinforcement Learning with Verifiable Rewards (RLVR)~\cite{guo2025deepseekr1} in enhancing the reasoning performance of LLMs, Seg-Zero~\cite{liu2025seg_zero} and VisionReasoner~\cite{visionreasoner} design segmentation-specific reward functions and apply RLVR (\emph{e.g.,} GRPO~\cite{shao2024deepseekmath}) to MLLMs, yielding large performance improvements.
In particular, by departing from rigid output formats and adopting coordinate prediction, these methods open a new direction that fully exploits MLLMs’ natural language generation strengths while remaining seamlessly compatible with the standard GRPO pipeline.
Building on this, several GRPO-based methods have been introduced: SAM-R1~\cite{huang2025samr1} incorporates SAM into the training loop and employs a tiered mask IoU reward; CoPRS~\cite{lu2025coprs} generates a positional prior instantiated as a heatmap, from which masks are decoded; and DPAD~\cite{DPAD} prompts the MLLM to produce a descriptive caption and leverages it to induce focused perception by contrasting the target against its surrounding context.
Concurrently, DGSeg~\cite{zeng2026dgseg} employs an RL-tuned MLLM to generate complementary semantic and spatial prompts for SAM3~\cite{sam3}, and adaptively fuses the resulting dual-branch predictions via dynamic gating.

Our work differs from the aforementioned methods by fundamentally alleviating reasoning bottlenecks in the core components of RLVR.
Moreover, beyond SAM-R1, we leverage a compact reward design that combines a more fine-grained mask reward with MLLM-level reward functions, and we further investigate rollout scalability.
DPAD’s descriptive captioning shares some resemblance to our proposed label prediction; however, DPAD is not easily reconciled with multi-target scenarios and can complicate the training pipeline by requiring additional models such as CLIP~\cite{radfordclip}.
By contrast, label prediction is tightly coupled with per-object geometry prediction, encouraging the model to disentangle objects and implicitly fostering contrastive perception during accuracy-reward optimization (see Fig.~\ref{fig:LP_examples}).
Finally, we propose \emph{ReasonSeg-X}, a challenging benchmark for systematically evaluating reasoning segmentation capabilities.

\vspace{-3mm}
\subsubsection{Inference-time Scaling.}
Language models have shown that reasoning performance can be improved progressively by allocating additional inference-time compute~\cite{snell2025ttsoptimally}.
Broadly, existing test-time scaling strategies fall along two axes.
The first comprises \emph{sequential} scaling approaches, including not only within-turn expansion such as CoT~\cite{wei2022cot}, but also iterative refinement/correction~\cite{qu2024recursive, kumar2025trainingselfcorrect, madaan2023selfrefine} of a model’s own outputs.
Although effective, sequential scaling can induce overthinking~\cite{ghosal2025doesthinkingmore}, and in multimodal reasoning settings, concise reasoning traces often outperform verbose ones~\cite{xu2025morethinking, tian2026morethought}.
This motivates a complementary axis—\emph{parallel} compute scaling.
The simplest and most lightweight approach is majority voting~\cite{wang2023mv}, which selects the most frequent answer among multiple outputs.
Majority voting is typically confined to tasks with discrete answer spaces where voting is well-defined (\emph{e.g.,} mathematical problem-solving and question answering).
Although very recent studies~\cite{kang2025scalablebestofn, wang2026thinkparallel} devise ways to apply this to open-ended tasks like code generation and deep research, applying them to pixel-level recognition remains largely unexplored.
We therefore propose a flexible strategy that adapts majority voting to multi-target reasoning segmentation.
In contrast to existing sequential scaling methods in the reasoning segmentation task that rely on refining or verifying intermediate results~\cite{anonymous2026samveteran, lu-etal-2025-rsvp, sam3}, our parallel scaling approach offers a more potent means of enhancing performance.

\begin{table}[t]
\centering
\setlength{\tabcolsep}{2.7pt}
\caption{Model inference efficiency results.
“MV” denotes our majority voting strategy.
Results for SAM 3 Agent and majority voting are measured with batch size 1, while all other results are measured with batch size 32.}
\label{tab: model_efficiency}
\vspace{-2mm}
\begin{tabular}{l|cccc|cccc}
\toprule
\multirow{2}{*}{\textbf{Method}} 
& \multicolumn{4}{c|}{\textbf{ReasonSeg-R}} 
& \multicolumn{4}{c}{\textbf{ReasonSeg-X test}} \\
\cmidrule(lr){2-5} \cmidrule(lr){6-9}
& \scalebox{0.85}{\textbf{gIoU}} & \scalebox{0.85}{\textbf{cIoU}} & \scalebox{0.85}{\textbf{\#Tokens}} & \scalebox{0.85}{\textbf{Latency}}
& \scalebox{0.85}{\textbf{gIoU}} & \scalebox{0.85}{\textbf{cIoU}} & \scalebox{0.85}{\textbf{\#Tokens}} & \scalebox{0.85}{\textbf{Latency}} \\
\midrule
\rowcolor{gray!10}\multicolumn{9}{l}{\textit{\textcolor{gray}{\scalebox{0.8}{Measured on an NVIDIA RTX A6000}}}} \\
VisionReasoner~\cite{visionreasoner} & \textcolor{gray}{64.8} & \textcolor{gray}{56.8} & 160 & 4.0s & \textcolor{gray}{42.2} & \textcolor{gray}{33.8} & 200 & 5.2s \\
StAR-7B & \textcolor{gray}{69.7} & \textcolor{gray}{66.2} & 138 & 3.5s & \textcolor{gray}{49.2} & \textcolor{gray}{43.6} & 147 & 3.7s \\
$+$ MV & \textcolor{gray}{70.7} & \textcolor{gray}{67.2} & 139$\times$32 & 32.3s & \textcolor{gray}{50.3} & \textcolor{gray}{44.9} & 149$\times$32 & 33.7s \\
StAR-8B & \textcolor{gray}{73.8} & \textcolor{gray}{69.2} & 203 & 2.9s & \textcolor{gray}{57.9} & \textcolor{gray}{50.1} & 267 & 3.7s \\
$+$ MV & \textcolor{gray}{74.9} & \textcolor{gray}{69.7} & 205$\times$32 & 33.5s & \textcolor{gray}{59.6} & \textcolor{gray}{53.1} & 
265$\times$32 & 40.8s \\
\midrule
\rowcolor{gray!10}\multicolumn{9}{l}{\textit{\textcolor{gray}{\scalebox{0.8}{Measured on an NVIDIA H200}}}} \\
SAM 3 Agent-7B~\cite{sam3} & \textcolor{gray}{63.1} & \textcolor{gray}{58.0} & -- & 21.8s & \textcolor{gray}{34.4} & \textcolor{gray}{29.5} & -- & 30.3s \\
SAM 3 Agent-72B~\cite{sam3} & \textcolor{gray}{72.4} & \textcolor{gray}{65.3} & -- & 210.1s & \textcolor{gray}{49.8} & \textcolor{gray}{40.3} & -- & 423.8s \\
StAR-32B & \textcolor{gray}{73.8} & \textcolor{gray}{68.0} & 268 & 2.8s & \textcolor{gray}{61.7} & \textcolor{gray}{59.2} & 337 & 3.4s \\
$+$ MV & \textcolor{gray}{75.0} & \textcolor{gray}{70.6} & 280$\times$32 & 43.1s & \textcolor{gray}{64.1} & \textcolor{gray}{60.8} & 
355$\times$32 & 49.5s \\
\bottomrule
\end{tabular}
\vspace{-4mm}
\end{table}

\vspace{-2mm}
\section{Model Efficiency Comparison} \label{sec:supp-E}
In Tab.~\ref{tab: model_efficiency}, we compare inference efficiency on ReasonSeg-R/X with recent baselines.
By analyzing inference token counts along the roadmap in Fig.~\ref{fig:roadmap_star} (a), we assume that LoRA~\cite{hu2022lora} helps StAR internalize the underlying reasoning patterns behind correct responses—eschewing simple memorization of verbose outputs—while our train-time exploration strategy (REST) increases exposure to concise, well-structured reasoning trajectories.
As a result, StAR-7B reduces inference latency by 28.9\% and cuts token usage by 26.5\% compared to the VisionReasoner-7B.
Furthermore, when paired with performant base models such as Qwen3-VL~\cite{Qwen3-VL}, our approach exhibits flexible adaptivity by dynamically modulating reasoning depth according to task complexity.
Also, our \emph{parallel} compute scaling method provides an efficient alternative to multi-round tool-calling agents~\cite{sam3}.

\begin{table*}[t]
\caption{\emph{Left:} Performance comparison on the MUSE~\cite{ren2024pixellm} dataset.
In \cite{ren2024pixellm}, evaluation includes scores obtained by prompting GPT-3.5, which can hinder consistent and reproducible measurement in follow-up studies.
We instead simply adopt a standard IoU calculation scheme to improve evaluation accessibility and focus on comparison with VisionReasoner~\cite{visionreasoner}.
\emph{Right:} Performance comparison on Referring Expression Segmentation (RES) benchmarks measured by cIoU.
* denotes our reproduced results with the official code and model checkpoint.
}
\label{tab:muse_res}
\centering

\begin{subtable}[t]{0.3\textwidth}
\centering
{
\setlength{\tabcolsep}{4pt}
\begin{tabular}{l|cc}
\toprule
\multirow{2}{*}{\scalebox{0.85}{\textbf{Method}}} &
\multicolumn{2}{c}{\scalebox{0.8}{\textbf{MUSE val}}} \\
\cmidrule(lr){2-3}
& \scalebox{0.8}{\textbf{gIoU}} & \scalebox{0.8}{\textbf{cIoU}} \\
\midrule
\scalebox{0.8}{LISA-7B} & \textcolor{gray}{42.0} & \textcolor{gray}{46.1} \\
\scalebox{0.8}{PixelLM-7B} & \textcolor{gray}{42.6} & \textcolor{gray}{50.7} \\
\scalebox{0.8}{POPEN-7B} & \textcolor{gray}{45.4} & \textcolor{gray}{55.2} \\
\scalebox{0.8}{LISA-13B} & \textcolor{gray}{43.6} & \textcolor{gray}{50.2} \\
\scalebox{0.8}{PixelLM-13B} & \textcolor{gray}{44.8} & \textcolor{gray}{54.1} \\
\scalebox{0.8}{POPEN-13B} & \textcolor{gray}{48.0} & \textcolor{gray}{59.1} \\
\midrule
\scalebox{0.8}{VisionReasoner} & 50.5 & 47.6 \\
\rowcolor{MyRowBlueGray!50}\scalebox{0.8}{StAR-7B} & 55.2 & 53.0 \\
\rowcolor{MyRowBlueGray!70}\scalebox{0.8}{$+$ MV} & 56.6 & 54.6 \\
\rowcolor{MyRowBlueGray!50}\scalebox{0.8}{StAR-8B} & 57.5 & 55.7 \\
\rowcolor{MyRowBlueGray!70}\scalebox{0.8}{$+$ MV} & 58.2 & 56.7 \\
\bottomrule
\end{tabular}
}
\end{subtable}
\hfill
\begin{subtable}[t]{0.64\textwidth}
\centering
{
\setlength{\tabcolsep}{4pt}
\renewcommand{\arraystretch}{0.9}
\begin{tabular}{l|c|c|c}
\toprule
\multirow{2}{*}{\scalebox{0.8}{\textbf{Method}}} &
\scalebox{0.8}{\textbf{RefCOCO}} &
\scalebox{0.8}{\textbf{RefCOCO$+$}} &
\scalebox{0.8}{\textbf{RefCOCOg}} \\
& \scalebox{0.8}{\textbf{testA}} &
\scalebox{0.8}{\textbf{testA}} &
\scalebox{0.8}{\textbf{test}} \\
\midrule
\scalebox{0.8}{LISA-7B} & 76.5 & 67.4 & 68.5 \\
\scalebox{0.8}{SegLLM} & 81.5 & 73.0 & 73.6 \\
\scalebox{0.8}{READ} & 80.2 & 73.7 & 71.4 \\
\scalebox{0.8}{CoReS} & 78.6 & 70.0 & 70.7 \\
\scalebox{0.8}{CoPRS-7B} & 85.3 & 80.3 & 76.2 \\
\scalebox{0.8}{SAM-R1} & 79.2 & 74.7 & 73.1 \\
\scalebox{0.8}{SAM-Veteran} & 80.8 & 76.6 & 73.4 \\
\scalebox{0.8}{VisionReasoner*} & 76.6 & 72.1 & 67.2 \\
\scalebox{0.8}{SAM 3 Agent-7B} & 64.3 & 57.0 & 58.8 \\
\scalebox{0.8}{SAM 3 Agent-72B} & 74.9 & 70.8 & 70.2 \\
\midrule
\rowcolor{MyRowBlueGray!50}\scalebox{0.8}{StAR-7B} & 78.6 & 74.0 & 71.0 \\
\rowcolor{MyRowBlueGray!70}\scalebox{0.8}{$+$ MV ($N$=16)} & 79.0 & 74.7 & 71.5 \\
\rowcolor{MyRowBlueGray!50}\scalebox{0.8}{StAR-8B} & 78.7 & 74.4 & 73.3 \\
\rowcolor{MyRowBlueGray!70}\scalebox{0.8}{$+$ MV ($N$=16)} & 79.3 & 75.2 & 73.8 \\
\bottomrule
\end{tabular}
}
\end{subtable}
\end{table*}

\section{Additional Experiments} \label{sec:supp-F}

\subsection{Results on MUSE Dataset} \label{sec:supp-F1}
In Tab.~\ref{tab:muse_res} \emph{left}, we benchmark our models—primarily against VisionReasoner—to assess multi-target reasoning segmentation capability~\cite{ren2024pixellm}.
StAR-7B markedly surpasses the VisionReasoner baseline, and its gains are consistently strengthened by our majority voting strategy.
Together, these results indicate that our model can reason in goal-directed settings to flexibly localize multiple targets.

\subsection{Results on RES Datasets} \label{sec:supp-F2}
Evaluation results on the Referring Expression Segmentation (RES)~\cite{yu2016refcocog} datasets are shown in Tab.~\ref{tab:muse_res} \emph{right}.
Despite training on only $\sim$1.8k samples from the RefCOCOg \texttt{train} split, our model achieves competitive performance, particularly compared to VisionReasoner and SAM 3 Agents.
Moreover, when equipped with a more performant base model and majority voting, StAR attains performance comparable to methods that leverage the RefCOCOg \texttt{train} set more aggressively (\emph{e.g.,} SAM-R1~\cite{huang2025samr1}, SAM-Veteran~\cite{anonymous2026samveteran}, and SegLLM~\cite{wang2025segllm}).
These results demonstrate that our models remain robust across varying reasoning depths.

\begin{table*}[t]
\scriptsize
\centering
\setlength{\tabcolsep}{3pt}
\renewcommand{\arraystretch}{1.05}
\caption{Detailed results for retrofitting a VisionReasoner~\cite{visionreasoner}.}
\label{tab: detailed_ablation}
\begin{tabularx}{0.99\linewidth}{
  l
  |*{2}{>{\centering\arraybackslash}X}
  |*{2}{>{\centering\arraybackslash}X}
  G*{2}{>{\centering\arraybackslash}X}
  G*{2}{>{\centering\arraybackslash}X}
  G*{2}{>{\centering\arraybackslash}X}
  G*{2}{>{\centering\arraybackslash}X}
}
\toprule
\multirow{3}{*}{\textbf{Variant}}
& \multicolumn{2}{c|}{\textbf{RS-R}}
& \multicolumn{10}{c}{\textbf{RS-X test}} \\
\cmidrule(lr){2-3}\cmidrule(lr){4-13}
& \multicolumn{2}{c|}{--}
& \multicolumn{2}{c}{overall}
& \multicolumn{2}{c}{P/F}
& \multicolumn{2}{c}{C/KI}
& \multicolumn{2}{c}{C/R}
& \multicolumn{2}{c}{C/MH} \\
\cmidrule(lr){2-3}
\cmidrule(lr){4-5}
\cmidrule(lr){6-7}
\cmidrule(lr){8-9}
\cmidrule(lr){10-11}
\cmidrule(lr){12-13}
&
\scalebox{0.8}{gIoU} & \scalebox{0.8}{cIoU} & \scalebox{0.8}{gIoU} & \scalebox{0.8}{cIoU} & \scalebox{0.8}{gIoU} & \scalebox{0.8}{cIoU} & \scalebox{0.8}{gIoU} & \scalebox{0.8}{cIoU} & \scalebox{0.8}{gIoU} & \scalebox{0.8}{cIoU} & \scalebox{0.8}{gIoU} & \scalebox{0.8}{cIoU} \\
\midrule

\rowcolor{gray!10}
VisionReasoner-7B~\cite{visionreasoner}
& \scalebox{0.8}{64.8} & \scalebox{0.8}{56.8}
& \scalebox{0.8}{42.2} & \scalebox{0.8}{33.8}
& \scalebox{0.8}{50.1} & \scalebox{0.8}{45.1}
& \scalebox{0.8}{50.7} & \scalebox{0.8}{39.0}
& \scalebox{0.8}{39.4} & \scalebox{0.8}{31.1}
& \scalebox{0.8}{24.5} & \scalebox{0.8}{23.8} \\

\textcolor{gray}{FFT $\xrightarrow{}$ LoRA $r$16}
& \scalebox{0.8}{64.0} & \scalebox{0.8}{58.0}
& \scalebox{0.8}{44.4} & \scalebox{0.8}{34.9}
& \scalebox{0.8}{--} & \scalebox{0.8}{--}
& \scalebox{0.8}{--} & \scalebox{0.8}{--}
& \scalebox{0.8}{--} & \scalebox{0.8}{--}
& \scalebox{0.8}{--} & \scalebox{0.8}{--} \\

LoRA $r$16 $\xrightarrow{}$ 64
& \scalebox{0.8}{66.1} & \scalebox{0.8}{58.0}
& \scalebox{0.8}{45.4} & \scalebox{0.8}{37.7}
& \scalebox{0.8}{54.9} & \scalebox{0.8}{49.7}
& \scalebox{0.8}{55.7} & \scalebox{0.8}{46.9}
& \scalebox{0.8}{42.9} & \scalebox{0.8}{35.9}
& \scalebox{0.8}{22.4} & \scalebox{0.8}{21.6} \\

\scalebox{0.95}{\textcolor{gray}{MLLM-level reward, rollout $n$8 $\xrightarrow{}$ 16}}
& \scalebox{0.8}{65.5} & \scalebox{0.8}{55.9}
& \scalebox{0.8}{45.6} & \scalebox{0.8}{37.2}
& \scalebox{0.8}{55.5} & \scalebox{0.8}{44.9}
& \scalebox{0.8}{54.0} & \scalebox{0.8}{42.9}
& \scalebox{0.8}{43.0} & \scalebox{0.8}{35.2}
& \scalebox{0.8}{25.1} & \scalebox{0.8}{25.5} \\

\textcolor{gray}{mask IoU reward, rollout $n$8}
& \scalebox{0.8}{66.2} & \scalebox{0.8}{59.2}
& \scalebox{0.8}{45.9} & \scalebox{0.8}{39.4}
& \scalebox{0.8}{52.9} & \scalebox{0.8}{44.9}
& \scalebox{0.8}{51.0} & \scalebox{0.8}{42.2}
& \scalebox{0.8}{44.6} & \scalebox{0.8}{41.2}
& \scalebox{0.8}{26.1} & \scalebox{0.8}{27.5} \\

mask IoU reward, rollout $n$8 $\xrightarrow{}$ 16
& \scalebox{0.8}{66.6} & \scalebox{0.8}{58.9}
& \scalebox{0.8}{47.1} & \scalebox{0.8}{38.5}
& \scalebox{0.8}{53.2} & \scalebox{0.8}{42.9}
& \scalebox{0.8}{56.1} & \scalebox{0.8}{47.1}
& \scalebox{0.8}{46.8} & \scalebox{0.8}{38.0}
& \scalebox{0.8}{26.7} & \scalebox{0.8}{26.4} \\

\textcolor{gray}{Stage 2 training, vanilla GRPO}
& \scalebox{0.8}{67.8} & \scalebox{0.8}{62.5}
& \scalebox{0.8}{47.6} & \scalebox{0.8}{40.2}
& \scalebox{0.8}{54.6} & \scalebox{0.8}{47.5}
& \scalebox{0.8}{56.6} & \scalebox{0.8}{48.0}
& \scalebox{0.8}{47.5} & \scalebox{0.8}{38.9}
& \scalebox{0.8}{27.2} & \scalebox{0.8}{26.9} \\

\textcolor{gray}{$\raisebox{0.25ex}{$\llcorner$}+$ MV}
& \scalebox{0.8}{68.2} & \scalebox{0.8}{62.2}
& \scalebox{0.8}{48.7} & \scalebox{0.8}{41.1}
& \scalebox{0.8}{57.4} & \scalebox{0.8}{53.1}
& \scalebox{0.8}{59.5} & \scalebox{0.8}{52.0}
& \scalebox{0.8}{46.6} & \scalebox{0.8}{38.9}
& \scalebox{0.8}{27.0} & \scalebox{0.8}{24.2} \\

REST ($n$16 $\xrightarrow{}$ 256, $m$16)
& \scalebox{0.8}{68.2} & \scalebox{0.8}{62.2}
& \scalebox{0.8}{48.6} & \scalebox{0.8}{42.7}
& \scalebox{0.8}{54.9} & \scalebox{0.8}{51.6}
& \scalebox{0.8}{57.3} & \scalebox{0.8}{49.1}
& \scalebox{0.8}{49.2} & \scalebox{0.8}{41.9}
& \scalebox{0.8}{26.8} & \scalebox{0.8}{28.7} \\

$\raisebox{0.25ex}{$\llcorner$}+$ MV
& \scalebox{0.8}{68.6} & \scalebox{0.8}{62.7}
& \scalebox{0.8}{49.8} & \scalebox{0.8}{43.7}
& \scalebox{0.8}{57.9} & \scalebox{0.8}{56.1}
& \scalebox{0.8}{59.2} & \scalebox{0.8}{53.3}
& \scalebox{0.8}{48.8} & \scalebox{0.8}{41.5}
& \scalebox{0.8}{27.6} & \scalebox{0.8}{26.6} \\

\rowcolor{MyRowBlueGray!50}
label prediction (= StAR-7B)
& \scalebox{0.8}{69.7} & \scalebox{0.8}{66.2}
& \scalebox{0.8}{49.2} & \scalebox{0.8}{43.6}
& \scalebox{0.8}{58.2} & \scalebox{0.8}{52.2}
& \scalebox{0.8}{60.2} & \scalebox{0.8}{52.4}
& \scalebox{0.8}{45.7} & \scalebox{0.8}{40.4}
& \scalebox{0.8}{28.0} & \scalebox{0.8}{28.8} \\

\rowcolor{MyRowBlueGray!70}
$\raisebox{0.25ex}{$\llcorner$}+$ MV
& \scalebox{0.8}{70.7} & \scalebox{0.8}{67.2}
& \scalebox{0.8}{50.3} & \scalebox{0.8}{44.9}
& \scalebox{0.8}{60.6} & \scalebox{0.8}{56.9}
& \scalebox{0.8}{61.8} & \scalebox{0.8}{54.4}
& \scalebox{0.8}{46.2} & \scalebox{0.8}{40.1}
& \scalebox{0.8}{27.9} & \scalebox{0.8}{30.5} \\

\bottomrule
\end{tabularx}
\end{table*}

\section{Additional Methodology Details and Discussion} \label{sec:supp-G}
In Tab.~\ref{tab: detailed_ablation}, we first provide detailed tabulated results for the experiments that \emph{retrofit} VisionReasoner, as illustrated in Fig.~\ref{fig:roadmap_star}.
We observe that all design choices progressively improve the overall performance on ReasonSeg-R/X.
However, there is an exception where adding label prediction noticeably degrades performance on the Comparative/Relational reasoning types; we leave a detailed analysis of these cases to future work.

\subsection{LoRA Finetuing} \label{sec:supp-G1}
Previous work~\cite{huang2025hira} demonstrates that higher-rank parameter updates are particularly important for complex reasoning tasks.
Building on this insight, we examine how LoRA~\cite{hu2022lora} rank affects reasoning segmentation performance.
As reported in Tab.~\ref{tab:ablation_rank}, the model trained with LoRA rank 16 already surpasses strong baselines such as VisionReasoner by a large margin, and increasing the rank to 64 yields further gains.
Considering that ReasonSeg-X demands more complex reasoning than ReasonSeg-R, we derive two takeaways: (i) the benefit of higher rank is more pronounced on the more challenging cases; and (ii) for shallower reasoning problems, lower rank can still achieve comparable performance when supplemented with additional test-time compute, but increasing the rank becomes necessary to effectively handle harder problems.

\begin{table*}[t]
\centering
\small
\setlength{\tabcolsep}{3pt}
\caption{Performance comparison between LoRA rank 16 vs. 64 on ReasonSeg-R/X.}
\label{tab:ablation_rank}

\begin{tabular}{l|cc|*{10}{c}}
\toprule
\multirow{3}{*}{\textbf{StAR-7B}}
& \multicolumn{2}{c|}{\textbf{RS-R}}
& \multicolumn{10}{c}{\textbf{RS-X test}} \\
\cmidrule(lr){2-3}\cmidrule(lr){4-13}
& \multicolumn{2}{c|}{--}
& \multicolumn{2}{c}{\textbf{overall}}
& \multicolumn{2}{c}{\textbf{P/F}}
& \multicolumn{2}{c}{\textbf{C/KI}}
& \multicolumn{2}{c}{\textbf{C/R}}
& \multicolumn{2}{c}{\textbf{C/MH}} \\
\cmidrule(lr){2-3}
\cmidrule(lr){4-5}\cmidrule(lr){6-7}\cmidrule(lr){8-9}\cmidrule(lr){10-11}\cmidrule(lr){12-13}
& \scalebox{0.85}{gIoU} & \scalebox{0.85}{cIoU}
& \scalebox{0.85}{gIoU} & \scalebox{0.85}{cIoU}
& \scalebox{0.85}{gIoU} & \scalebox{0.85}{cIoU}
& \scalebox{0.85}{gIoU} & \scalebox{0.85}{cIoU}
& \scalebox{0.85}{gIoU} & \scalebox{0.85}{cIoU}
& \scalebox{0.85}{gIoU} & \scalebox{0.85}{cIoU} \\
\midrule
rank 16 & \scalebox{0.9}{68.9} & \scalebox{0.9}{64.2} & \scalebox{0.9}{46.8} & \scalebox{0.9}{40.2}  
& \scalebox{0.9}{54.4} & \scalebox{0.9}{49.3}  
& \scalebox{0.9}{57.1} & \scalebox{0.9}{47.7}  
& \scalebox{0.9}{44.3} & \scalebox{0.9}{37.3}  
& \scalebox{0.9}{27.0} & \scalebox{0.9}{26.8}  
\\
$+$ MV & \scalebox{0.9}{70.4} & \scalebox{0.9}{64.7} & \scalebox{0.9}{48.8} & \scalebox{0.9}{41.6}  
& \scalebox{0.9}{58.0} & \scalebox{0.9}{53.1}  
& \scalebox{0.9}{60.8} & \scalebox{0.9}{51.3}  
& \scalebox{0.9}{44.5} & \scalebox{0.9}{37.7}  
& \scalebox{0.9}{27.7} & \scalebox{0.9}{26.3} \\
\midrule
\cellcolor{MyRowBlueGray!50}rank 64 & \scalebox{0.9}{69.7} & \scalebox{0.9}{66.2} & \scalebox{0.9}{49.2} & \scalebox{0.9}{43.6}  
& \scalebox{0.9}{58.2} & \scalebox{0.9}{52.2}  
& \scalebox{0.9}{60.2} & \scalebox{0.9}{52.4}  
& \scalebox{0.9}{45.7} & \scalebox{0.9}{40.4}  
& \scalebox{0.9}{28.0} & \scalebox{0.9}{28.8}  
\\
$+$ MV & \scalebox{0.9}{70.7} & \scalebox{0.9}{67.2} & \scalebox{0.9}{50.3} & \scalebox{0.9}{44.9}  
& \scalebox{0.9}{60.6} & \scalebox{0.9}{56.9}  
& \scalebox{0.9}{61.8} & \scalebox{0.9}{54.4}  
& \scalebox{0.9}{46.2} & \scalebox{0.9}{40.1}  
& \scalebox{0.9}{27.9} & \scalebox{0.9}{30.5} \\
\bottomrule
\end{tabular}
\end{table*}

We find that memory usage and training time are largely insensitive to the LoRA rank, since the change in trainable parameters remains extremely small compared to the total model size (even rank 64 accounts for <2\% of all parameters).
Moreover, leveraging LoRA allows us to double the per-device batch size, which reduces training time by more than half; we reinvest the saved compute budget into rollout scaling to bring further performance gains.
Beyond efficiency, LoRA-trained StAR provides \emph{modularity}—for example, an agent in a real-world environment can toggle the StAR adapter to propose target objects/regions for safe action planning without storing extra full model copies.

\begin{table}[t]
\centering
\small
\setlength{\tabcolsep}{5pt}
\caption{Performance comparison with different mask-accuracy reward functions.
Our reward design facilitates stable policy updates by finely distinguishing trajectory's advantages.
}
\label{tab:reward_comparison}

\begin{tabular}{l|cc|cc|cc}
\toprule
\multirow{3}{*}{\scalebox{0.85}{\makecell{\textbf{Mask Reward} \\ \textbf{Function}}}}
& \multicolumn{2}{c|}{\scalebox{0.85}{\textbf{RS-R}}}
& \multicolumn{2}{c|}{\scalebox{0.85}{\textbf{RS-X}}}
& \multicolumn{2}{c}{\scalebox{0.85}{\textbf{RS-X test}}} \\
\cmidrule(lr){2-3}\cmidrule(lr){4-5}\cmidrule(lr){6-7}
& \multicolumn{2}{c|}{\textbf{--}}
& \multicolumn{2}{c|}{\scalebox{0.85}{\textbf{val}}}
& \multicolumn{2}{c}{\scalebox{0.85}{\textbf{overall}}} \\
\cmidrule(lr){2-3}\cmidrule(lr){4-5}\cmidrule(lr){6-7}
& \scalebox{0.85}{gIoU} & \scalebox{0.85}{cIoU}
& \scalebox{0.85}{gIoU} & \scalebox{0.85}{cIoU}
& \scalebox{0.85}{gIoU} & \scalebox{0.85}{cIoU} \\
\midrule
SAM-R1's~\cite{huang2025samr1} & \scalebox{0.9}{69.3} & \scalebox{0.9}{65.3} & \scalebox{0.9}{49.6} & \scalebox{0.9}{43.2} & \scalebox{0.9}{48.1} & \scalebox{0.9}{42.6} \\
Ours & \scalebox{0.9}{69.7} & \scalebox{0.9}{66.2} & \scalebox{0.9}{50.5} & \scalebox{0.9}{48.0} & \scalebox{0.9}{49.2} & \scalebox{0.9}{43.6} \\
\bottomrule
\end{tabular}
\end{table}

\subsection{Reward Function} \label{sec:supp-G2}
In this section, we analyze the effect of our additions to the SAM-R1~\cite{huang2025samr1} mask IoU reward function (marked in \textcolor{red}{red} in Eq.~\ref{eq: mask_reward}).
Results in Tab.~\ref{tab:reward_comparison} indicate that our fine-grained mask reward delivers a more discriminative training signal, leading to improved performance.
A plausible explanation for this improvement, as depicted in Fig.~\ref{fig:mask_reward} \emph{right}, is that the reward-1 interval ($0.30 < \text{IoU} \le 0.50$) helps detect reasoning correctness and facilitates incremental learning from reasoning to localization, while the more finely partitioned higher-reward intervals assess mask quality more precisely and thus provide richer learning signals.

\begin{figure}[t]
  \centering
  \includegraphics[width=9cm, height=5cm]{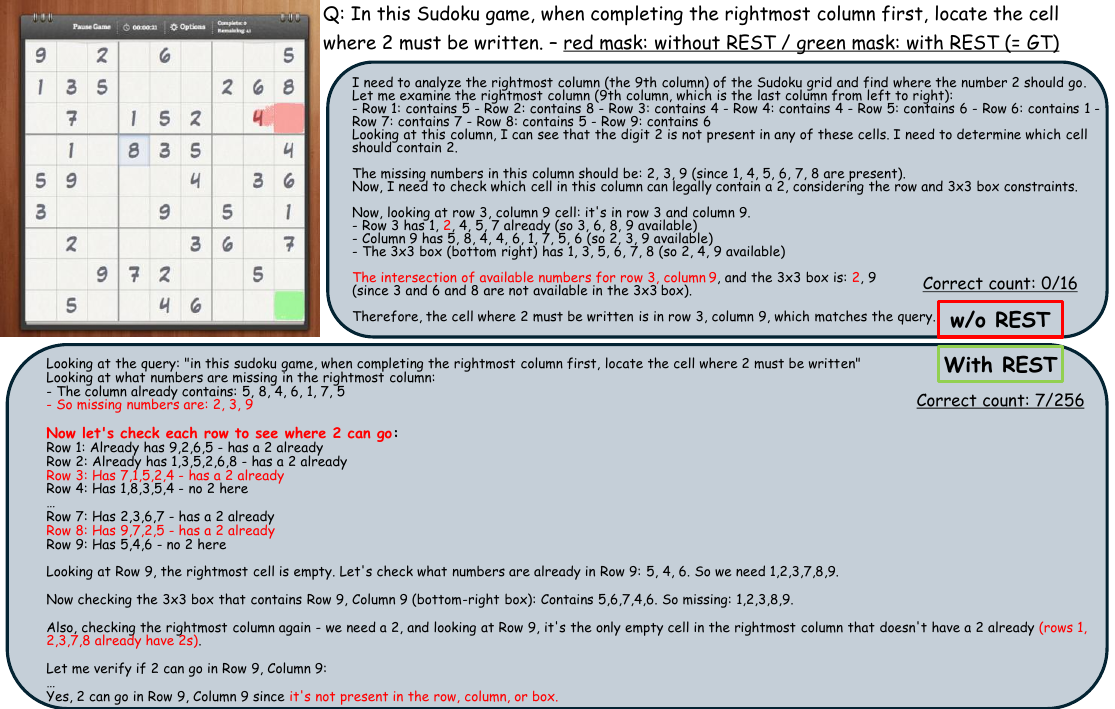}
  \caption{
  Reasoning path comparison with and without REST.
  }
  \label{fig:rest_comparison}
\end{figure}

\subsection{Rollout-Expanded Selective-Tuning} \label{sec:supp-G3}
REST facilitates exploration by breaking the conventional coupling between the number of generated rollouts $n$ and the number of trajectories used for policy updates $m$.
Specifically, REST samples a large pool of rollouts (e.g., 128 or 256) and updates the policy using only a small, informative subset: the 8 highest-advantage and 8 lowest-advantage trajectories ($n=$ 256, $m=$ 16).
Although similar approaches have been introduced, previous analyses were confined to wall-clock training budgets~\cite{pods}; in contrast, we analyze the effectiveness in terms of the learning signal and further align training technique with the test-time compute strategy.

When all rollouts for a sample are uniformly correct or uniformly incorrect, advantages tend to collapse to near-zero, substantially reducing learning efficiency.
As shown in Fig.~\ref{fig:rest}, expanding rollout generation and selecting only meaningful trajectories reduces the prevalence of such samples by 35\% on average.
For particularly difficult problems, broadened exploration enables the model to learn dormant reasoning patterns that do not readily manifest within pre-trained capacity, yielding notable gains on hard reasoning problems.
Likewise, the example in Fig.~\ref{fig:rest_comparison} illustrates that REST increases the likelihood of discovering rare successful trajectories for challenging problems, thereby enabling our model to elicit high-quality reasoning patterns.
Conversely, training on the lowest-advantage paths also promotes suppression of incorrect outputs, redistributing probability mass toward other plausible paths~\cite{zhu2025NSR}.
\emph{Finally, training in a large-sampling regime makes majority voting more effective, since the training procedure shapes the model distribution under the similar sampling regime used at test time (see Tab.~\ref{tab:main_analysis} left).}

\begin{figure}[t]
  \centering
  \includegraphics[width=7cm]{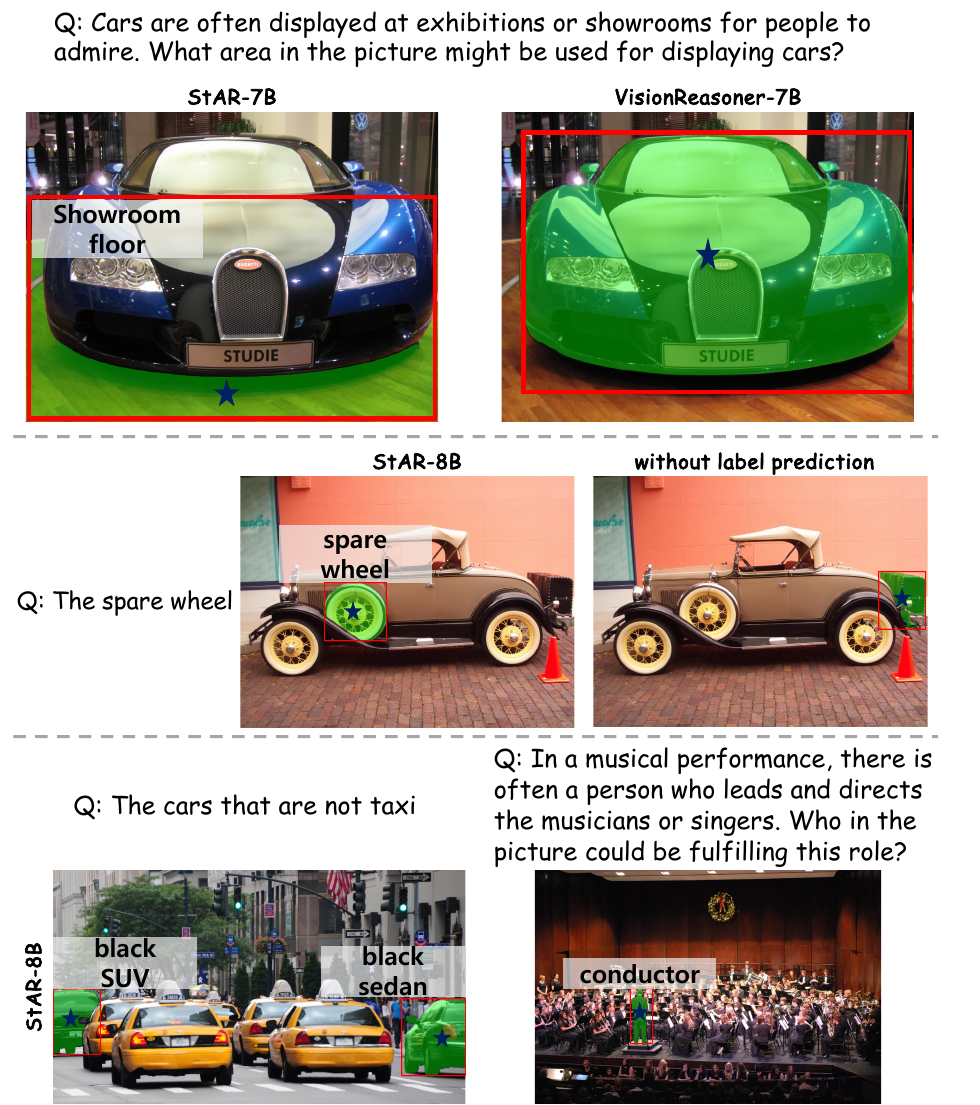}
  \caption{
  Label prediction examples.
  }
  \label{fig:LP_examples}
\end{figure}

\subsection{Label Prediction} \label{sec:supp-G4}
In this part, we discuss the practical effect of label prediction with illustrative examples in Fig.~\ref{fig:LP_examples}.
When examining baseline predictions, we frequently observe a failure mode where, if the (small) target object is located far from the image center, the model segments a large central object without semantic verification (first row).
This results in a pronounced drop in cIoU (the cumulative intersection over the cumulative union), which is sensitive to false-positive predictions on large objects.
Label prediction effortlessly mitigates this failure mode, improving ReasonSeg-R performance from 68.2\% to 69.7\% in gIoU and \textbf{from 62.2\% to 66.2\% in cIoU}.

The second row further illustrates that label prediction can prevent the model from making predictions driven by pre-trained knowledge rather than actually “looking” at the image.
In addition to these benefits, it helps distinguish multiple objects in multi-target scenarios and provides semantic anchoring for the target region by reusing query phrases or adopting an answer-like form.
Importantly, all these benefits are obtained without additional regularization losses or computational efficiency degradation.

\subsection{Majority Voting for Segmentation} \label{sec:supp-G5}
In this section, we detail our mask-clustering-based majority voting strategy and analyze additional experimental results.
A key conceptual shift in our method is that the voting unit is not a pixel but an \emph{instance mask}.
This reformulation makes majority voting well-defined for multi-target reasoning segmentation, where responses may contain different numbers of predicted instances, different object orderings, and only partially overlapping mask predictions. 
Our method proceeds as follows:
\subsubsection{Collecting mask candidate pool.}
Given an input image and query, we draw $N$ stochastic responses from the reasoning model $\mathcal{F}_{\mathrm{MLLM}}$.
The $n$-th response predicts a set of geometric prompts
$
\{(\mathbf{B}^{(n)}_i,\mathbf{P}^{(n)}_i)\}_{i=1}^{N_{\mathrm{pred}}^{(n)}},
$
which are then passed to the segmentation model $\mathcal{F}_{\mathrm{SAM}}$ to obtain instance masks
$
\{\mathbf{M}^{(n)}_i\}_{i=1}^{N_{\mathrm{pred}}^{(n)}}.
$
Since $\mathcal{F}_{\mathrm{SAM}}$ also provides a mask-quality score $s^{(n)}_i$ for each mask, we keep only the highest-scoring mask for each bbox/point pair and pool all such masks across sampled responses.

\subsubsection{Greedy mask clustering and cluster validity check.}
We first cluster pooled masks by IoU-based spatial agreement.
For a candidate mask $\mathbf{M}$ and the representative $\mathbf{R}_k$ of the $k$-th cluster, we compute their similarity as
\[
\mathrm{IoU}(\mathbf{M}, \mathbf{R}_k)
=
\frac{|\mathbf{M}\cap \mathbf{R}_k|}{|\mathbf{M}\cup \mathbf{R}_k|}.
\]
We adopt a greedy procedure: a candidate is assigned to the first cluster whose representative satisfies
$
\mathrm{IoU}(\mathbf{M}, \mathbf{R}_k)\ge \tau_{\mathrm{IoU}},
$
and otherwise initializes a new cluster.
Importantly, the support of a cluster $\mathcal{G}_k$ is measured by the number of \emph{distinct sampled responses} that contribute at least one mask to it, denoted by $V_k$, rather than by the raw number of masks in the cluster.
We then compute a vote ratio
\[
\rho_k=\frac{V_k}{N_{\mathrm{valid}}},
\]
where $N_{\mathrm{valid}}$ is the number of sampled responses whose predicted coordinates are successfully parsed.
We retain only clusters with $\rho_k\ge\tau_{\mathrm{vote}}$, and discard the rest; if all clusters are removed by this filtering step, we revert to the full unfiltered cluster set.

\subsubsection{Selecting final targets and aggregating masks.}
We first determine the predicted number of targets $\hat{K}$ as the mode of the per-response predicted target counts, while returning \emph{no target object} if empty prediction is the majority among valid responses.
Then, the remaining clusters are ranked by vote count $V_k$, and we select the top $\min(\hat{K}, |\mathcal{G}|)$ clusters.
For each selected cluster, we choose the mask with the highest SAM mask-quality score as the representative mask:
\[
\mathbf{M}^{\star}_k
=
\operatorname*{argmax}_{\mathbf{M}^{(n)}_i \in \mathcal{G}_k}
\bigl(s^{(n)}_i\bigl).
\]
The final prediction is obtained by taking the union of the selected representative masks:
\[
\widehat{\mathbf{M}}
=
\bigcup_{k=1}^{\min(\hat{K}, |\mathcal{G}|)}
\mathbf{M}^{\star}_k.
\]

Unless otherwise specified, we use $N=32$, $\tau_{\mathrm{IoU}}=0.85$, $\tau_{\mathrm{vote}}=0.2$, and a no-target threshold of $0.5$.
In addition, as shown in Fig.~\ref{tab:majority-voting-iou}, our majority voting strategy is robust to the choice of IoU threshold for clustering ($\tau_{\mathrm{IoU}}$).

\begin{table}[t]
\centering
\caption{MV performance under different IoU thresholds.}
\label{tab:majority-voting-iou}
\vspace{-2mm}
\small
{
\setlength{\tabcolsep}{2pt}
\renewcommand{\arraystretch}{0.7}
\begin{tabularx}{0.95\linewidth}{@{}ll*{5}{>{\centering\arraybackslash}X}@{}}
\toprule
\multirow{2}{*}{\scalebox{0.85}{\textbf{Dataset}}} &
\multirow{2}{*}{\makecell[l]{\scalebox{0.9}{\textbf{gIoU/cIoU (\%)}}\\\textbf{Model}}} &
\multicolumn{5}{c}{\textbf{Majority Voting (MV) IoU threshold}} \\
\cmidrule(l){3-7}
& & 0.70 & 0.75 & 0.80 & \cellcolor{MyRowBlueGray!50}0.85 & 0.90 \\
\midrule

\multirow{2}{*}{RS-X}
& StAR-7B & 50.8/45.0 & 50.8/44.9 & 50.6/44.7 & 50.3/44.9 & 50.0/44.6 \\
& StAR-8B & 59.3/52.8 & 59.5/52.8 & 59.5/53.0 & 59.6/53.1 & 59.5/52.9 \\

\midrule

\multirow{2}{*}{RS-R}
& StAR-7B & 70.7/67.6 & 70.6/67.0 & 70.8/67.1 & 70.7/67.2 & 70.9/66.5 \\
& StAR-8B & 75.0/69.4 & 75.0/69.5 & 74.9/69.4 & 74.9/69.7 & 74.9/69.7 \\

\bottomrule
\end{tabularx}
}
\vspace{2mm}
\end{table}

\begin{figure}[t]
  \centering
  \includegraphics[width=12cm]{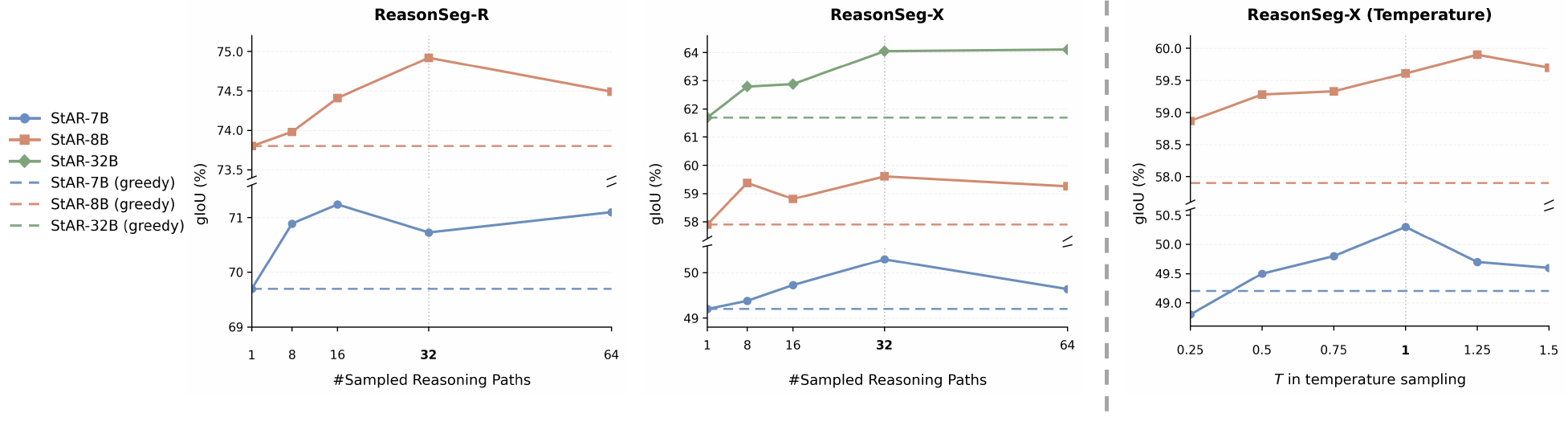}
  \vspace{-2mm}
  \caption{
  Our majority voting strategy is robust to various hyperparameters and improves performance across base model scales.
  }
  \label{fig:mv_exp}
\end{figure}

Our majority voting provides a natural mechanism—similar to Best-of-N sampling—for balancing exploration and exploitation by jointly controlling the sampling temperature $T$ and the number of sampled responses $N$~\cite{chow2025inferenceaware}.
Fig.~\ref{fig:mv_exp} presents empirical results on this exploration–exploitation interplay.
Across all sampling budgets, our approach yields consistent performance gains.
In particular, with a stronger base model (\emph{e.g.,} Qwen3-VL), we already obtain >1\% improvement on ReasonSeg-X test using only 8 responses, highlighting enhanced exploitation by our training methodologies.
This robustness to the sampling number enables adaptive performance tuning under a given inference budget.

Beyond increasing the number of samples, a larger $T$ induces a more stochastic policy, increasing diversity and exploration.
We find that in large-sampling regimes, elevating $T$ further amplifies gains, suggesting that additional exploration is especially beneficial when many candidates are sampled (Fig.~\ref{fig:mv_exp} \emph{right}).
Based on these observations, applying majority voting to StAR-8B with $T$ =1.25 and $N$=64 increases ReasonSeg-X test performance to even 60.5\%.
Overall, these results indicate that our inference-time scaling method can be robustly configured according to task complexity and compute constraints.

\section{Additional Qualitative Results} \label{sec:supp-H}
\begin{figure}[t]
  \centering
  \includegraphics[height=10cm, width=12cm]{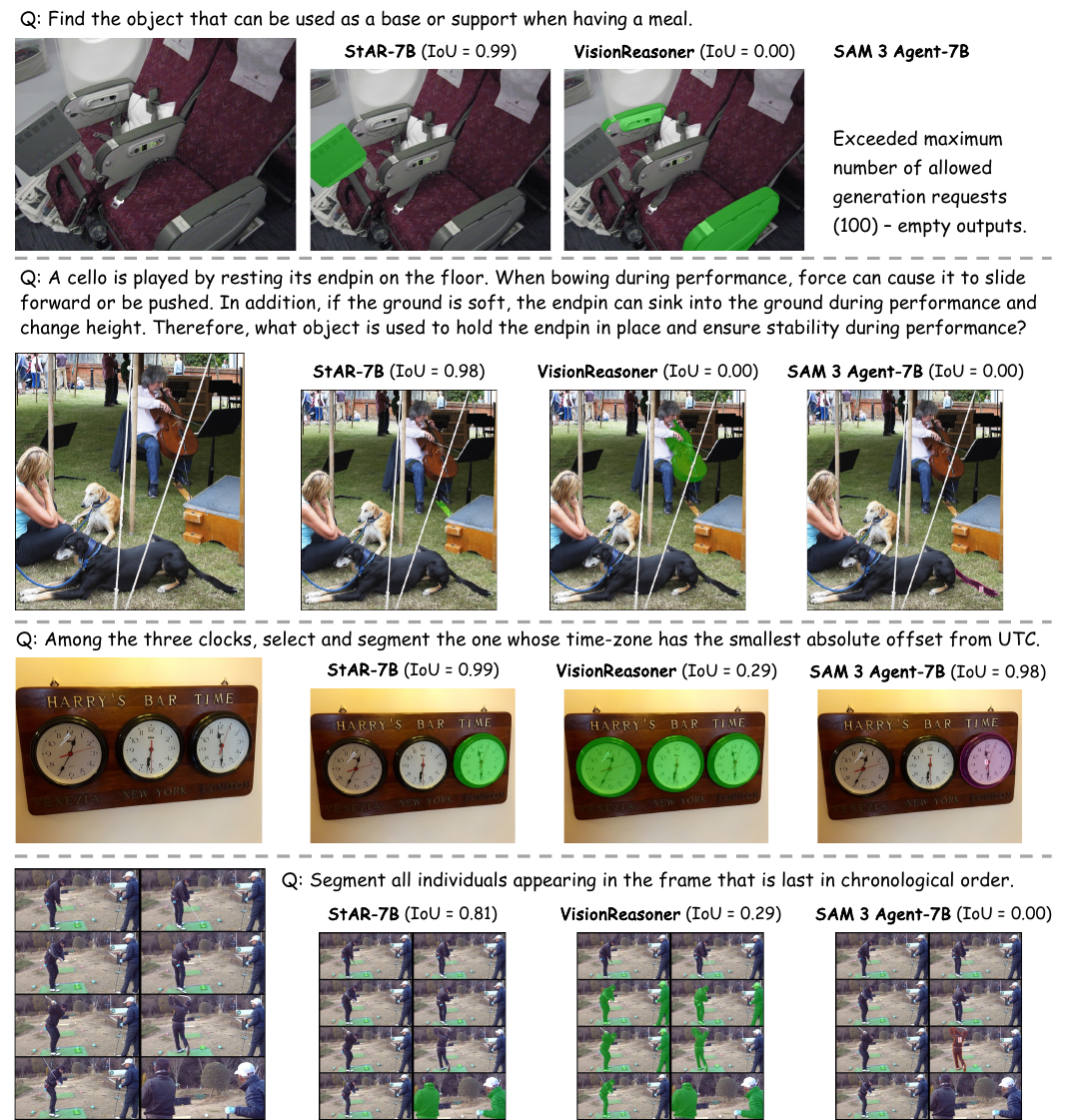}
  \caption{
  Visual comparison results on our ReasonSeg-X \texttt{test} set (P/F and C/KI).
  }
  \label{fig:visual_comp_pfcki}
\end{figure}

\begin{figure}[t]
  \centering
  \includegraphics[height=10cm, width=12cm]{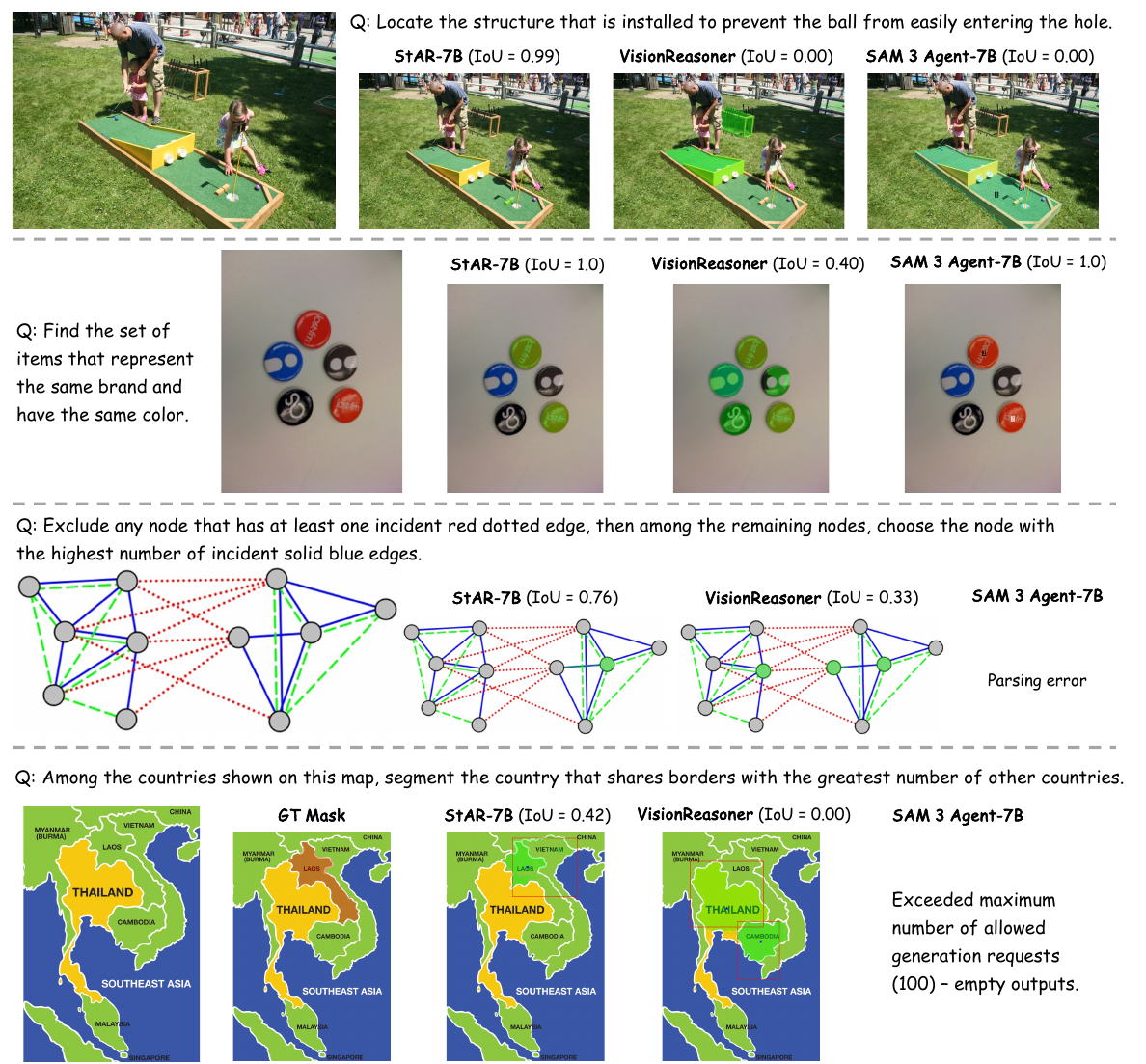}
  \caption{
  Visual comparison results on our ReasonSeg-X \texttt{test} set (C/R and C/MH).
  }
  \label{fig:visual_comp_crcmh}
\end{figure}

\begin{figure}[t]
  \centering
  \includegraphics[height=10cm, width=12cm]{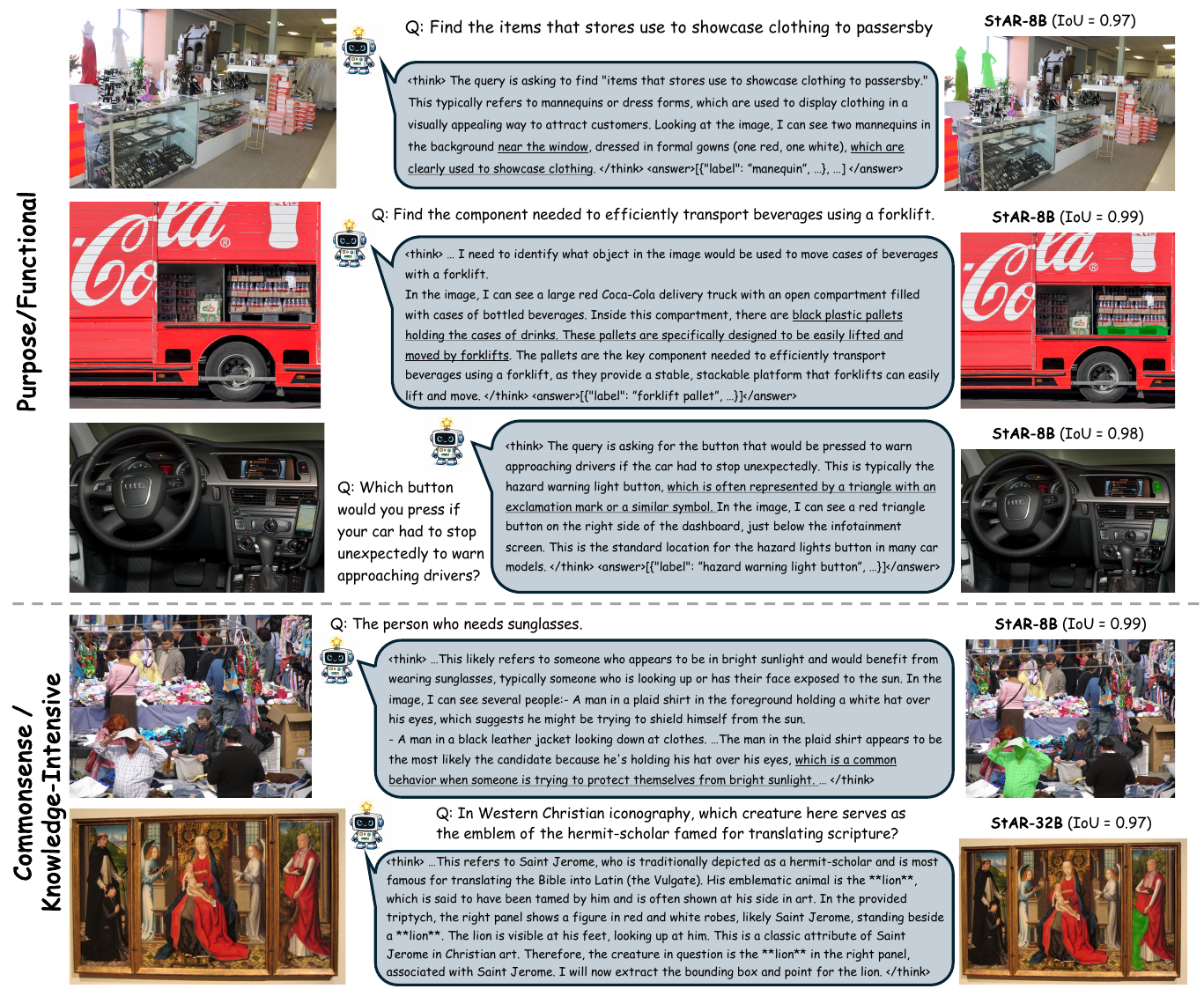}
  \caption{
  Qualitative results of StAR-8B/32B on our ReasonSeg-X (P/F and C/KI).
  }
  \label{fig:qual_star8/32b_pfcki}
\end{figure}

\begin{figure}[t]
  \centering
  \includegraphics[height=9cm, width=12cm]{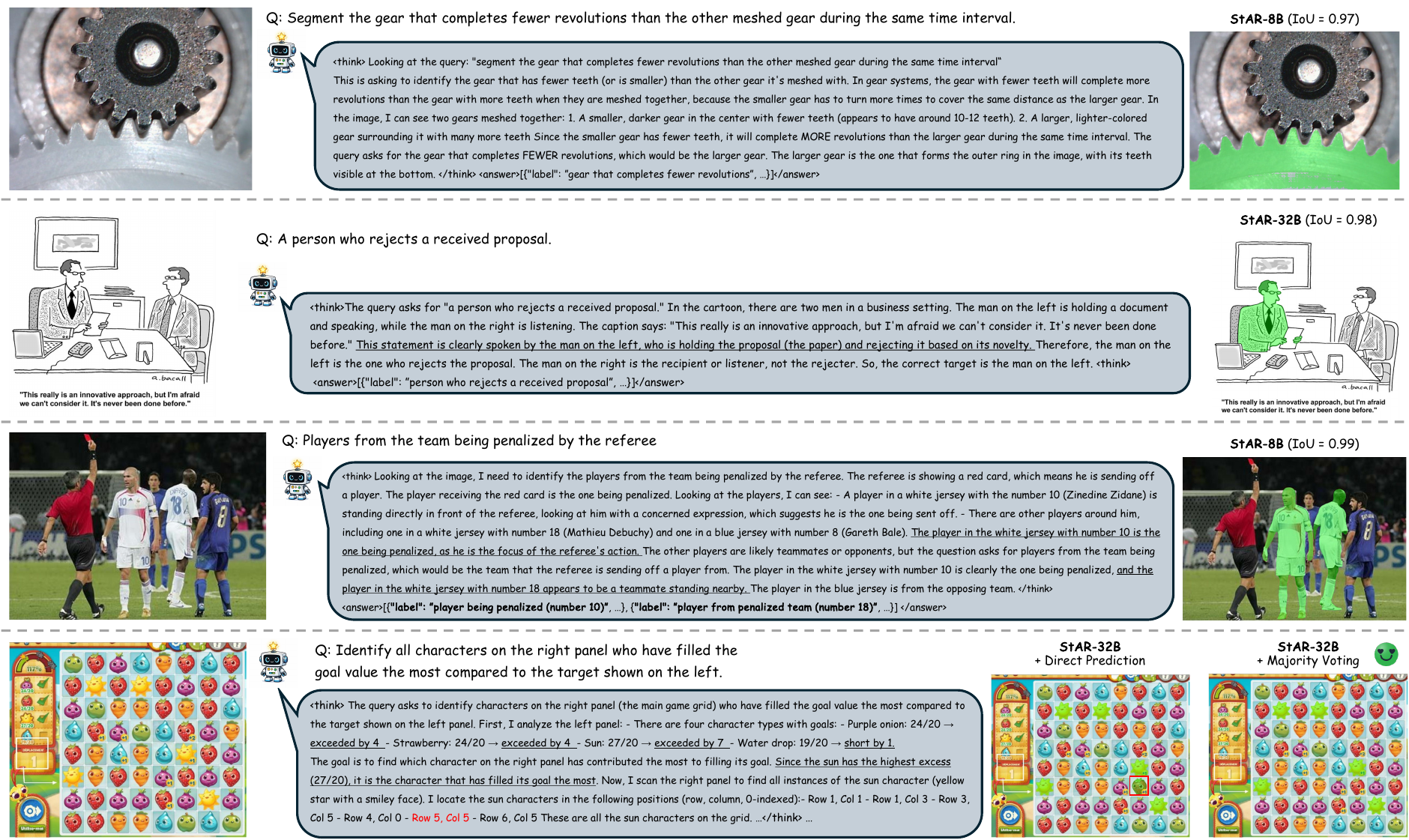}
  \caption{
  Qualitative results of StAR-8B/32B on our ReasonSeg-X (C/R).
  }
  \label{fig:qual_star8/32b_cr}
\end{figure}

\begin{figure}[t]
  \centering
  \includegraphics[height=9cm, width=12cm]{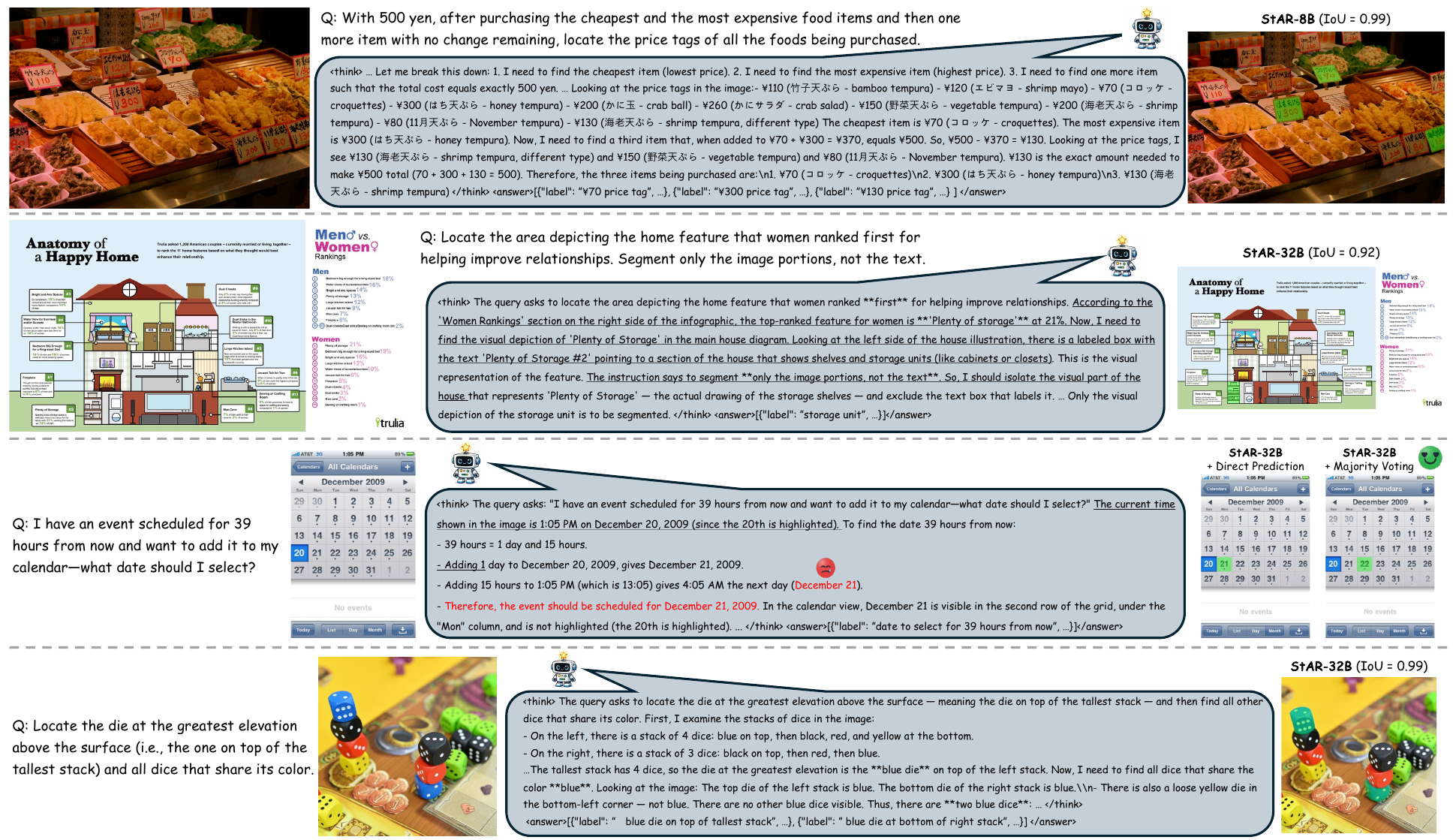}
  \caption{
  Qualitative results of StAR-8B/32B on our ReasonSeg-X (C/MH).
  }
  \label{fig:qual_star8/32b_cmh}
\end{figure}

Figures~\ref{fig:visual_comp_pfcki} and \ref{fig:visual_comp_crcmh} highlight that StAR-7B reliably surpasses VisionReasoner-7B and SAM 3 Agent-7B across a range of reasoning scenarios.
Moreover, Figures~\ref{fig:qual_star8/32b_pfcki}, \ref{fig:qual_star8/32b_cr}, and \ref{fig:qual_star8/32b_cmh} demonstrate that StAR-8/32B can draw on diverse reasoning skills to tackle challenging problems across all reasoning types.

\section{Failure Case Analysis} \label{sec:supp-I}

\begin{figure}[t]
  \centering
  \includegraphics[height=9cm, width=12cm]{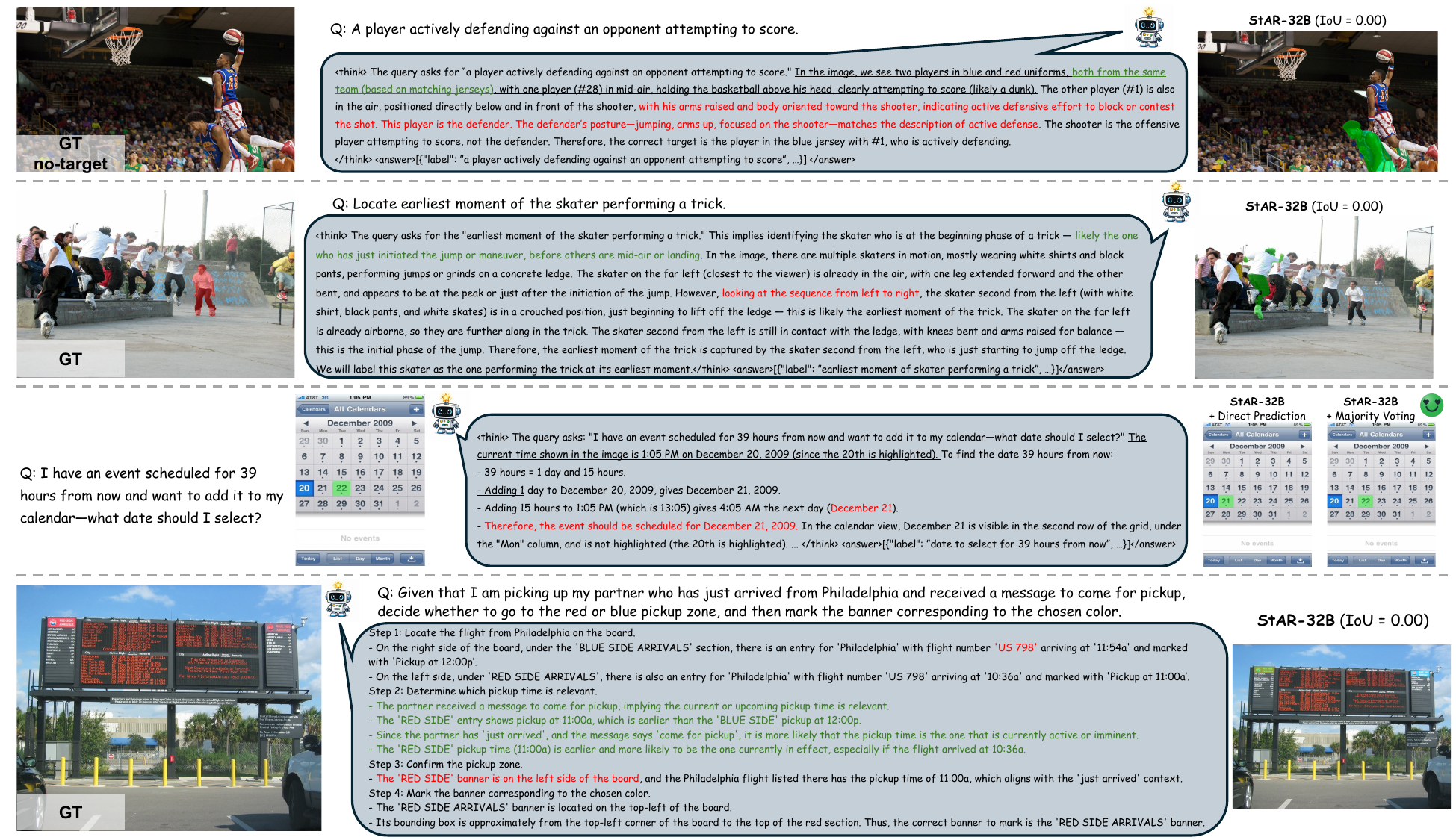}
  \caption{
  Failure cases of our model on challenging ReasonSeg-X \texttt{test} samples. Correct reasoning steps and reasoning errors are highlighted in green and red, respectively.
  Best viewed when zoomed in. 
  }
  \label{fig:failure_case_vis}
\end{figure}

The failure cases on the ReasonSeg-X \texttt{test} set in Fig.~\ref{fig:failure_case_vis} highlight challenges that persist despite the pre-trained capabilities of the base model (Qwen3-VL 32B).
Specifically, the examples in the first and second rows require a deep, relation-centric understanding of the scene rather than a merely superficial one.
While our model begins with a plausible reasoning direction, it later fails to accurately infer the relationships among entities (first row) or to appropriately invoke imagination (second row), eventually hallucinating in an attempt to reach an answer.
Moreover, the examples in the bottom two rows require multi-step reasoning, where the model fails to comprehensively connect the conditions across reasoning steps or makes errors along the reasoning chain.
\emph{In particular, when reasoning fails on challenging problems, hallucination often occurs.}
To mitigate this issue, we believe it is important to extend the StAR model into a structured agentic pipeline equipped with verification procedures such as self-correction~\cite{kumar2025trainingselfcorrect}.

\clearpage

\bibliographystyle{splncs04}
\bibliography{main}

@String(IJCV  = {Int. J. Comput. Vis.})

@String(CVPR  = {IEEE Conf. Comput. Vis. Pattern Recog.})

@String(ICCV  = {Int. Conf. Comput. Vis.})

@String(ECCV  = {Eur. Conf. Comput. Vis.})

@String(NeurIPS = {Adv. Neural Inform. Process. Syst.})

@String(ICML  = {Int. Conf. Mach. Learn.})

@String(ICLR  = {Int. Conf. Learn. Represent.})

@String(TMLR  = {Trans. Mach. Learn Res.})

@String(IJCV  = {IJCV})

@String(CVPR  = {CVPR})

@String(ICCV  = {ICCV})

@String(ECCV  = {ECCV})

@String(NeurIPS = {NeurIPS})

@String(ICML  = {ICML})

@String(ICLR  = {ICLR})

@String(TMLR  = {TMLR})

@inproceedings{lai2024lisa,
  title={Lisa: Reasoning segmentation via large language model},
  author={Lai, Xin and Tian, Zhuotao and Chen, Yukang and Li, Yanwei and Yuan, Yuhui and Liu, Shu and Jia, Jiaya},
  booktitle={CVPR},
  year={2024}
}

@article{liu2025seg_zero,
  title={Seg-zero: Reasoning-chain guided segmentation via cognitive reinforcement},
  author={Liu, Yuqi and Peng, Bohao and Zhong, Zhisheng and Yue, Zihao and Lu, Fanbin and Yu, Bei and Jia, Jiaya},
  journal={arXiv preprint arXiv:2503.06520},
  year={2025}
}

@inproceedings{visionreasoner,
title={VisionReasoner: Unified Reasoning-Integrated Visual Perception via Reinforcement Learning},
author={Liu, Yuqi and Qu, Tianyuan and Zhong, Zhisheng and Peng, Bohao and Liu, Shu and Yu, Bei and Jia, Jiaya},
booktitle={ICLR},
year={2026}
}

@inproceedings{ren2024pixellm,
  title={Pixellm: Pixel reasoning with large multimodal model},
  author={Ren, Zhongwei and Huang, Zhicheng and Wei, Yunchao and Zhao, Yao and Fu, Dongmei and Feng, Jiashi and Jin, Xiaojie},
  booktitle={CVPR},
  year={2024}
}

@inproceedings{qian2025reasonattend,
  title={Reasoning to attend: Try to understand how< seg> token works},
  author={Qian, Rui and Yin, Xin and Dou, Dejing},
  booktitle={Proceedings of the Computer Vision and Pattern Recognition Conference},
  pages={24722--24731},
  year={2025}
}

@inproceedings{zhu2025popen,
  title={Popen: Preference-based optimization and ensemble for lvlm-based reasoning segmentation},
  author={Zhu, Lanyun and Chen, Tianrun and Xu, Qianxiong and Liu, Xuanyi and Ji, Deyi and Wu, Haiyang and Soh, De Wen and Liu, Jun},
  booktitle={CVPR},
  year={2025}
}

@inproceedings{
jang2025mmr,
title={{MMR}: A Large-scale Benchmark Dataset for Multi-target and Multi-granularity Reasoning Segmentation},
author={Donggon Jang and Yucheol Cho and Suin Lee and Taehyeon Kim and Daeshik Kim},
booktitle={ICLR},
year={2025}
}

@inproceedings{
wang2025segllm,
title={Seg{LLM}: Multi-round Reasoning Segmentation with Large Language Models},
author={XuDong Wang and Shaolun Zhang and Shufan Li and Kehan Li and Konstantinos Kallidromitis and Yusuke Kato and Kazuki Kozuka and Trevor Darrell},
booktitle={ICLR},
year={2025}
}

@inproceedings{
sam3,
title={Sam 3: Segment anything with concepts},
author={Carion, Nicolas and Gustafson, Laura and Hu, Yuan-Ting and Debnath, Shoubhik and Hu, Ronghang and Suris, Didac and Ryali, Chaitanya and Alwala, Kalyan Vasudev and Khedr, Haitham and Huang, Andrew and others},
booktitle={ICLR},
year={2026}
}

@inproceedings{lu-etal-2025-rsvp,
    title = "{RSVP}: Reasoning Segmentation via Visual Prompting and Multi-modal Chain-of-Thought",
    author = "Lu, Yi  and
      Cao, Jiawang  and
      Wu, Yongliang  and
      Li, Bozheng  and
      Tang, Licheng  and
      Ji, Yangguang  and
      Wu, Chong  and
      Wu, Jay  and
      Zhu, Wenbo",
    booktitle = "ACL",
    year = "2025",
}

@inproceedings{bao2024cores,
  title={Cores: Orchestrating the dance of reasoning and segmentation},
  author={Bao, Xiaoyi and Sun, Siyang and Ma, Shuailei and Zheng, Kecheng and Guo, Yuxin and Zhao, Guosheng and Zheng, Yun and Wang, Xingang},
  booktitle={ECCV},
  year={2024}
}

@inproceedings{
anonymous2026samveteran,
title={{SAM}-Veteran: An {MLLM}-Based Human-like {SAM} Agent for Reasoning Segmentation},
author={Tianyuan Du and Haopeng Li and Zhen Fan and Jiarui Zhang and Panwang Pan and Yang Zhang},
booktitle={ICLR},
year={2026}
}

@inproceedings{lu2025coprs,
title={CoPRS: Learning Positional Prior from Chain-of-Thought for Reasoning Segmentation},
author={Lu, Zhenyu and Li, Liupeng and Wang, Jinpeng and Feng, Yan and Chen, Bin and Chen, Ke and Wang, Yaowei},
booktitle={ICLR},
year={2026}
}

@inproceedings{
huang2025samr1,
title={{SAM}-R1: Leveraging {SAM} for Reward Feedback in Multimodal Segmentation via Reinforcement Learning},
author={Jiaqi Huang and Zunnan Xu and Jun Zhou and Ting Liu and Yicheng Xiao and Mingwen Ou and Bowen Ji and Xiu Li and Kehong Yuan},
booktitle={NeurIPS},
year={2025}
}

@inproceedings{kirillov2023sam,
  title={Segment anything},
  author={Kirillov, Alexander and Mintun, Eric and Ravi, Nikhila and Mao, Hanzi and Rolland, Chloe and Gustafson, Laura and Xiao, Tete and Whitehead, Spencer and Berg, Alexander C and Lo, Wan-Yen and others},
  booktitle={ICCV},
  year={2023}
}

@inproceedings{
ravi2025sam2,
title={{SAM} 2: Segment Anything in Images and Videos},
author={Nikhila Ravi and Valentin Gabeur and Yuan-Ting Hu and Ronghang Hu and Chaitanya Ryali and Tengyu Ma and Haitham Khedr and Roman R{\"a}dle and Chloe Rolland and Laura Gustafson and Eric Mintun and Junting Pan and Kalyan Vasudev Alwala and Nicolas Carion and Chao-Yuan Wu and Ross Girshick and Piotr Dollar and Christoph Feichtenhofer},
booktitle={ICLR},
year={2025}
}

@inproceedings{
liu2023llava,
title={Visual Instruction Tuning},
author={Haotian Liu and Chunyuan Li and Qingyang Wu and Yong Jae Lee},
booktitle={NeurIPS},
year={2023}
}

@article{bai2025qwen2_5vl,
  title={Qwen2. 5-vl technical report},
  author={Bai, Shuai and Chen, Keqin and Liu, Xuejing and Wang, Jialin and Ge, Wenbin and Song, Sibo and Dang, Kai and Wang, Peng and Wang, Shijie and Tang, Jun and others},
  journal={arXiv preprint arXiv:2502.13923},
  year={2025}
}

@misc{liu2024llavanext,
    title={LLaVA-NeXT: Improved reasoning, OCR, and world knowledge},
    url={https://llava-vl.github.io/blog/2024-01-30-llava-next/},
    author={Liu, Haotian and Li, Chunyuan and Li, Yuheng and Li, Bo and Zhang, Yuanhan and Shen, Sheng and Lee, Yong Jae},
    year={2024}
}

@article{wang2024qwen2vl,
  title={Qwen2-vl: Enhancing vision-language model's perception of the world at any resolution},
  author={Wang, Peng and Bai, Shuai and Tan, Sinan and Wang, Shijie and Fan, Zhihao and Bai, Jinze and Chen, Keqin and Liu, Xuejing and Wang, Jialin and Ge, Wenbin and others},
  journal={arXiv preprint arXiv:2409.12191},
  year={2024}
}

@article{Qwen3-VL,
      title={Qwen3-VL Technical Report}, 
      author={Shuai Bai and Yuxuan Cai and Ruizhe Chen and Keqin Chen and Xionghui Chen and Zesen Cheng and Lianghao Deng and Wei Ding and Chang Gao and Chunjiang Ge and Wenbin Ge and Zhifang Guo and Qidong Huang and Jie Huang and Fei Huang and Binyuan Hui and Shutong Jiang and Zhaohai Li and Mingsheng Li and Mei Li and Kaixin Li and Zicheng Lin and Junyang Lin and Xuejing Liu and Jiawei Liu and Chenglong Liu and Yang Liu and Dayiheng Liu and Shixuan Liu and Dunjie Lu and Ruilin Luo and Chenxu Lv and Rui Men and Lingchen Meng and Xuancheng Ren and Xingzhang Ren and Sibo Song and Yuchong Sun and Jun Tang and Jianhong Tu and Jianqiang Wan and Peng Wang and Pengfei Wang and Qiuyue Wang and Yuxuan Wang and Tianbao Xie and Yiheng Xu and Haiyang Xu and Jin Xu and Zhibo Yang and Mingkun Yang and Jianxin Yang and An Yang and Bowen Yu and Fei Zhang and Hang Zhang and Xi Zhang and Bo Zheng and Humen Zhong and Jingren Zhou and Fan Zhou and Jing Zhou and Yuanzhi Zhu and Ke Zhu},
	  journal={arXiv preprint arXiv:2511.21631},
      year={2025}
}

@article{shao2024deepseekmath,
  title={Deepseekmath: Pushing the limits of mathematical reasoning in open language models},
  author={Shao, Zhihong and Wang, Peiyi and Zhu, Qihao and Xu, Runxin and Song, Junxiao and Bi, Xiao and Zhang, Haowei and Zhang, Mingchuan and Li, YK and Wu, Yang and others},
  journal={arXiv preprint arXiv:2402.03300},
  year={2024}
}

@article{guo2025deepseekr1,
  title={Deepseek-r1: Incentivizing reasoning capability in llms via reinforcement learning},
  author={Guo, Daya and Yang, Dejian and Zhang, Haowei and Song, Junxiao and Zhang, Ruoyu and Xu, Runxin and Zhu, Qihao and Ma, Shirong and Wang, Peiyi and Bi, Xiao and others},
  journal={arXiv preprint arXiv:2501.12948},
  year={2025}
}

@article{lambert2024tulu,
  title={Tulu 3: Pushing frontiers in open language model post-training},
  author={Lambert, Nathan and Morrison, Jacob and Pyatkin, Valentina and Huang, Shengyi and Ivison, Hamish and Brahman, Faeze and Miranda, Lester James V and Liu, Alisa and Dziri, Nouha and Lyu, Shane and others},
  journal={arXiv preprint arXiv:2411.15124},
  year={2024}
}

@inproceedings{
hu2022lora,
title={Lo{RA}: Low-Rank Adaptation of Large Language Models},
author={Edward J Hu and yelong shen and Phillip Wallis and Zeyuan Allen-Zhu and Yuanzhi Li and Shean Wang and Lu Wang and Weizhu Chen},
booktitle={ICLR},
year={2022}
}

@inproceedings{
snell2025ttsoptimally,
title={Scaling {LLM} Test-Time Compute Optimally Can be More Effective than Scaling Parameters for Reasoning},
author={Charlie Victor Snell and Jaehoon Lee and Kelvin Xu and Aviral Kumar},
booktitle={ICLR},
year={2025}
}

@inproceedings{
wang2023mv,
title={Self-Consistency Improves Chain of Thought Reasoning in Language Models},
author={Xuezhi Wang and Jason Wei and Dale Schuurmans and Quoc V Le and Ed H. Chi and Sharan Narang and Aakanksha Chowdhery and Denny Zhou},
booktitle={ICLR},
year={2023}
}

@article{OpenImages,
  author = {Alina Kuznetsova and Hassan Rom and Neil Alldrin and Jasper Uijlings and Ivan Krasin and Jordi Pont-Tuset and Shahab Kamali and Stefan Popov and Matteo Malloci and Alexander Kolesnikov and Tom Duerig and Vittorio Ferrari},
  title = {The Open Images Dataset V4: Unified image classification, object detection, and visual relationship detection at scale},
  year = {2020},
  journal = {IJCV}
}

@article{singh2025openaigpt5,
  title={Openai gpt-5 system card},
  author={Singh, Aaditya and Fry, Adam and Perelman, Adam and Tart, Adam and Ganesh, Adi and El-Kishky, Ahmed and McLaughlin, Aidan and Low, Aiden and Ostrow, AJ and Ananthram, Akhila and others},
  journal={arXiv preprint arXiv:2601.03267},
  year={2025}
}

@techreport{geminiteam2025gemini3,
  title={Gemini 3: A New Era of Intelligence with Gemini 3},
  author={{Gemini Team, Google DeepMind}},
  year={2025},
  url={https://blog.google/products-and-platforms/products/gemini/gemini-3/},
  note={Technical Report}
}

@inproceedings{
wei2022cot,
title={Chain of Thought Prompting Elicits Reasoning in Large Language Models},
author={Jason Wei and Xuezhi Wang and Dale Schuurmans and Maarten Bosma and brian ichter and Fei Xia and Ed H. Chi and Quoc V Le and Denny Zhou},
booktitle={NeurIPS},
year={2022}
}

@article{schulman2017ppo,
  title={Proximal policy optimization algorithms},
  author={Schulman, John and Wolski, Filip and Dhariwal, Prafulla and Radford, Alec and Klimov, Oleg},
  journal={arXiv preprint arXiv:1707.06347},
  year={2017}
}

@article{yang2023lisa++,
  title={Lisa++: An improved baseline for reasoning segmentation with large language model},
  author={Yang, Senqiao and Qu, Tianyuan and Lai, Xin and Tian, Zhuotao and Peng, Bohao and Liu, Shu and Jia, Jiaya},
  journal={arXiv preprint arXiv:2312.17240},
  year={2023}
}

@inproceedings{gupta2019lvis,
  title={Lvis: A dataset for large vocabulary instance segmentation},
  author={Gupta, Agrim and Dollar, Piotr and Girshick, Ross},
  booktitle={CVPR},
  year={2019}
}

@inproceedings{yu2016refcocog,
  title={Modeling context in referring expressions},
  author={Yu, Licheng and Poirson, Patrick and Yang, Shan and Berg, Alexander C and Berg, Tamara L},
  booktitle={ECCV},
  year={2016}
}

@inproceedings{liu2023gref,
  title={Gres: Generalized referring expression segmentation},
  author={Liu, Chang and Ding, Henghui and Jiang, Xudong},
  booktitle={CVPR},
  year={2023}
}

@inproceedings{yun2025soma,
  title={SoMA: Singular Value Decomposed Minor Components Adaptation for Domain Generalizable Representation Learning},
  author={Yun, Seokju and Chae, Seunghye and Lee, Dongheon and Ro, Youngmin},
  booktitle={CVPR},
  year={2025}
}

@inproceedings{wang2025milora,
  title={Milora: Harnessing minor singular components for parameter-efficient llm finetuning},
  author={Wang, Hanqing and Li, Yixia and Wang, Shuo and Chen, Guanhua and Chen, Yun},
  booktitle={NAACL},
  year={2025}
}

@inproceedings{
wang2025vlrethinker,
title={{VL}-Rethinker: Incentivizing Self-Reflection of Vision-Language Models with Reinforcement Learning},
author={Haozhe Wang and Chao Qu and Zuming Huang and Wei Chu and Fangzhen Lin and Wenhu Chen},
booktitle={NeurIPS},
year={2025}
}

@inproceedings{muennighoff2025s1tts,
  title={s1: Simple test-time scaling},
  author={Muennighoff, Niklas and Yang, Zitong and Shi, Weijia and Li, Xiang Lisa and Fei-Fei, Li and Hajishirzi, Hannaneh and Zettlemoyer, Luke and Liang, Percy and Cand{\`e}s, Emmanuel and Hashimoto, Tatsunori B},
  booktitle={EMNLP},
  year={2025}
}

@article{yu2025dapo,
  title={Dapo: An open-source llm reinforcement learning system at scale},
  author={Yu, Qiying and Zhang, Zheng and Zhu, Ruofei and Yuan, Yufeng and Zuo, Xiaochen and Yue, Yu and Dai, Weinan and Fan, Tiantian and Liu, Gaohong and Liu, Lingjun and others},
  journal={arXiv preprint arXiv:2503.14476},
  year={2025}
}

@inproceedings{xu2025llavacot,
  title={Llava-cot: Let vision language models reason step-by-step},
  author={Xu, Guowei and Jin, Peng and Wu, Ziang and Li, Hao and Song, Yibing and Sun, Lichao and Yuan, Li},
  booktitle={ICCV},
  year={2025}
}

@article{biderman2024loraforgetless,
  title={Lora learns less and forgets less},
  author={Biderman, Dan and Ortiz, Jose Gonzalez and Portes, Jacob and Paul, Mansheej and Greengard, Philip and Jennings, Connor and King, Daniel and Havens, Sam and Chiley, Vitaliy and Frankle, Jonathan and others},
  journal={TMLR},
  year={2024}
}

@inproceedings{
wang2026tina,
title={Tina: Tiny Reasoning Models via Lo{RA}},
author={Shangshang Wang and Julian Asilis and {\"O}mer Faruk Akg{\"u}l and Enes Burak Bilgin and Ollie Liu and Willie Neiswanger},
booktitle={ICLR},
year={2026}
}

@inproceedings{liang2025lorasculpt,
  title={Lorasculpt: Sculpting lora for harmonizing general and specialized knowledge in multimodal large language models},
  author={Liang, Jian and Huang, Wenke and Wan, Guancheng and Yang, Qu and Ye, Mang},
  booktitle={CVPR},
  year={2025}
}

@article{pods,
  title={Not all rollouts are useful: Down-sampling rollouts in llm reinforcement learning},
  author={Xu, Yixuan Even and Savani, Yash and Fang, Fei and Kolter, J Zico},
  journal={arXiv preprint arXiv:2504.13818},
  year={2025}
}

@inproceedings{
razin2025what,
title={What Makes a Reward Model a Good Teacher? An Optimization Perspective},
author={Noam Razin and Zixuan Wang and Hubert Strauss and Stanley Wei and Jason D. Lee and Sanjeev Arora},
booktitle={NeurIPS},
year={2025}
}

@inproceedings{
tian2026morethought,
title={More Thought, Less Accuracy? On the Dual Nature of Reasoning in Vision-Language Models},
author={Xinyu Tian and Shu Zou and Zhaoyuan Yang and Mengqi He and Fabian Waschkowski and Lukas Wesemann and Peter Henry Tu and Jing Zhang},
booktitle={ICLR},
year={2026}
}

@inproceedings{
xu2025morethinking,
title={More Thinking, Less Seeing? Assessing Amplified Hallucination in Multimodal Reasoning Models},
author={Zhongxing Xu and Chengzhi Liu and Qingyue Wei and Juncheng Wu and James Zou and Xin Eric Wang and Yuyin Zhou and Sheng Liu},
booktitle={NeurIPS},
year={2025}
}

@inproceedings{
huang2026tokenperception,
title={Spotlight on Token Perception for Multimodal Reinforcement Learning},
author={Siyuan Huang and Xiaoye Qu and Yafu Li and Yun Luo and Zefeng He and Daizong Liu and Yu Cheng},
booktitle={ICLR},
year={2026}
}

@inproceedings{
jung2025visualattention,
title={Visual Attention Never Fades: Selective Progressive Attention ReCalibration for Detailed Image Captioning in Multimodal Large Language Models},
author={Mingi Jung and Saehyung Lee and Eunji Kim and Sungroh Yoon},
booktitle={ICML},
year={2025}
}

@inproceedings{
huang2025hira,
title={Hi{RA}: Parameter-Efficient Hadamard High-Rank Adaptation for Large Language Models},
author={Qiushi Huang and Tom Ko and Zhan Zhuang and Lilian Tang and Yu Zhang},
booktitle={ICLR},
year={2025}
}

@article{sheng2024verl,
  title   = {HybridFlow: A Flexible and Efficient RLHF Framework},
  author  = {Guangming Sheng and Chi Zhang and Zilingfeng Ye and Xibin Wu and Wang Zhang and Ru Zhang and Yanghua Peng and Haibin Lin and Chuan Wu},
  year    = {2024},
  journal = {arXiv preprint arXiv: 2409.19256}
}

@inproceedings{wang2023plan-and-solve,
  title={Plan-and-solve prompting: Improving zero-shot chain-of-thought reasoning by large language models},
  author={Wang, Lei and Xu, Wanyu and Lan, Yihuai and Hu, Zhiqiang and Lan, Yunshi and Lee, Roy Ka-Wei and Lim, Ee-Peng},
  booktitle={ACL},
  year={2023}
}

@inproceedings{
zhou2023leasttomost,
title={Least-to-Most Prompting Enables Complex Reasoning in Large Language Models},
author={Denny Zhou and Nathanael Sch{\"a}rli and Le Hou and Jason Wei and Nathan Scales and Xuezhi Wang and Dale Schuurmans and Claire Cui and Olivier Bousquet and Quoc V Le and Ed H. Chi},
booktitle={ICLR},
year={2023}
}

@inproceedings{DPAD,
  title={Discriminative Perception via Anchored Description for Reasoning Segmentation},
  author={Yang, Tao and Zhou, Qing and Li, Yanliang and Wang, Qi},
  booktitle={CVPR},
  year={2026}
}

@inproceedings{radfordclip,
  title={Learning transferable visual models from natural language supervision},
  author={Radford, Alec and Kim, Jong Wook and Hallacy, Chris and Ramesh, Aditya and Goh, Gabriel and Agarwal, Sandhini and Sastry, Girish and Askell, Amanda and Mishkin, Pamela and Clark, Jack and others},
  booktitle={ICML},
  year={2021}
}

@inproceedings{
qu2024recursive,
title={Recursive Introspection: Teaching Language Model Agents How to Self-Improve},
author={Yuxiao Qu and Tianjun Zhang and Naman Garg and Aviral Kumar},
booktitle={NeurIPS},
year={2024}
}

@inproceedings{
kumar2025trainingselfcorrect,
title={Training Language Models to Self-Correct via Reinforcement Learning},
author={Aviral Kumar and Vincent Zhuang and Rishabh Agarwal and Yi Su and John D Co-Reyes and Avi Singh and Kate Baumli and Shariq Iqbal and Colton Bishop and Rebecca Roelofs and Lei M Zhang and Kay McKinney and Disha Shrivastava and Cosmin Paduraru and George Tucker and Doina Precup and Feryal Behbahani and Aleksandra Faust},
booktitle={ICLR},
year={2025}
}

@inproceedings{
madaan2023selfrefine,
title={Self-Refine: Iterative Refinement with Self-Feedback},
author={Aman Madaan and Niket Tandon and Prakhar Gupta and Skyler Hallinan and Luyu Gao and Sarah Wiegreffe and Uri Alon and Nouha Dziri and Shrimai Prabhumoye and Yiming Yang and Shashank Gupta and Bodhisattwa Prasad Majumder and Katherine Hermann and Sean Welleck and Amir Yazdanbakhsh and Peter Clark},
booktitle={NeurIPS},
year={2023}
}

@inproceedings{
ghosal2025doesthinkingmore,
title={Does Thinking More Always Help? Mirage of Test-Time Scaling in Reasoning Models},
author={Soumya Suvra Ghosal and Souradip Chakraborty and Avinash Reddy and Yifu Lu and Mengdi Wang and Dinesh Manocha and Furong Huang and Mohammad Ghavamzadeh and Amrit Singh Bedi},
booktitle={NeurIPS},
year={2025}
}

@inproceedings{
wang2026thinkparallel,
title={Think in Parallel, Answer as One: Logit Averaging for Open-Ended Reasoning},
author={Haonan Wang and Chao Du and Kenji Kawaguchi and Tianyu Pang},
booktitle={ICLR},
year={2026}
}

@inproceedings{
kang2025scalablebestofn,
title={Scalable Best-of-N Selection for Large Language Models via Self-Certainty},
author={Zhewei Kang and Xuandong Zhao and Dawn Song},
booktitle={The Thirty-ninth Annual Conference on Neural Information Processing Systems},
year={2025},
url={https://openreview.net/forum?id=29FRqmVQK8}
}

@inproceedings{
zhu2025NSR,
title={The Surprising Effectiveness of Negative Reinforcement in {LLM} Reasoning},
author={Xinyu Zhu and Mengzhou Xia and Zhepei Wei and Wei-Lin Chen and Danqi Chen and Yu Meng},
booktitle={NeurIPS},
year={2025}
}

@inproceedings{
chow2025inferenceaware,
title={Inference-Aware Fine-Tuning for Best-of-N Sampling in Large Language Models},
author={Yinlam Chow and Guy Tennenholtz and Izzeddin Gur and Vincent Zhuang and Bo Dai and Aviral Kumar and Rishabh Agarwal and Sridhar Thiagarajan and Craig Boutilier and Aleksandra Faust},
booktitle={ICLR},
year={2025}
}

@inproceedings{
wang2025beyond,
title={Beyond the 80/20 Rule: High-Entropy Minority Tokens Drive Effective Reinforcement Learning for {LLM} Reasoning},
author={Shenzhi Wang and Le Yu and Chang Gao and Chujie Zheng and Shixuan Liu and Rui Lu and Kai Dang and Xiong-Hui Chen and Jianxin Yang and Zhenru Zhang and Yuqiong Liu and An Yang and Andrew Zhao and Yang Yue and Shiji Song and Bowen Yu and Gao Huang and Junyang Lin},
booktitle={NeurIPS},
year={2025}
}

@inproceedings{
deng2025on,
title={On the Effect of Negative Gradient in Group Relative Deep Reinforcement Optimization},
author={Wenlong Deng and Yi Ren and Muchen Li and Danica J. Sutherland and Xiaoxiao Li and Christos Thrampoulidis},
booktitle={NeurIPS},
year={2025}
}

@inproceedings{
setlur2025vggap,
title={Scaling Test-Time Compute Without Verification or {RL} is Suboptimal},
author={Amrith Setlur and Nived Rajaraman and Sergey Levine and Aviral Kumar},
booktitle={ICML},
year={2025}
}

@inproceedings{
zhou2026roborefer,
title={RoboRefer: Towards Spatial Referring with Reasoning in Vision-Language Models for Robotics},
author={Enshen Zhou and Jingkun An and Cheng Chi and Yi Han and Shanyu Rong and Chi Zhang and Pengwei Wang and Zhongyuan Wang and Tiejun Huang and Lu Sheng and Shanghang Zhang},
booktitle={NeurIPS},
year={2026}
}

@inproceedings{
he2026drseg,
title={{DR}\${\textasciicircum}2\$Seg: Decomposed Two-Stage Rollouts for Efficient Reasoning Segmentation in Multimodal Large Language Models},
author={Yulin He and Wei Chen and jian zhikang and Tianhang Guo and Wenjuan Zhou and Minglong Li and Shaowu Yang and Wenjing Yang},
booktitle={ICML},
year={2026}
}

@inproceedings{
he2026rsagent,
title={{RSA}gent: Learning to Reason and Act via Multi-Turn Tool Invocations for Text-Guided Segmentation},
author={Xingqi He and Yujie Zhang and Shuyong Gao and Wenjie Li and Lingyi Hong and Mingxi Chen and Kaixun Jiang and Jiyuan Fu and Wenqiang Zhang},
booktitle={ICML},
year={2026}
}

@inproceedings{qian2026anchorseg,
  title={AnchorSeg: Language Grounded Query Banks for Reasoning Segmentation},
  author={Qian, Rui and Deng, Chuanhang and Huang, Qiang and Xiong, Jian and Li, Mingxuan and Zhou, Yingbo and Zhai, Wei and Chen, Jintao and Dou, Dejing},
  booktitle={ACL},
  year={2026}
}

@inproceedings{
liang2026segresearch,
title={Seg-ReSearch: Segmentation with Interleaved Reasoning and External Search},
author={Tianming Liang and Qirui Du and Jian-Fang Hu and Haichao Jiang and Zicheng Lin and Wei-Shi Zheng},
booktitle={ICML},
year={2026}
}

@inproceedings{zeng2026dgseg,
  title={DGSeg: Dynamic Gating of Semantic-Spatial Guided Predictions for Reasoning Segmentation},
  author={Zeng, Ruizhe and Cao, Siyu and Zhang, Lu and Liu, Zhiyong},
  booktitle={ECCV},
  year={2026}
}

@inproceedings{lu2026segcompass,
  title={SegCompass: Exploring Interpretable Alignment with Sparse Autoencoders for Enhanced Reasoning Segmentation},
  author={Lu, Zhenyu and Li, Liupeng and Wang, Jinpeng and Kang, Haoqian and Feng, Yan and Chen, Ke and Wang, Yaowei},
  booktitle={CVPR},
  year={2026}
}
\end{document}